%% file: template.tex
\RequirePackage{fix-cm}
\documentclass[twocolumn]{svjour3}          % twocolumn
\smartqed  % flush right qed marks, e.g. at end of proof

\usepackage{arrayjobx}
\usepackage{epsfig}
\usepackage{graphicx}
\usepackage{amssymb}
\usepackage{amsmath}
\usepackage{subcaption}
\usepackage{enumitem}
\usepackage{microtype}

\usepackage{lipsum}
\usepackage{cite}
\usepackage{epigraph}
\usepackage{multirow}
\usepackage{booktabs}
\usepackage{tabularx}
\usepackage{colortbl}
\usepackage{epstopdf}
\usepackage[font=footnotesize,labelfont=bf]{caption}
\usepackage{lib}
\usepackage{url}
\usepackage{mathptmx}      % use Times fonts if available on your TeX system
\newcommand{\dt}[0]{\Delta t}
\DeclareMathOperator*{\argmin}{arg\,min}
\newcommand{\secrefext}[1]{\secref{#1}}
\usepackage[pagebackref=true,breaklinks=true,letterpaper=true,colorlinks,bookmarks=false,citecolor=green]{hyperref}
\definecolor{Gray}{gray}{0.85}
\newcolumntype{g}{>{\columncolor{Gray}}c}
\definecolor{DarkGray}{gray}{0.65}
\newcolumntype{b}{>{\columncolor{DarkGray}}c}
\setlist[enumerate]{leftmargin=*}
\newcommand{\rgbd}[0]{{RGB-D}\xspace}
\newcommand{\resnet}[0]{{ResNet-50}\xspace}
\newcommand{\rgb}[0]{{RGB}\xspace}

\newcommand{\pparagraph}[1]{\textbf{#1}}
\newcommand{\tb}[1]{\textbf{#1}}
\newcommand{\eat}[1]{}
\newcommand{\dd}[0]{{depth}\xspace}
\newcommand{\rldagger}[0]{{\small DAGGER}\xspace}

\newcommand{\insertW}[2]{\IfFileExists{#2}{\includegraphics[width=#1\textwidth]{#2}}{\includegraphics[width=#1\textwidth]{figures/blank.png}}}
\newcommand{\insertWL}[2]{\IfFileExists{#2}{\includegraphics[width=#1\linewidth]{#2}}{\includegraphics[width=#1\linewidth]{figures/blank.png}}}
\newcommand{\insertH}[2]{\IfFileExists{#2}{\includegraphics[height=#1\textwidth]{#2}}{\includegraphics[height=#1\textwidth]{figures/blank.png}}}
\newcommand{\insertHW}[3]{\IfFileExists{#2}{\includegraphics[height=#1\textwidth,width=#2\textwidth]{#3}}{\includegraphics[height=#1\textwidth,width=#2\textwidth]{figures/blank.png}}}

\newcommand{\drl}[0]{DRL\xspace}

\begin{document}

\title{Cognitive Mapping and Planning for Visual Navigation\thanks{Work done
in part at Google, and in part at UC Berkeley.  Project website with simulator,
code, models, and videos:
\url{https://sites.google.com/view/cognitive-mapping-and-planning/}.}}


\author{Saurabh Gupta \and Varun Tolani \and James Davidson \\ Sergey Levine
\and Rahul Sukthankar \and Jitendra Malik}

\authorrunning{S. Gupta et al.} % if too long for running head

\institute{\footnotesize
S. Gupta$^{1,2}$ V. Tolani$^1$ J. Davidson$^2$ \\
S. Levine$^{1,2}$ R. Sukthankar$^1$ J. Malik$^{1,2}$ \\ 
$^1$UC Berkeley, $^2$Google \\
{\small \{sgupta, svlevine, malik\}@eecs.berkeley.edu, vtolani@berkeley.edu, \{jcdavidson,
sukthankar\}@google.com}}

\date{Received: date / Accepted: date}

% \input{coverletter}
\twocolumn

\maketitle

\begin{abstract}
We introduce a neural architecture for navigation in novel environments.  Our
proposed architecture learns to map from first-person views and plans a
sequence of actions towards goals in the environment. The Cognitive Mapper and
Planner (CMP) is based on two key ideas: a) a unified joint architecture for
mapping and planning, such that the mapping is driven by the needs of the
task, and b) a spatial memory with the ability to plan given an incomplete
set of observations about the world. CMP constructs a top-down belief map of
the world and applies a differentiable neural net planner to produce the next
action at each time step. The accumulated belief of the world enables the agent
to track visited regions of the environment.  We train and test CMP on
navigation problems in simulation environments derived from scans of real world
buildings.  Our experiments demonstrate that CMP outperforms alternate
learning-based architectures, as well as, classical mapping and path planning
approaches in many cases. Furthermore, it naturally extends to semantically
specified goals, such as ``going to a chair''.  We also deploy CMP on physical
robots in indoor environments, where it achieves reasonable performance, even
though it is trained entirely in simulation.
\end{abstract}

\input{intro}

\input{cmp}

\input{delta}

\input{related}

\input{recent_related}

\input{prelim}

\input{mapping}

\input{planning}

\input{joint}

\input{experiments}

\input{visualizations}

\input{realrobot}

\input{discussion}

\textbf{Acknowledgments:} We thank Shubham Tulsiani, David Fouhey, Somil Bansal and
Christian H\"{a}ne for useful discussion and feedback on the manuscript, as
well as Marek Fiser and Sergio Guadarrama for help with the Google
infrastructure and Tensorflow.

{\small
\bibliographystyle{spmpsci}      % mathematics and physical sciences
\bibliography{refs-vision,refs-da,refs-robo,refs-nav}
}

\clearpage

\renewcommand{\thefigure}{A\arabic{figure}}
\renewcommand{\thetable}{A\arabic{table}}
\setcounter{figure}{0}
\setcounter{table}{0}
\appendix
\renewcommand{\thesection}{A\arabic{section}}
\setcounter{section}{0}
\input{supp_text}

\end{document}

%% file: intro.tex
\begin{figure}
  \centering
  \insertWL{1.00}{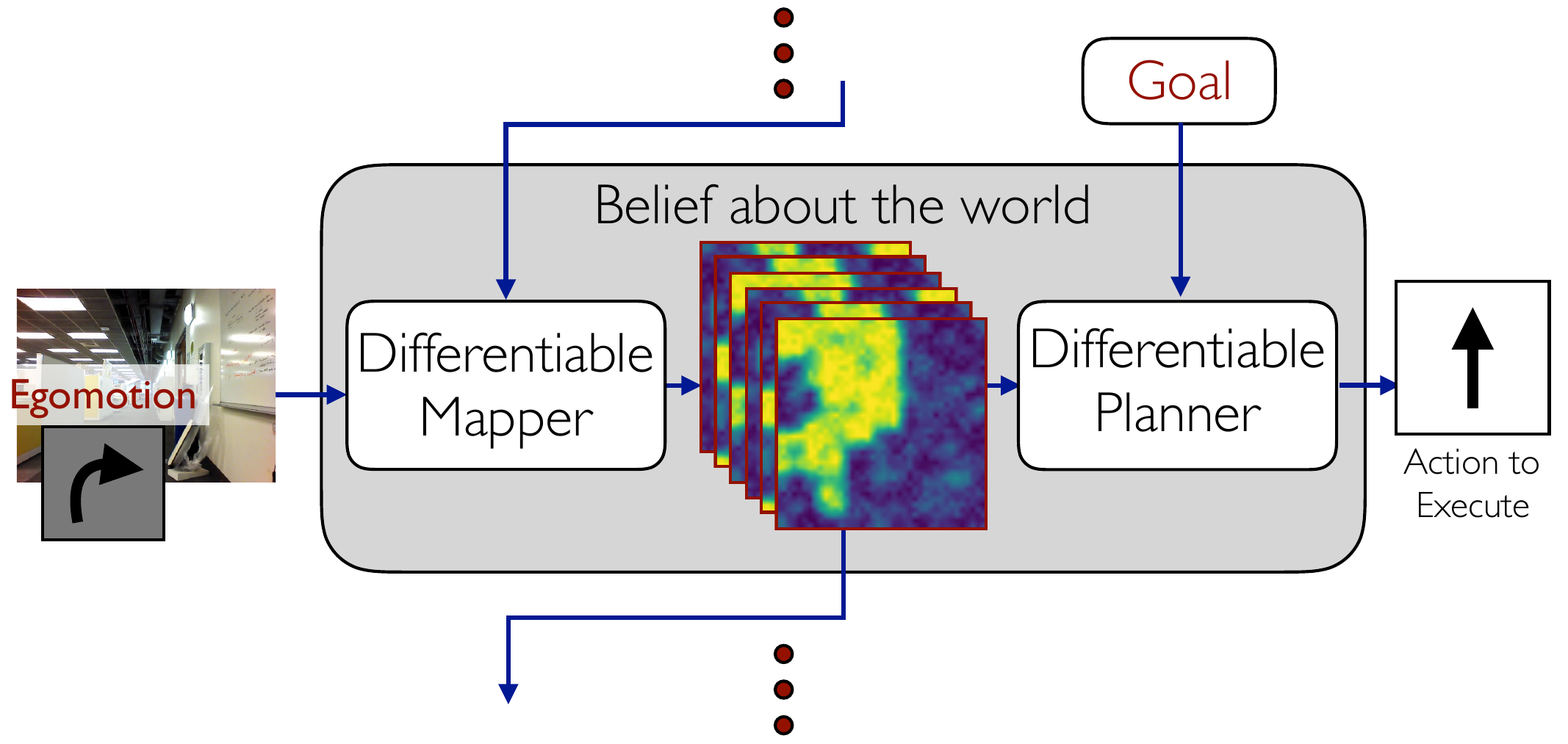}
  \includegraphics[trim={10cm 5cm 10cm 30cm},clip,width=1.0\linewidth]{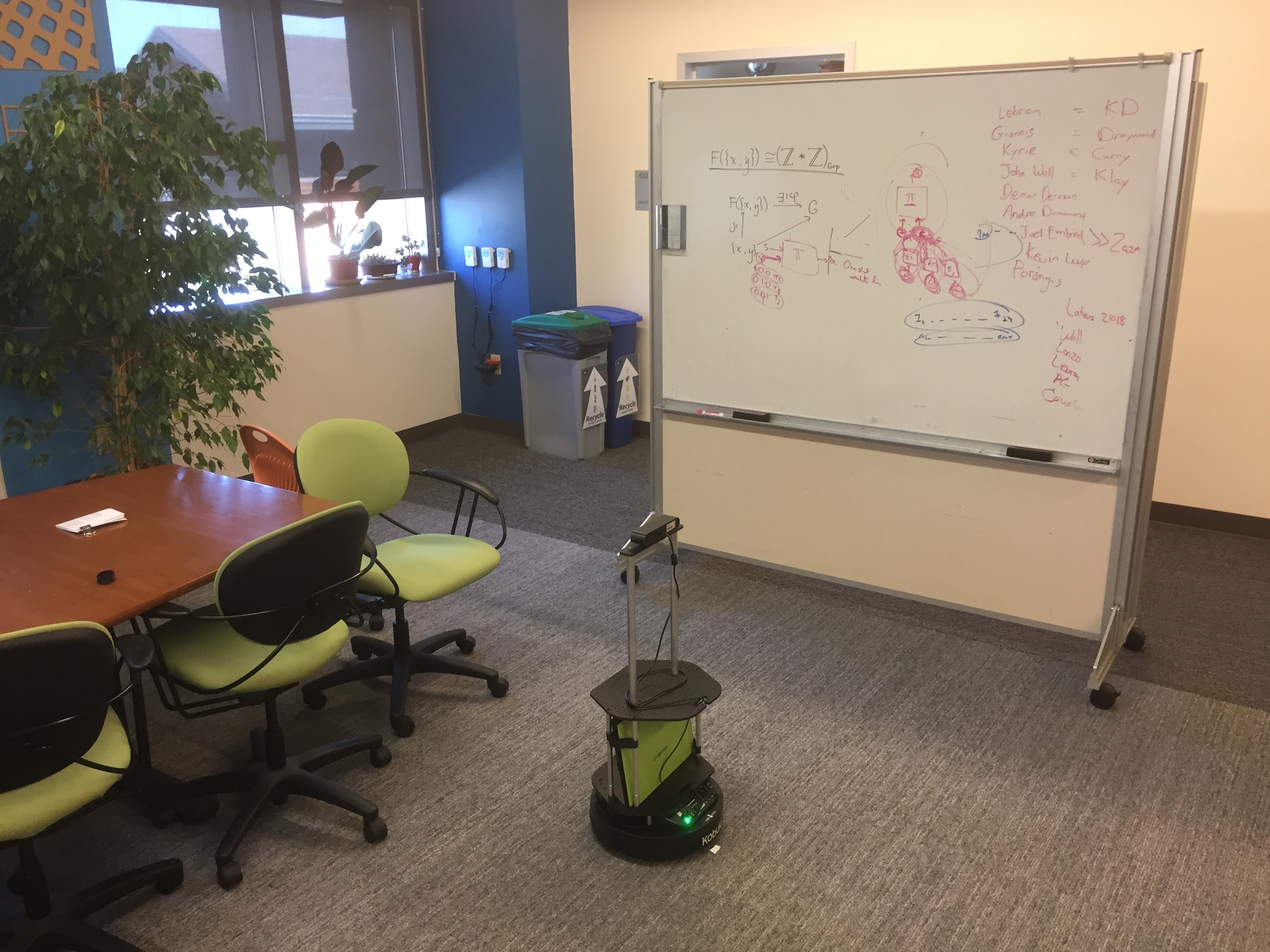}
  \caption{\textbf{Top: Network architecture:} Our learned navigation network
  consists of mapping and planning modules. The mapper writes into a latent
  spatial memory that corresponds to an egocentric map of the environment,
  while the planner uses this memory alongside the goal to output navigational
    actions. The map is not supervised explicitly, but rather emerges naturally
    from the learning process. \textbf{Bottom:} We also describe experiments
    where we deploy our learned navigation policies on a physical robot.}
  \figlabel{arch}
\end{figure}
\section{Introduction}
As humans, when we navigate through novel environments, we draw on our previous
experience in similar conditions.  We reason about free-space, obstacles and
the topology of the environment, guided by common sense rules and heuristics
for navigation. For example, to go from one room to another, I must first exit
the initial room; to go to a room at the other end of the building, getting
into a hallway is more likely to succeed than entering a conference room; a
kitchen is more likely to be situated in open areas of the building than in
the middle of cubicles. The goal of this paper is to design a learning
framework for acquiring such expertise, and demonstrate this for the problem
of robot navigation in novel environments.

However, classic approaches to navigation  rarely make use of such
common sense patterns.
Classical SLAM based approaches \cite{davison1998mobile,
thrun2005probabilistic} first build a 3D map using LIDAR, depth, or
structure from motion, and then plan paths in this map. These maps are built
purely geometrically, and nothing is known until it has been explicitly observed,
even when there are obvious patterns. This becomes a problem for goal directed
navigation. Humans can often guess, for example, where they will find a chair or that
a hallway will probably lead to another hallway
but a classical robot agent can
at best only do uninformed exploration. The separation between mapping and
planning also makes the overall system unnecessarily fragile. For example, the
mapper might fail on texture-less regions in a corridor, leading to failure of
the whole system, but precise geometry may not even be necessary if the robot
just has to keep traveling straight.

Inspired by this reasoning, recently there has been an increasing interest in
more end-to-end learning-based approaches that go directly from pixels to
actions \cite{zhu2016target, mnih2015human, levine2016end} without going
through explicit model or state estimation steps. These methods thus enjoy the
power of being able to learn behaviors from experience. However, it is
necessary to carefully design architectures that can capture the structure of
the task at hand. For instance Zhu \etal \cite{zhu2016target} use reactive
memory-less vanilla feed forward architectures for solving visual navigation
problems, In contrast, experiments by Tolman \cite{tolman1948cognitive} have
shown that even rats build sophisticated representations for space in the form
of `cognitive maps' as they navigate, giving them the ability to reason
about shortcuts, something that a reactive agent is unable to.

\begin{figure*}
  \centering
  \insertW{1.0}{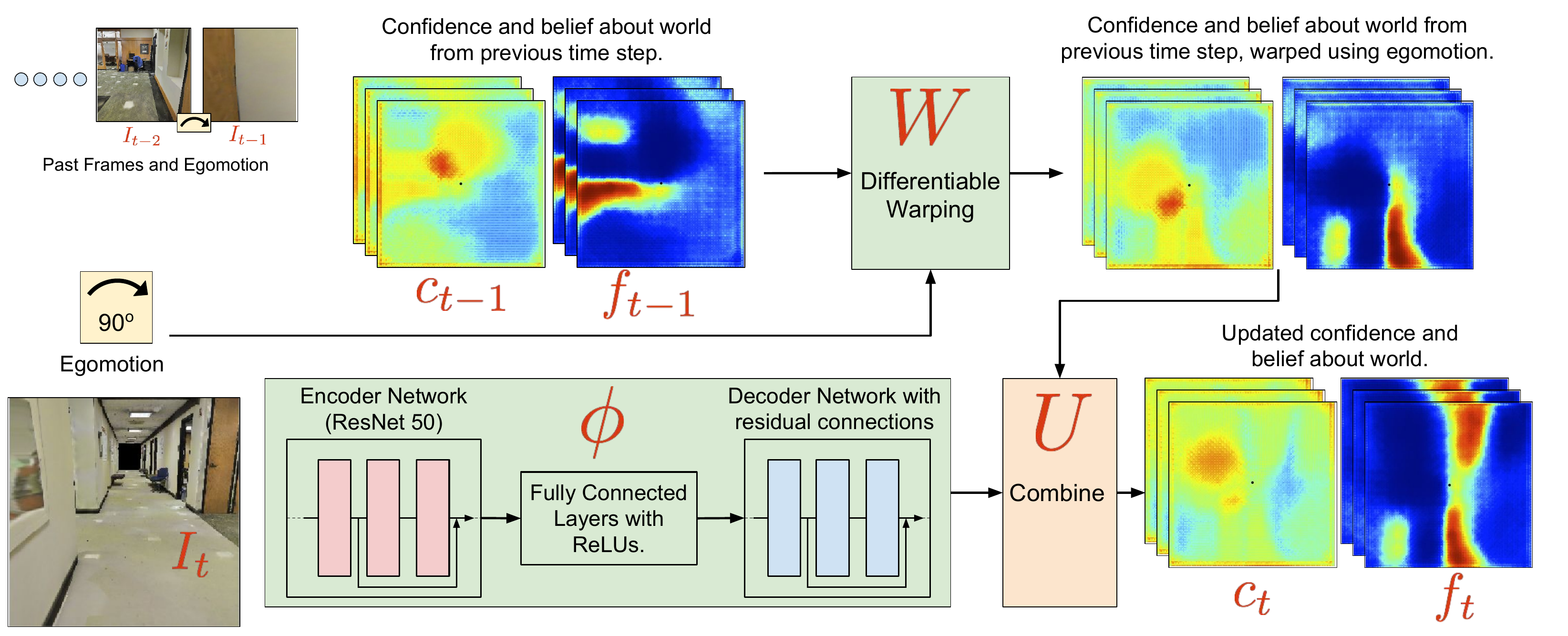}
  \caption{\textbf{Architecture of the mapper}: The mapper module processes
  first person images from the robot and integrates the observations into a
  latent memory, which corresponds to an egocentric map of the top-view of the
  environment. The mapping operation is not supervised explicitly -- the mapper
  is free to write into memory whatever information is most useful for the
  planner. In addition to filling in obstacles, the mapper also stores
  confidence values in the map, which allows it to make probabilistic
  predictions about unobserved parts of the map by exploiting learned
  patterns.}
  \figlabel{mapper}
\vspace{-0.5cm}
\end{figure*}

%% file: cmp.tex
This motivates our Cognitive Mapping and Planning (CMP) approach for visual
navigation (\figref{arch}). CMP consists of a) a spatial memory to capture the
layout of the world, and b) a planner that can plan paths given partial
information. The mapper and the planner are put together into a unified
architecture that can be trained to leverage regularities of the world.
The mapper fuses information from input views as observed by the agent over
time to produce a metric egocentric multi-scale belief about the world in a
top-down view. The planner uses this multi-scale egocentric belief of the world
to plan paths to the specified goal and outputs the optimal action to take.
This process is repeated at each time step to convey the agent to the goal.

At each time step, the agent updates the belief of the world from the previous
time step by a) using the ego-motion to transform the belief from the previous
time step into the current coordinate frame and b) incorporating information
from the current view of the world to update the belief. This allows the agent
to progressively improve its model of the world as it moves around.  The most
significant contrast with prior work is that our approach is trained end-to-end
to take good actions in the world. To that end, instead of analytically
computing the update to the belief (via classical structure from motion) we
frame this as a learning problem and train a convolutional neural network to
predict the update based on the observed first person view. We make the belief
transformation and update operations differentiable thereby allowing for
end-to-end training. This allows our method to adapt to the statistical
patterns in real indoor scenes without the need for any explicit supervision of
the mapping stage.
 
Our planner uses the metric belief of the world obtained through the mapping
operation described above to plan paths to the goal. We use value iteration as
our planning algorithm but crucially use a trainable, differentiable and
hierarchical version of value iteration. This has three advantages, a) being
trainable it naturally deals with partially observed environments by explicitly
learning when and where to explore, b) being differentiable it enables us to
train the mapper for navigation, and c) being hierarchical it allows us to plan
paths to distant goal locations in time complexity that is logarithmic in the
number of steps to the goal.

Our approach is a reminiscent of classical work in navigation that also
involves building maps and then planning paths in these maps to reach desired
target locations. However, our approach differs from classical work in the
following significant way: except for the architectural choice of maintaining a
metric belief, everything else is learned from data. This leads to some very
desirable properties: a) our model can learn statistical regularities of indoor
environments in a task-driven manner, b) jointly training the mapper and the
planner makes our planner more robust to errors of the mapper, and c) our model
can be used in an online manner in novel environments without requiring a
pre-constructed map.

%% file: delta.tex
This paper originally appeared at CVPR 2017. In this journal article, we
additionally describe real world deployment of our learned policies on a
TurtleBot~2 platform, and report results of our deployment on indoor test
environments. We have also incorporated feedback from the community.  In
particular, we have added comparisons to a policy that very closely resembles a
classical mapping and planning method. We have also included more
visualizations of the representations produced by the mapper. Finally, we
situate the current work in context of the new research directions that are
emerging in the field at the intersection of machine learning, robotics and
computer vision.

%% file: related.tex
\section{Related Work}
\seclabel{related} Navigation is one of the most fundamental problems in mobile
robotics.  The standard approach is to decompose the problem into two separate
stages: (1) mapping the environment, and (2) planning a path through the
constructed map \cite{khatib1986real, elfes1987sonar}. Decomposing navigation
in this manner allows each stage to be developed independently, but prevents
each from exploiting the specific needs of the other.  A comprehensive survey
of classical approaches for mapping and planning can be found in
\cite{thrun2005probabilistic}.

Mapping has been well studied in computer vision and robotics in the form of
structure from motion and simultaneous localization and mapping
\cite{slam-survey:2015, izadiUIST11, henry2010rgb, snavely2008modeling} with a
variety of sensing modalities such as range sensors, \rgb cameras and \rgbd
cameras. These approaches take a purely geometric approach.  Learning based
approaches \cite{zamir2016generic, hadsell2009learning} study the problem in
isolation thus only learning generic task-independent maps.  Path planning in
these inferred maps has also been well studied, with pioneering works from
Canny \cite{canny1988complexity}, Kavraki \etal \cite{kavraki1996probabilistic} and
LaValle and Kuffner \cite{lavalle2000rapidly}. Works such as \cite{elfes1989using,
fraundorfer2012vision} have studied the joint problem of mapping and planning.
While this relaxes the need for pre-mapping by incrementally updating the map
while navigating, but still treat navigation as a  purely geometric problem,
Konolige \etal \cite{konolige2010view} and Aydemir \etal
\cite{aydemir2013active} proposed approaches which leveraged semantics for more
informed navigation. Kuipers \etal \cite{kuipers1991robot}
introduce a cognitive mapping model using hierarchical abstractions of maps.
Semantics have also been associated with 3D environments more generally
\cite{koppulaNIPS11,guptaCVPR13}.

As an alternative to separating out discrete mapping and planning phases,
reinforcement learning (RL) methods directly learn policies for robotic
tasks~\cite{kim2003autonomous, peters2008reinforcement, kohl2004policy}. A
major challenge with using RL for this task is the need to process complex
sensory input, such as camera images.  Recent works in deep reinforcement
learning (\drl) learn policies in an end-to-end manner \cite{mnih2015human}
going from pixels to actions. Follow-up works \cite{mnih2016asynchronous,
gu2016continuous, schulman2015trust} propose improvements to \drl algorithms,
\cite{oh2016control, mnih2016asynchronous, wierstra2010recurrent,
heess2015memory, zhang2016learning} study how to incorporate memory into such
neural network based models. We build on the work from Tamar \etal
\cite{tamar2016value} who study how explicit planning can be incorporated in
such agents, but do not consider the case of first-person visual navigation,
nor provide a framework for memory or mapping.  \cite{oh2016control} study the
generalization behavior of these algorithms to novel environments they have not
been trained on.

In context of navigation, learning and \drl has been used to obtain policies
\cite{toussaint2003learning, zhu2016target, oh2016control, tamar2016value,
kahn2016plato, giusti2016machine, daftry2016learning,
abel2016exploratory}. Some of these works \cite{kahn2016plato,
giusti2016machine}, focus on the problem of learning controllers for
effectively maneuvering around obstacles directly from raw sensor data. Others,
such as \cite{tamar2016value, blundell2016model, oh2016control}, focus on the
planning problem associated with navigation under full state information
\cite{tamar2016value}, designing strategies for faster learning via episodic
control \cite{blundell2016model}, or incorporate memory into \drl algorithms to
ease generalization to new environments. Most of this research (except
\cite{zhu2016target}) focuses on navigation in synthetic mazes which have
little structure to them.  Given these environments are randomly generated, the
policy learns a random exploration strategy, but has no statistical
regularities in the layout that it can exploit. We instead test on layouts
obtained from real buildings, and show that our architecture consistently
outperforms feed forward and LSTM models used in prior work.

The research most directly relevant to our work is the contemporary work of Zhu
\etal \cite{zhu2016target}. Similar to us, Zhu \etal also study first-person
view navigation using macro-actions in more realistic environments instead of
synthetic mazes. Zhu \etal propose a feed forward model which when trained in
one environment can be finetuned in another environment. Such a memory-less
agent cannot map, plan or explore the environment, which our expressive model
naturally does. Zhu \etal also don't consider zero-shot generalization to
previously unseen environments, and focus on smaller worlds where memorization
of landmarks is feasible. In contrast, we explicitly handle generalization to
new, never before seen interiors, and show that our model generalizes
successfully to floor plans not seen during training.

%% file: recent_related.tex
\textbf{Relationship to contemporary research.}
In this paper, we used scans of real world environments to construct visually
realistic simulation environments to study representations that can enable
navigation in novel previously unseen environments. Since conducting this
research, over the last year, there has been a major thrust in this direction
in computer vision and related communities. Numerous works such as
\cite{Matterport3D, dai2017scannet, active-vision-dataset2017} have collected
large-scale datasets consisting of scans of real world environments, while
\cite{savva2017minos, wu2018building, xiazamirhe2018gibsonenv} have built more
sophisticated simulation environments based on such scans. A related and
parallel stream of research studies whether or not models trained in simulators
can be effectively transferred to the real world \cite{sadeghi2016cadrl,
bruce2018learning}, and how the domain gap between simulation and the real
world may be reduced \cite{xiazamirhe2018gibsonenv}. A number of works have
studied related navigation problems in such simulation environments
\cite{henriques2018mapnet, chaplot2018active, anderson2017vision}.  Researchers
have also gone beyond specifying goals as a desired location in space and
finding objects of interest as done in this paper, for example, Wu \etal
\cite{wu2018building} generalize the goal specification to also include rooms
of interest, and Das \etal \cite{embodiedqa} allow goal specification via
templated questions.  Finally, a number of works have also pursued the problem
of building representation for space in context of navigation.
\cite{parisotto2017neural, bhatti2016playing, khan2017memory, gordon2017iqa}
use similar $2D$ spatial representations, Mirowski \etal
\cite{mirowski2016learning} use fully-connected LSTMs, while Savinov \etal
\cite{savinov2018semi} develop topological representations. Interesting
reinforcement learning techniques have also been explored for the task of
navigation \cite{mirowski2016learning, duan2016rl}.

%% file: prelim.tex
\section{Problem Setup}
\seclabel{prelim}
To be able to focus on the high-level mapping and planning problem we remove
confounding factors arising from low-level control by conducting our
experiments in simulated real world indoor environments. Studying the problem
in simulation makes it easier to run exhaustive evaluation experiments, while
the use of scanned real world environments allows us to retains the richness and complexity of real scenes.
We also only study the static version of the problem, though extensions to dynamic
environments would be interesting to explore in future work. 

We model the robot as a cylinder of a fixed radius and height, equipped
with vision sensors (\rgb cameras or \dd cameras) mounted at a fixed height and
oriented at a fixed pitch. The robot is equipped with low-level controllers
which provide relatively high-level macro-actions $\mathcal{A}_{x,\theta}$.
These macro-actions are a) stay in place, b) rotate left by $\theta$, c) rotate
right by $\theta$, and d) move forward $x$ cm, denoted by $a_0, a_1, a_2 \text{
and } a_3$, respectively. We further assume that the environment is a grid world
and the robot uses its macro-actions to move between nodes on this graph.
The robot also has access to its precise egomotion. This amounts to assuming
perfect visual odometry \cite{nister2004visual}, which can itself be learned
\cite{haarnoja2016backprop}, but we defer the joint learning problem to future
work.

We want to learn policies for this robot for navigating in \textit{novel}
environments that it has not previously encountered.  We study two navigation
tasks, a \textit{geometric} task where the robot is required to go to a target
location specified in robot's coordinate frame (\eg $250cm$ forward, $300cm$
left) and a \textit{semantic} task where the robot is required to go to an
object of interest (\eg a chair). These tasks are to be performed in novel
environments, neither the exact environment map nor its topology is available
to the robot. 

Our navigation problem is defined as follows.  At a given time step $t$, let us
assume the robot is at a global position (position in the world coordinate
frame) $P_t$. At each time step the robot receives as input the image of the
environment $\mathcal{E}$, $I_t = I(\mathcal{E}, P_t)$ and a target location
$(x^g_t, y^g_t, \theta^g_t)$ (or a semantic goal) specified in the coordinate
frame of the robot. The navigation problem is to learn a policy that at every
time steps uses these inputs (current image, egomotion and target
specification) to output the action that will convey the robot to the target as
quickly as possible.

\pparagraph{Experimental Testbed.}
We conduct our experiments on the Stanford large-scale 3D Indoor Spaces (S3DIS)
dataset introduced by Armeni \etal \cite{armeni20163d}. The dataset consists of
3D scans (in the form of textured meshes) collected in 6 large-scale indoor
areas that originate from 3 different buildings of educational and office use.
The dataset was collected using the Matterport scanner \cite{matterport}. Scans
from 2 buildings were used for training and the agents were tested on scans
from the 3rd building. We pre-processed the meshes to compute space traversable
by the robot. We also precompute a directed graph $\mathcal{G}_{x,\theta}$
consisting of the set of locations the robot can visit as nodes and a
connectivity structure based on the set of actions $\mathcal{A}_{x,\theta}$
available to the robot to efficiently generate training problems. More details
in \secrefext{envvis}.

%% file: mapping.tex
\section{Mapping}
\seclabel{mapping}
We describe how the mapping portion of our learned network can integrate
first-person camera images into a top-down 2D representation of the
environment, while learning to leverage statistical structure in the world.
Note that, unlike analytic mapping systems, the map in our model amounts to a
latent representation. Since it is fed directly into a learned planning module,
it need not encode purely free space representations, but can instead function
as a general spatial memory. The model learns to store inside the map whatever
information is most useful for generating successful plans.  However to make
description in this section concrete, we assume that the mapper predicts free
space.

The mapper architecture is illustrated in \figref{mapper}.  At every time step
$t$ we maintain a cumulative estimate of the free space $f_t$ in the coordinate
frame of the robot. $f_t$ is represented as a multi-channel 2D feature map that
metrically represents space in the top-down view of the world.  $f_t$ is
estimated from the current image $I_t$, cumulative estimate from the
previous time step $f_{t-1}$ and egomotion between the last and this step
$e_{t}$ using the following update rule:
\begin{eqnarray}
  f_t = U\left( W\left(f_{t-1}, e_{t}\right), f'_{t}\right) \quad \text{where}, f'_{t} = \phi\left(I_t \right).
\end{eqnarray}
here, $W$ is a function that transforms the free space prediction from the
previous time step $f_{t-1}$ according to the egomotion in the last step
$e_{t}$, $\phi$ is a function that takes as input the current image $I_t$ and
outputs an estimate of the free space based on the view of the environment from
the current location (denoted by $f'_{t}$). $U$ is a function which accumulates
the free space prediction from the current view with the accumulated prediction
from previous time steps. 
Next, we describe how each of the functions $W$, $\phi$ and $U$ are realized.

\begin{figure*}
\centering
  \insertW{0.98}{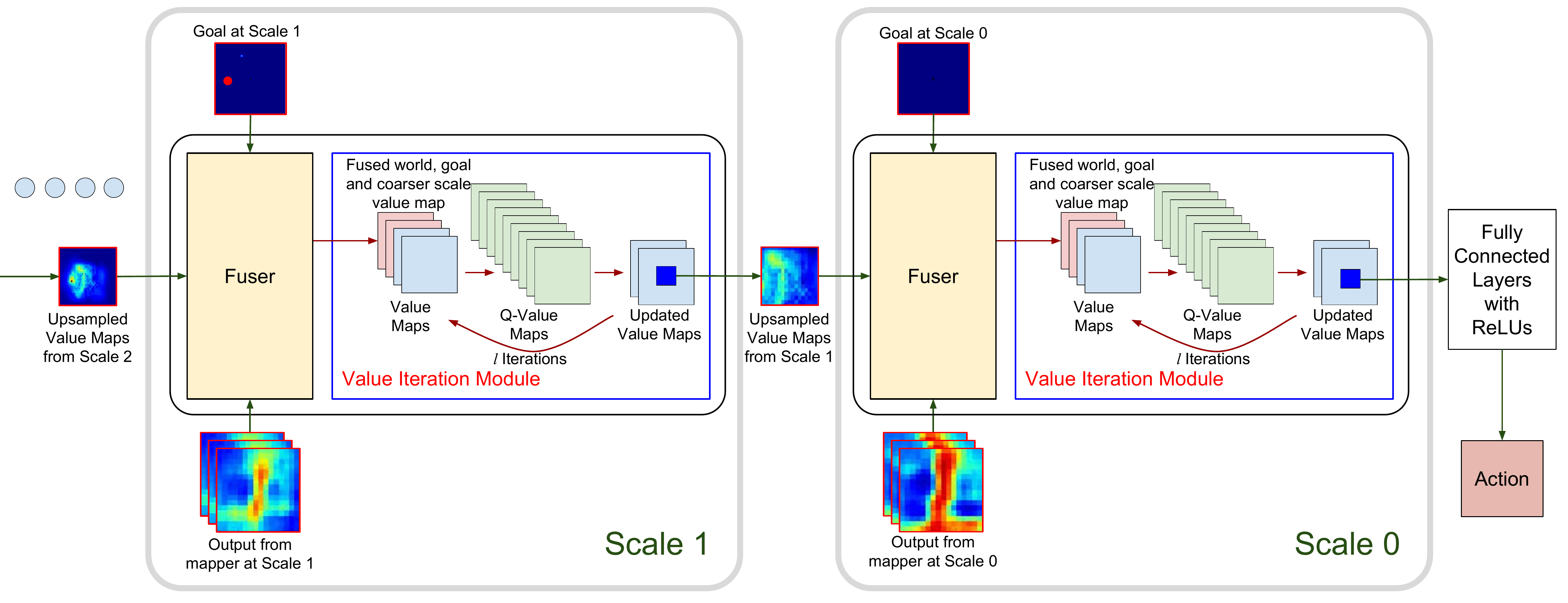}
  \caption{\textbf{Architecture of the hierarchical planner}: The hierarchical
planner takes the egocentric multi-scale belief of the world output by the
mapper and uses value iteration expressed as convolutions and channel-wise
max-pooling to output a policy. The planner is trainable and differentiable and
back-propagates gradients to the mapper. The planner operates at multiple
scales (scale 0 is the finest scale) of the problem which leads to efficiency
in planning.}
  \figlabel{planner}
\end{figure*}

The function $W$ is realized using bi-linear sampling. Given the ego-motion, we
compute a backward flow field $\rho(e_{t})$.  This backward flow maps each
pixel in the current free space image $f_t$ to the location in the previous
free space image $f_{t-1}$ where it should come from.  This backward flow
$\rho$ can be analytically computed from the ego-motion (as shown in
\secrefext{ego}). 
The function $W$ uses bi-linear sampling to apply this flow field to
the free space estimate from the previous frame. Bi-linear sampling allows us
to back-propagate gradients from $f_t$ to $f_{t-1}$ \cite{jaderberg2015spatial},
which will make it possible to train this model end to end.

The function $\phi$ is realized by a convolutional neural network. Because of
our choice to represent free space always in the coordinate frame of the robot,
this becomes a relatively easy function to learn, given the network only has to
output free space in the current coordinate, rather than in an arbitrary world
coordinate frame determined by the cumulative egomotion of the robot so far.

Intuitively, the network can use semantic cues
(such as presence of scene surfaces like floor and walls, common furniture
objects like chairs and tables) alongside other learned priors about size and
shapes of common objects to generate free space estimates, even for object
that may only be partiality visible. Qualitative results in
\secrefext{mapper_vis} show an example for this where our proposed mapper
is able to make predictions for spaces that haven't been observed. 

The architecture of the neural network that realizes function $\phi$ is shown
in \figref{mapper}. It is composed of a convolutional encoder
which uses residual
connections \cite{he2015deep} and produces a representation of the scene in the
2D image space. This representation is transformed into one that is in the
egocentric 2D top-down view via fully connected layers. This representation is
up-sampled using up-convolutional layers (also with residual connections) to
obtain the update to the belief about the world from the current frame.

In addition to producing an estimate of the free space from the current view
$f'_{t}$ the model also produces a confidence $c'_t$.  This estimate is also
warped by the warping function $W$ and accumulated over time into $c_t$. This
estimate allows us to simplify the update function, and can be thought of as
playing the role of the update gate in a gated recurrent unit.
The update function $U$ takes in the tuples $(f_{t-1}, c_{t-1})$, and
$(f'_{t}, c'_t)$ and produces $(f_t, c_t)$ as follows:
\begin{eqnarray}
  f_t = \frac{f_{t-1}c_{t-1} + f'_{t}c'_{t}}{c_{t-1} + c'_{t}} \quad \text{and} \quad c_t = c_{t-1} + c'_{t}
\end{eqnarray}
We chose an analytic update function to keep the overall architecture simple.
This can be replaced with more expressive functions like those realized by
LSTMs \cite{hochreiter1997long}.

\pparagraph{Mapper performance in isolation.}
To demonstrate that our proposed mapper architecture works we test it in
isolation on the task of free space prediction. \secrefext{mapper_performance}
shows qualitative and quantitative results.

%% file: planning.tex
\section{Planning}
\seclabel{planning}
Our planner is based on value iteration networks proposed by Tamar \etal
\cite{tamar2016value}, who observed that a particular type of planning
algorithm called value iteration \cite{bellman1957markovian} can be implemented
as a neural network with alternating convolutions and channel-wise max pooling
operations, allowing the planner to be differentiated with respect to its
inputs. Value iteration can be thought of as a generalization of Dijkstra's
algorithm, where the value of each state is iteratively recalculated at each
iteration by taking a max over the values of its neighbors plus the reward of
the transition to those neighboring states.
This plays nicely with 2D grid world navigation problems, where these
operations can be implemented with small $3 \times 3$ kernels followed by
max-pooling over channels. Tamar \etal \cite{tamar2016value} also showed that
this reformulation of value iteration can also be used to learn the planner
(the parameters in the convolutional layer of the planner) by providing
supervision for the optimal action for each state. Thus planning can be done in
a trainable and differentiable manner by very deep convolutional network (with
channel wise max-pooling).
For our problem, the mapper produces the 2D top-view of the world
which shares the same 2D grid world structure as described above, and we use
value iteration networks as a trainable and differentiable planner.

\pparagraph{Hierarchical Planning.}
Value iteration networks as presented in \cite{tamar2016value}(v2) are
impractical to use for any long-horizon planning problem. This is because the
planning step size is coupled with the action step size thus leading to a) high
computational complexity at run time, and b) a hard learning problem as
gradients have to flow back for as many steps. To alleviate this problem, we
extend the hierarchical version presented in \cite{tamar2016value}(v1).

Our hierarchical planner plans at multiple spatial scales. We start with a $k$
times spatially downsampled environment and conduct $l$ value iterations in
this downsampled environment. The output of this value iteration process is
center cropped, upsampled, and used for doing value iterations at a finer
scale. This process is repeated to finally reach the resolution of the original
problem. This procedure allows us to plan for goals which are as far as $l2^k$
steps away while performing (and backpropagating through) only $lk$ planning
iterations.  This efficiency increase comes at the cost of approximate
planning.

\pparagraph{Planning in Partially Observed Environments.}
Value iteration networks have only been evaluated when the environment is fully
observed, \ie the entire map is known while planning.  However, for our
navigation problem, the map is only partially observed.  Because the planner is
not hand specified but learned from data, it can learn policies which naturally
take partially observed maps into account. Note that the mapper produces not
just a belief about the world but also an uncertainty $c_t$, the planner knows
which parts of the map have and haven't been observed.

%% file: joint.tex
\section{Joint Architecture}
\seclabel{joint}

Our final architecture, Cognitive Mapping and Planning (CMP) puts together the
mapper and planner described above. At each time step, the mapper updates its
multi-scale belief about the world based on the current observation. This
updated belief is input to the planner which outputs the action to take. As
described previously, all parts of the network are differentiable and allow for
end-to-end training, and no additional direct supervision is used to train the
mapping module -- rather than producing maps that match some ground truth free
space, the mapper produces maps that allow the planner to choose effective
actions.

\textbf{Training Procedure.} We optimize the CMP network with fully supervised
training using \rldagger \cite{ross2011reduction}. We generate training
trajectories by sampling arbitrary start and goal locations on the graph
$\mathcal{G}_{x,\theta}$. We generate supervision for training by computing
shortest paths on the graph. We use an online version of \rldagger, where
during each episode we sample the next state based on the action from the
agent's current policy, or from the expert policy. We use scheduled sampling
and anneal the probability of sampling from the expert policy using
inverse sigmoid decay.

Note that the focus of this work is on studying different architectures for
navigation. Our proposed architecture can also be trained with alternate
paradigms for learning such policies, such as reinforcement learning. We chose
\rldagger for training our models because we found it to be significantly more sample
efficient and stable in our domain, allowing us to focus on the architecture design.

%% file: experiments.tex
\section{Experiments}
\seclabel{experiments}
The goal of this paper is to learn policies for visual navigation for different
navigation tasks in novel indoor environments. We first describe these
different navigation tasks, and performance metrics.  We then discuss different
comparison points that quantify the novelty of our test environments,
difficulty of tasks at hand. Next, we compare our proposed CMP architecture to
other learning-based methods and to classical mapping and planning based
methods. We report all numbers on the test set. The test set consists of a
floor from an altogether different building not contained in the training set.
(See dataset website and \secrefext{envvis} for environment visualizations.)

\textbf{Tasks.} We study two tasks: a \textit{geometric task}, where the goal is
to reach a point in space, and a \textit{semantic task}, where the goal is to
find objects of interest. We provide more details about both these tasks below:
\begin{enumerate}
\item \textbf{Geometric Task:} The goal is specified geometrically in terms of
position of the goal in robot's coordinate frame. Problems for this task are
generated by sampling a start node on the graph and then sampling an end node
which is within 32 steps from the starting node and preferably in another room
or in the hallway (we use room and hallway annotations from the
dataset~\cite{armeni20163d}). This is same as the \textit{PointGoal} task as
described in \cite{anderson2018evaluation}. 
\item \textbf{Semantic Task:} We consider three tasks: `go to a chair', `go to
a door' and `go to a table'. The agent receives a one-hot vector indicating the
object category it must go to and is considered successful if it can reach
\textit{any} instance of the indicated object category.  We use object
annotations from the S3DIS dataset \cite{armeni20163d} to setup this task. We
initialize the agent such that it is within 32 time steps of at least one
instance of the indicated category, and train it to go towards the nearest
instance. This is same as the \textit{ObjectGoal} task as described in
\cite{anderson2018evaluation}. 
\end{enumerate}
The same sampling process is used during training and testing.  For testing, we
sample 4000 problems on the test set.  The test set consists of a floor from an
altogether different building not contained in the training set.  These
problems remain fixed across different algorithms that we compare. We measure
performance by measuring the distance to goal after running the policies for a
maximum number of time steps (200), or if they emit the \textit{stop} action.

\textbf{Performance Metrics.}
We report multiple performance metrics: a) the mean distance to goal, b) the
75\textsuperscript{th} percentile distance to goal, and c) the success rate
(the agent succeeds if it is within a distance of $3$ steps of the goal
location) as a function of number of time-steps. We plot these metrics as a
function of time-steps and also report performance at $39$ and $199$ time steps
in the various tables. For the most competitive methods, we also report the SPL
metric\footnote{For computing SPL, we use the shortest-path on the graph as the
shortest-path length. We count both rotation and translation actions for both
the agent's path and the shortest path. An episode is considered successful, if
the agent ends up within 3 steps of the goal location. For the geometric task,
we run the agent till it outputs the `stay-in-place' action, or for a maximum
of 200 time steps. For the semantic task, we train a separate `stop' predictor.
This 'stop' predictor is trained to predict if the agent is within 3 steps of
the goal or not. The probability at which the episode should be terminated is
determined on the validation set.} (higher is better) as introduced in
\cite{anderson2018evaluation}. In addition to measuring whether the agent
reaches the goal, SPL additionally also measures the efficiency of the path
used and whether the agent reliably determines that it has reached the goal or
not. 

\textbf{Training Details.}
Models are trained asynchronously using TensorFlow \cite{tf}. We used
ADAM~\cite{kingma2014adam} to optimize our loss function and trained for 60K
iterations with a learning rate of $0.001$ which was dropped by a factor of 10
every 20K iterations (we found this necessary for consistent training across
different runs). We use weight decay of $0.0001$ to regularize the network and
use batch-norm \cite{ioffe2015batch}.  We use \resnet \cite{he2016identity}
pre-trained on ImageNet \cite{imagenet_cvpr09} to represent \rgb images. We
transfer supervision from \rgb images to \dd images using cross modal
distillation \cite{gupta2016cross} between \rgbd image pairs rendered from
meshes in the training set to obtain a pre-trained \resnet model to represent
\dd images.

\subsection{Baselines}
Our experiments are designed to test performance at visual navigation in novel
environments. We first quantify the differences between training and test
environments using a nearest neighbor trajectory method. We next quantify the
difficulty of our environments and evaluation episodes by training a blind
policy that only receives the relative goal location at each time step.
Next, we test the effectiveness of our memory-based architecture. We compare
to a purely reactive agent to understand the role of memory for this task, and
to a LSTM-based policy to test the effectiveness of our specific memory
architecture. Finally, we make comparisons with classical mapping and planning
based techniques.  Since the goal of this paper is to study various
architectures for navigation we train all these architectures the same way
using \rldagger \cite{ross2011reduction} as described earlier. We provide more
details for each of these baselines below. 

\begin{enumerate}
\item \textbf{Nearest Neighbor Trajectory Transfer}: To quantify similarity between
training and testing environments, we transfer optimal trajectories from the
train set to the test set using visual nearest neighbors (in \rgb \  \resnet
feature space). At each time step, we pick the location in the training set
which results in the most similar view to that seen by the agent at the current
time step. We then compute the optimal action that conveys the robot to the
same relative offset in the training environment from this location and execute
this action at the current time step. This procedure is repeated at each time
step. Such a transfer leads to very poor results. 
\item \textbf{No image, goal location only with LSTM}: Here, we ignore the image and
simply use the relative goal location (in robot's current coordinate frame) as
input to a LSTM, and predict the action that the agent should take. The
relative goal location is embedded into a $K$ dimensional space via fully
connected layers with ReLU non-linearities before being input to the LSTM. 
\item \textbf{Reactive Policy, Single Frame}: We next compare to a reactive agent
that uses the first-person view of the world. As described above we use \resnet
to extract features. These features are passed through a few fully connected
layers, and combined with the representation for the relative goal location
which is used to predict the final action. We experimented with additive and
multiplicative combination strategies and both performed similarly. 
\item \textbf{Reactive Policy, Multiple Frames}: We also consider the case where the
reactive policy receives 3 previous frames in addition to the current view.
Given the robot's step-size is fairly large we consider a late fusion
architecture and fuse the information extracted from \resnet. Note that this
architecture is similar to the one used in \cite{zhu2016target}. The primary
differences are: goal is specified in terms of relative offset (instead of an
image), training uses \rldagger (which utilizes denser supervision) instead of
A3C, and testing is done in novel environments. These adaptations are necessary
to make an interpretable comparison on our task. 
\item \textbf{LSTM Based Agent}: Finally, we also compare to an agent which uses an
LSTM based memory. We introduce LSTM units on the multiplicatively combined
image and relative goal location representation. Such an architecture also
gives the LSTM access to the egomotion of the agent (via how the relative goal
location changes between consecutive steps). Thus this model has access to all
the information that our method uses. We also experimented with other LSTM
based models (ones without egomotion, inputting the egomotion more explicitly,
\etc), but weren't able to reliably train them in early experiments and did not
pursue them further. 
\item \textbf{Purely Geometric Mapping}: We also compare to a
purely geometric incremental mapping and path planning policy. We projected
observed 3D points incrementally into a top-down occupancy map using the ground
truth egomotion and camera extrinics and intrinsics. When using depth images as
input, these 3D points are directly available. When using \rgb images as input,
we triangulated SIFT feature points in the \rgb images (registered using the
ground truth egomotion) to obtain the observed 3D points (we used the COLMAP
library \cite{schoenberger2016sfm}). This occupancy map was used to compute a
grid-graph (unoccupied cells are assumed free). For the \underline{\smash{geometric
task}}, we mark the goal location on this grid-graph and execute the action that
minimizes the distance to the goal node.  For the \underline{\smash{semantic task}}, we
use purely geometric exploration along with a semantic segmentation network
trained\footnote{We train this semantic segmentation network to segment chairs,
doors and table on the S3DIS dataset \cite{armeni20163d}.} 
to identify object categories of interest. The agent systematically
explores the environment using frontier-based exploration
\cite{yamauchi1997frontier}\footnote{We sample a \textit{goal} location outside
the map, and try to reach it, as for the geometric task. As there is no path to
this location, the agent ends up systematically exploring the environment.}
till it detects the specified category using the semantic segmentation network.
These labels are projected onto the occupancy map using the 3D points, and
nodes in the grid-graph are labelled as goals. We then output the action that
minimizes the distance to these inferred goal nodes.

For this baseline, we experimented with different input image sizes, and
increased the frequency at which \rgb or depth images were captured. We
validated a number of other hyper-parameters: a) number of points in a cell
before it is considered occupied, b) number of intervening cell to be occupied
before it is considered non-traversable, c) radius for morphological opening of
the semantic labels on the map. 3D reconstruction from \rgb images was
computationally expensive, and thus we report comparisons to these classical
baselines on a subset of the test cases. 
\end{enumerate}

\subsection{Results}
\input{geom_spl}

\begin{figure*}
\centering
{\insertW{0.490}{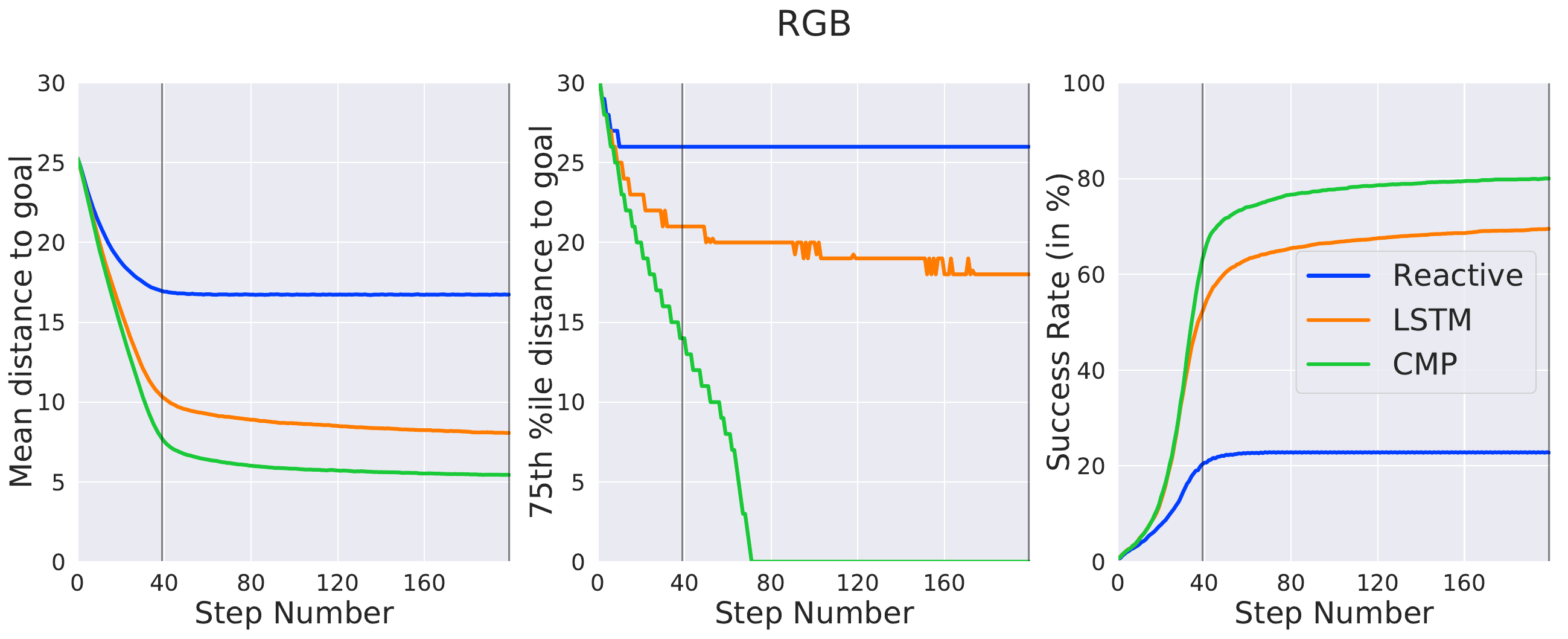}} \hfill \unskip\ \vrule\  \hfill 
{\insertW{0.490}{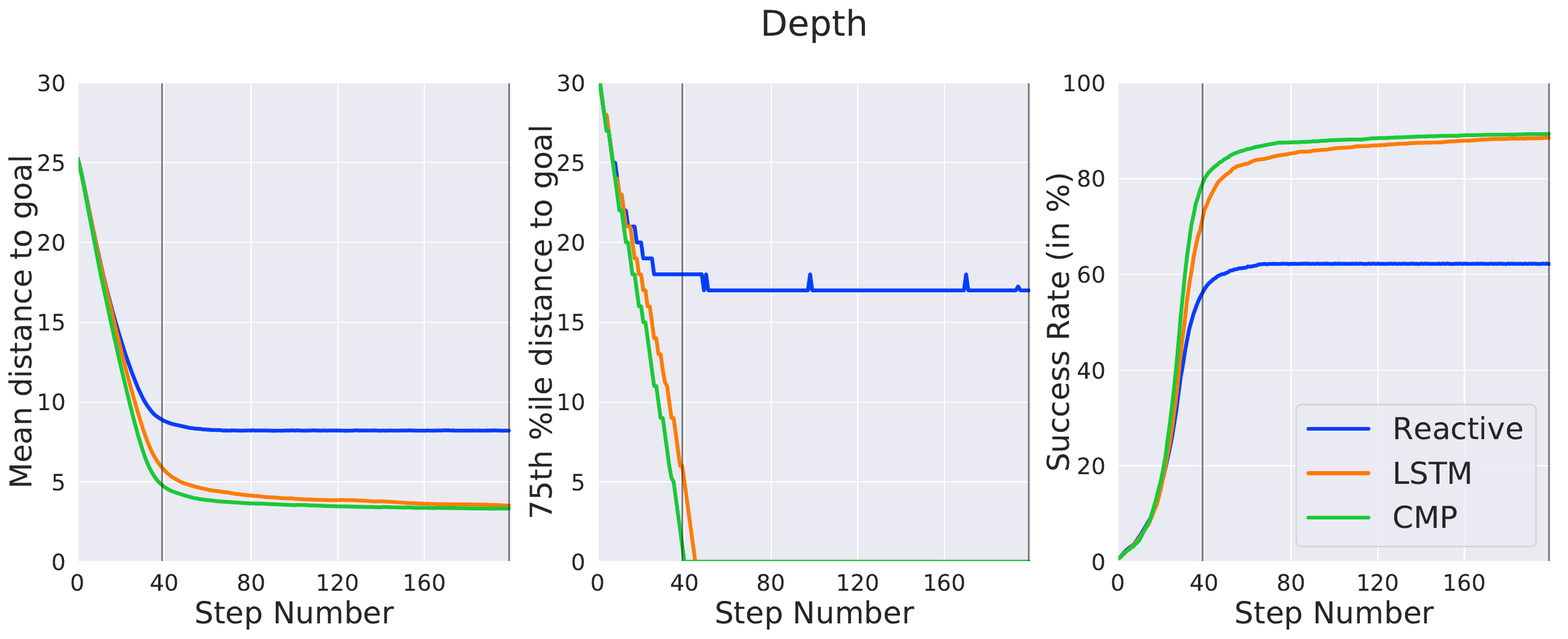}} \\ % \midrule
{\insertW{0.490}{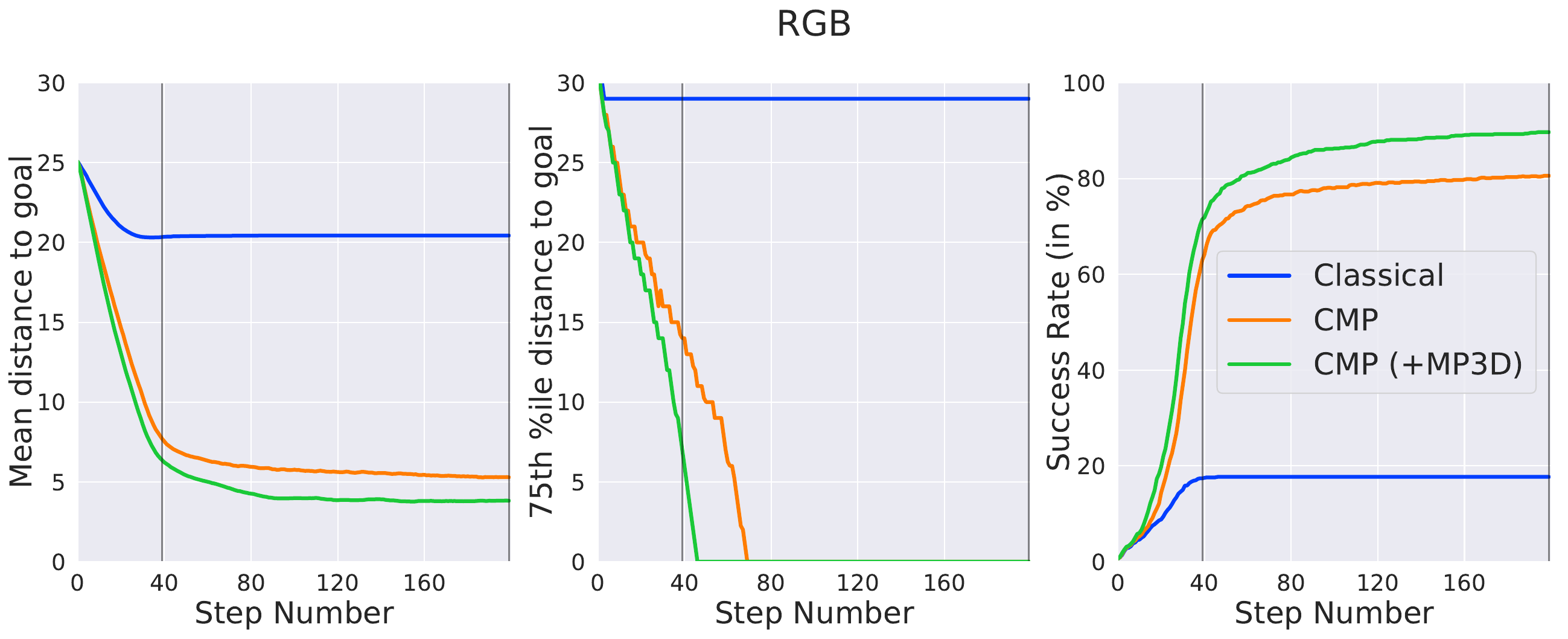}} \hfill \unskip\ \vrule\  \hfill 
{\insertW{0.490}{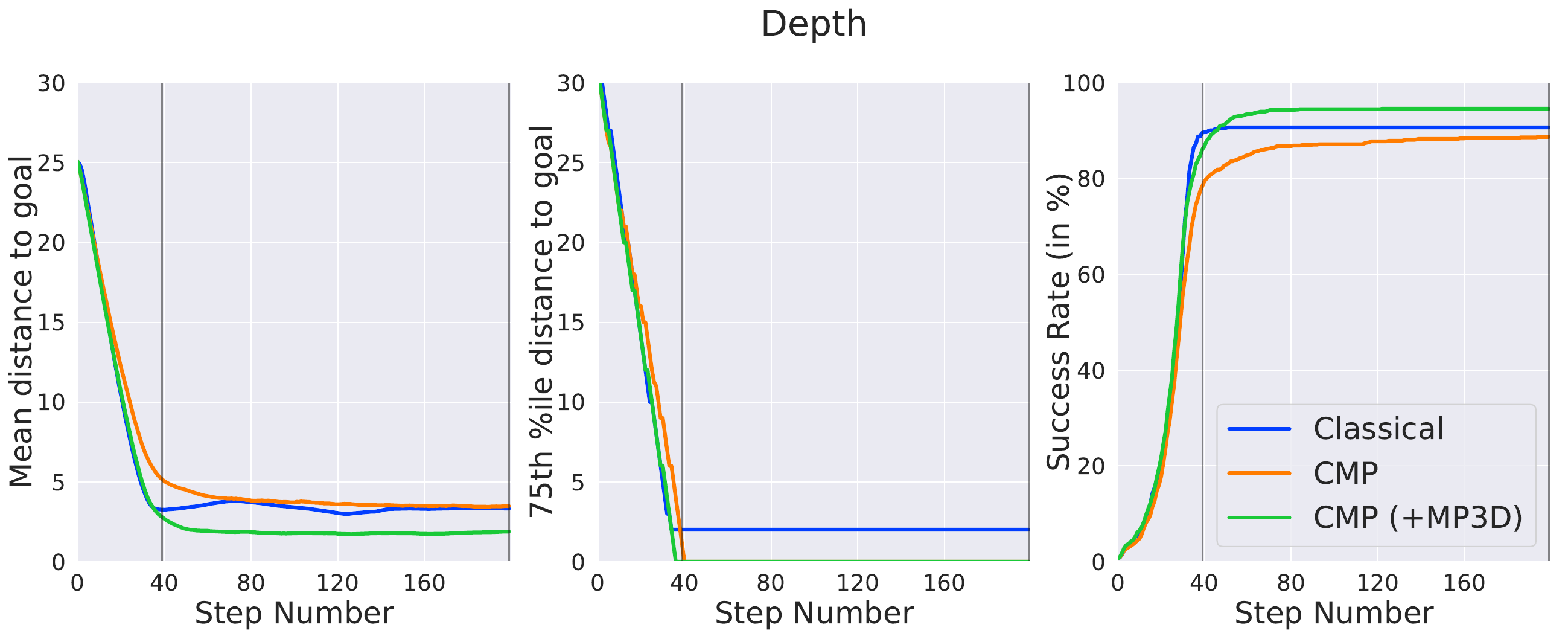}} 
\caption{\textbf{Geometric Task:} We plot the mean distance to goal,
75\textsuperscript{th} percentile distance to goal (lower is better) and
success rate (higher is better) as a function of the number of steps. Top row
compares the 4 frame reactive agent, LSTM based agent and our proposed CMP
based agent when using RGB images as input (left three plots) and when using
depth images as input (right three plots). Bottom row compares classical
mapping and planning with CMP (again, left is with \rgb input and right with
depth input). We note that CMP outperforms all these baselines, and using depth
input leads to better performance than using RGB input.}
\figlabel{geom-plot}
\end{figure*}

\input{semantic_spl}

\begin{figure*}
\centering
{\insertW{0.490}{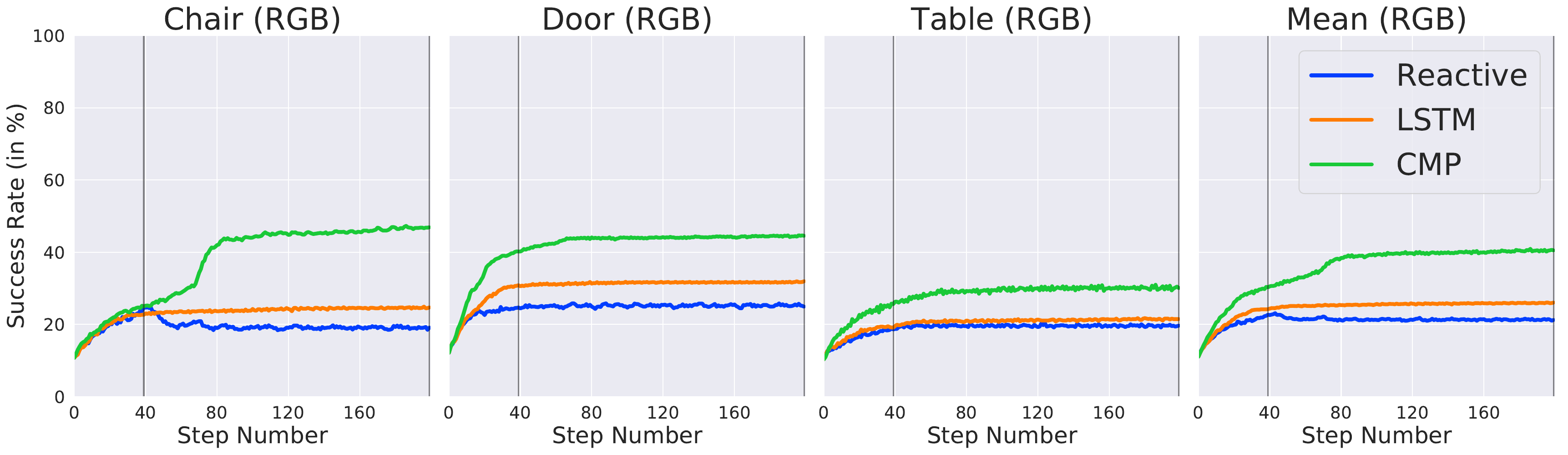}} \hfill \unskip\ \vrule\  \hfill 
{\insertW{0.490}{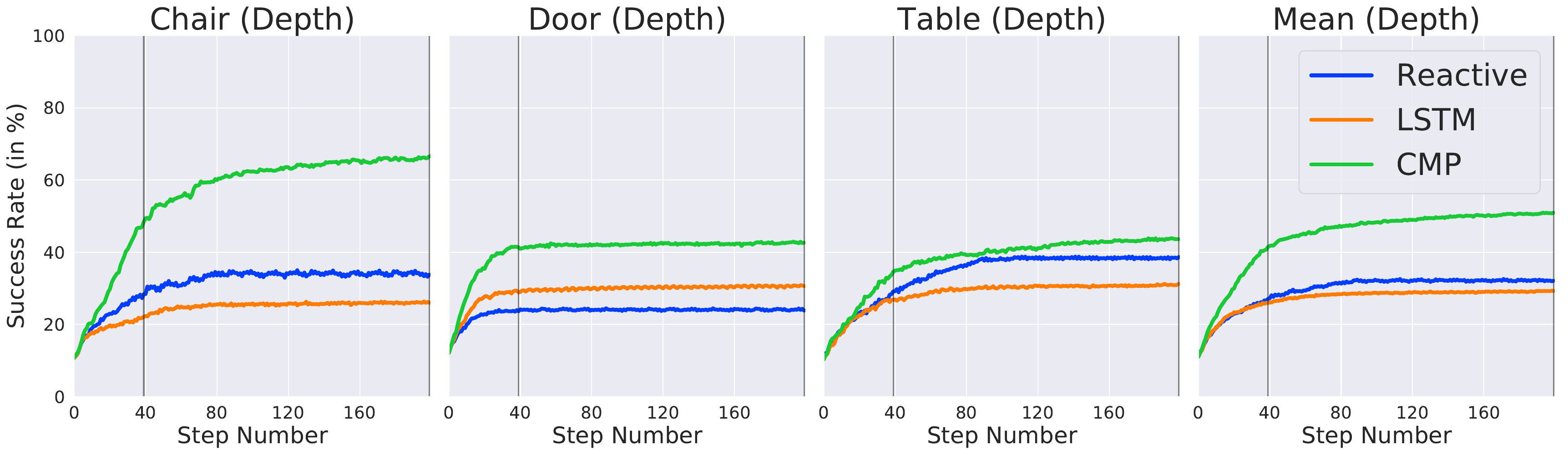}} \\ 
{\insertW{0.490}{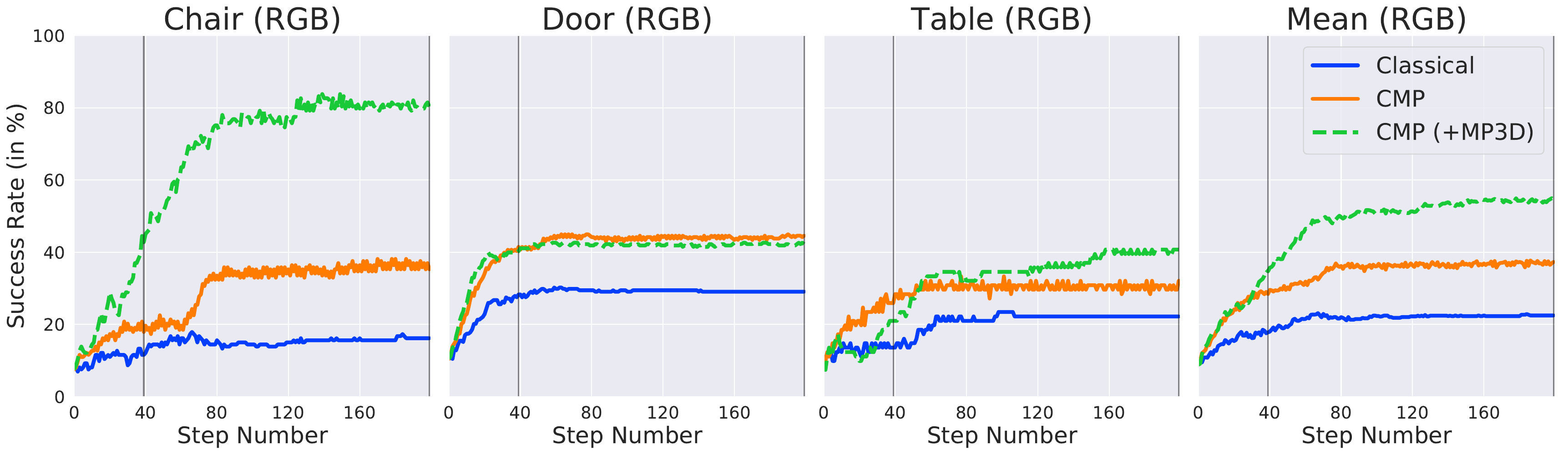}} \hfill \unskip\ \vrule\  \hfill 
{\insertW{0.490}{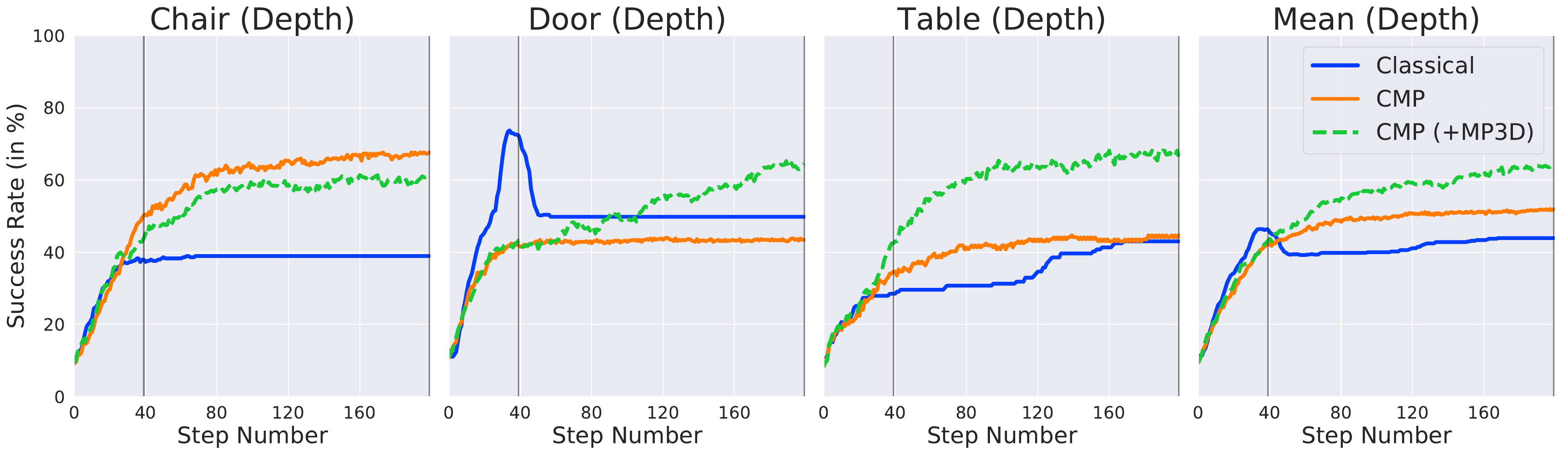}} 
\caption{\textbf{Semantic Task:} We plot the success rate as a function of the
number of steps for different categories. Top row compares learning based
approaches (4 frame reactive agent, LSTM based agent and our proposed CMP based
agent). Bottom row compares a classical approach (using exploration along with
semantic segmentation) and CMP. Left plots show performance when using \rgb
input, right plots show performance with depth input. See text for more
details.}
\figlabel{sem-plot}
\end{figure*}

\underline{\textbf{Geometric Task.}} We first present results for the geometric
task. \figref{geom-plot} plots the error metrics over time (for 199 time
steps), while \tableref{geom-result} reports these metrics at $39$ and $199$ time
steps, and SPL (with a max episode length of 199). We summarize the results
below:

\begin{enumerate}
\item We first note that nearest neighbor trajectory transfer does not work well,
with the mean and median distance to goal being $22$ and $25$ steps
respectively. This highlights the differences between the train and test
environments in our experiments.
\item Next, we note that the `No Image LSTM' baseline performs poorly as well, with a
success rate of $6.2\%$ only. This suggests that our testing episodes aren't
trivial. They don't just involve going straight to the goal, but require
understanding the layout of the given environment.
\item Next, we observe that the reactive baseline with a single frame also performs
poorly, succeeding only $8.2\%$ of the time. Note that this reactive baseline
is able to perform well on the training environments obtaining a mean distance
to goal of about 9 steps, but perform poorly on the test set only being able to
get to within 17 steps of the goal on average. This suggests that a reactive
agent is able to effectively memorize the environments it was trained on, but
fails to generalize to novel environments, this is not surprising given it does
not have any form of memory to allow it to map or plan. We also experimented
with using Drop Out in the fully connected layers for this model but found that
to hurt performance on both the train and the test sets.
\item Using additional frames as input to the reactive policy leads to a large
improvement in performance, and boosts performance to $20\%$, and to $56\%$
when using depth images. 
\item The LSTM based model is able to consistently outperform these reactive
baseline across all metrics. This indicates that memory does have a role to
play in navigation in novel environments.
\item Our proposed method CMP, outperforms all of these learning based
methods, across all metrics and input modalities. CMP achieves a lower
75\textsuperscript{th} \%ile distance to goal (14 and 1 as compared to 21 and 5
for the LSTM) and improves the success rate to 62.5\% and 78.3\% from 53.0\%
and 71.8\%. CMP also obtains higher SPL (59.4\% \vs 51.3\% and 73.7\% \vs
69.1\% for \rgb and depth input respectively).
\item We next compare to classical mapping and path planning. We first note
that a purely geometric approach when provided with depth images does really
really well, obtaining a SPL of 80.6\%. Access to depth images and perfect pose
allows efficient and accurate mapping, leading to high performance. In
contrast, when using only RGB images as input (but still with perfect pose),
performance drops sharply to only 15.9\%. There are two failure modes: spurious
stray points in reconstruction that get treated as obstacles, and failure to
reconstruct texture-less obstacles (such as walls) and bumping into them.  In
comparison, CMP performs well even when presented with just \rgb images, at
59.6\% SPL. Furthermore, when CMP is trained with more data (6 additional large
buildings from the Matterport3D dataset \cite{Matterport3D}), performance
improves further, to 70.8\% SPL for \rgb input and 82.3\% SPL for depth input.
Though we tried our best at implementing the classical purely geometry-based
method, we note that they may be improved further by introducing and validating
over more and more hyper-parameters, specially for the case where depth
observations are available as input. \end{enumerate}

\pparagraph{Variance Over Multiple Runs.}
We also report variance in performance over five re-trainings from different
random initializations of the network for the 3 most competitive methods
(Reactive with 4 frames, LSTM, and CMP) for the depth image case.
\figref{all_error} shows the performance, the solid line shows the
median metric value and the surrounding shaded region represents the minimum
and maximum metric value over the five re-trainings. As we are using imitation
learning (instead of reinforcement learning) for training our models, variation
in performance is reasonably small for all models and CMP leads to significant
improvements.

\pparagraph{Ablations.} We also studied ablated versions of our proposed
method. We summarize the key takeaways, a learned mapper leads to better
navigation performance than an analytic mapper, planning is crucial (specially
for when using RGB images as input) and single-scale planning works slightly
better than the multi-scale planning at the cost of increased planning cost.
More details in \secrefext{supp_exp}.

\setlength{\tabcolsep}{1.8pt}
\begin{figure*}
\centering
\begin{tabular}{ccccc|cc}
\insertW{0.1900}{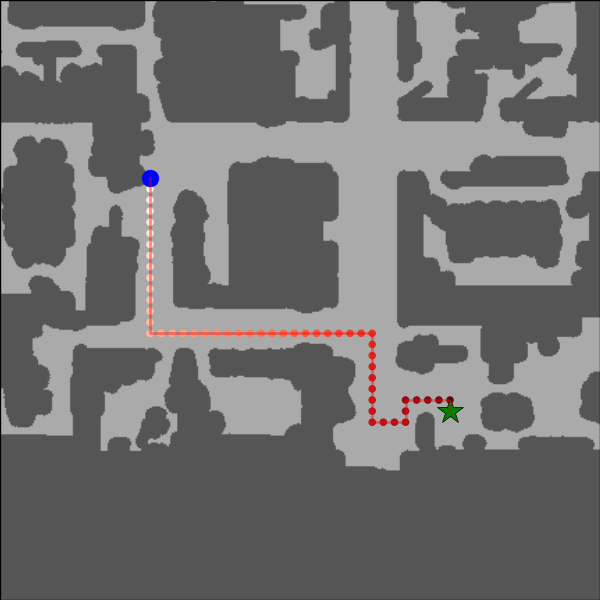} &
\insertW{0.1900}{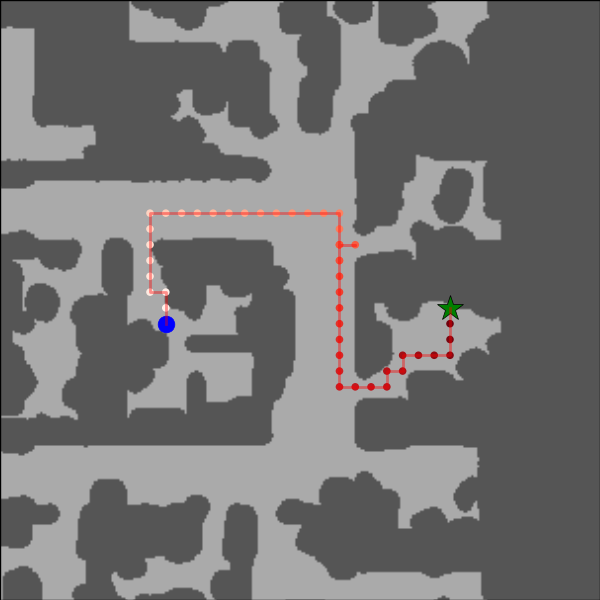} &
\insertW{0.1900}{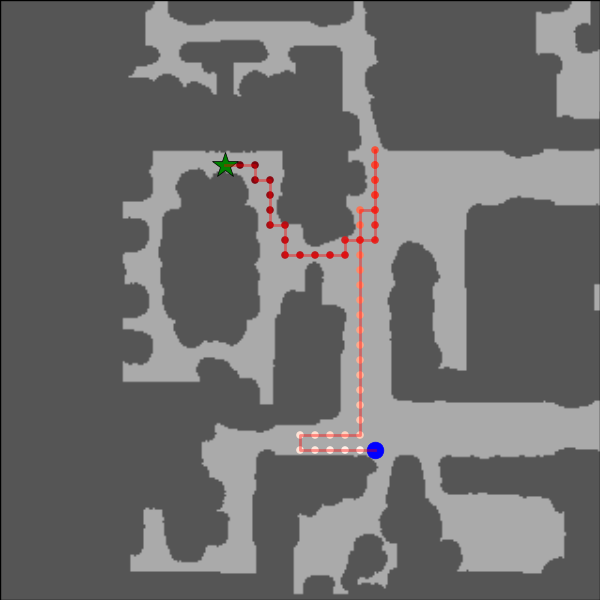} &
\insertW{0.1900}{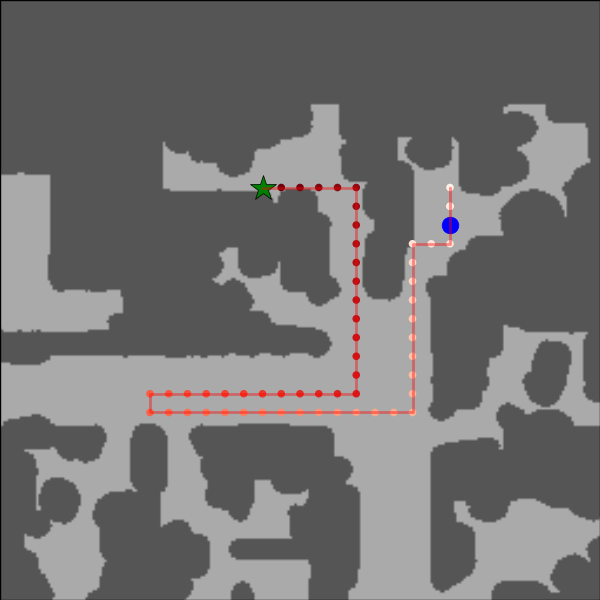} & & &%\unskip\ \vrule\
\insertW{0.1900}{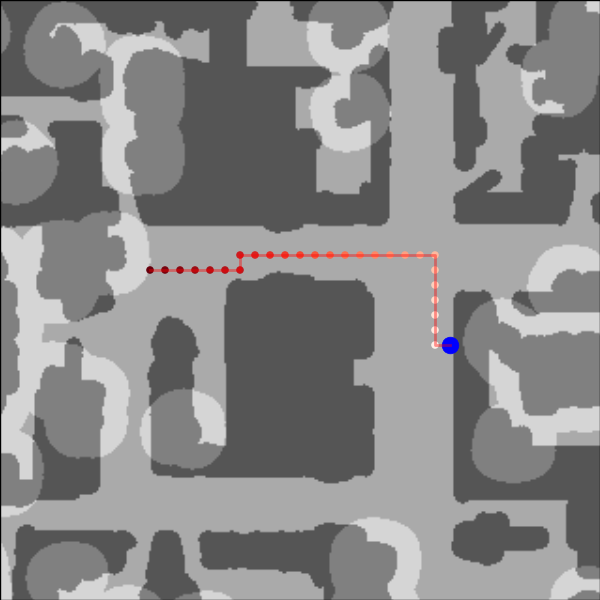} \\ 
\insertW{0.1900}{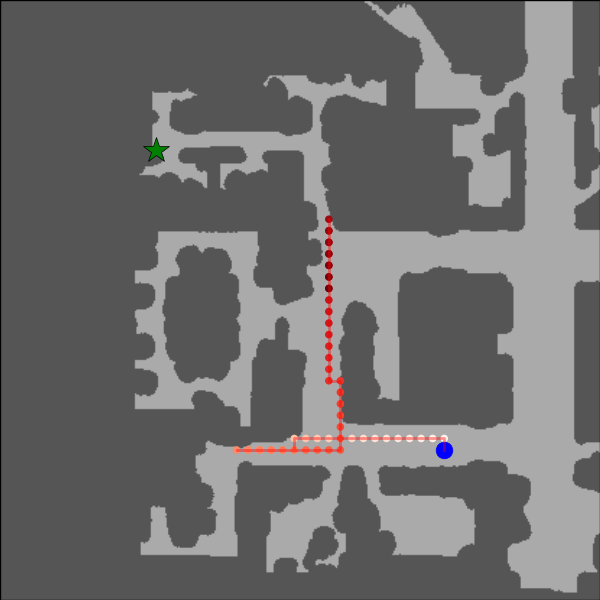} & 
\insertW{0.1900}{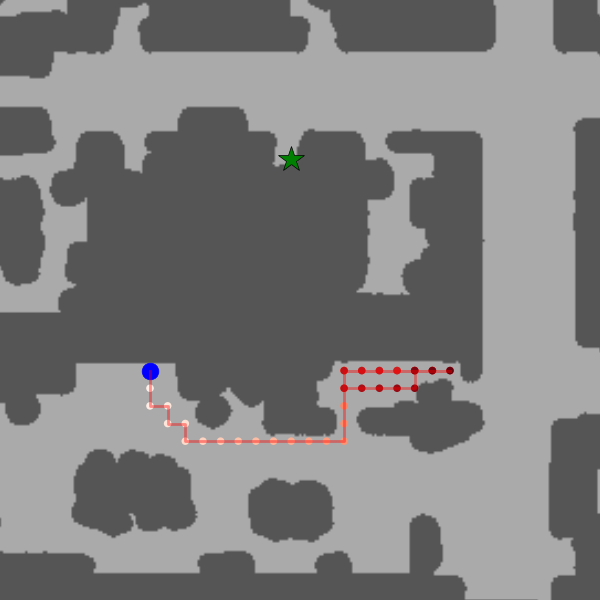} &
\insertW{0.1900}{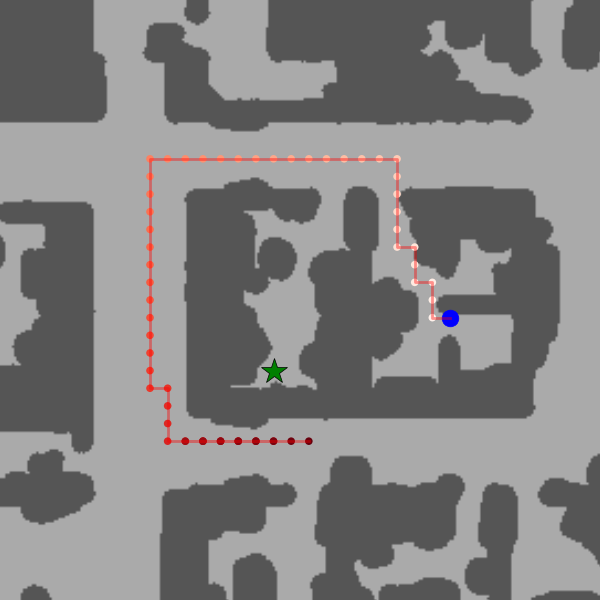}  &
\insertW{0.1900}{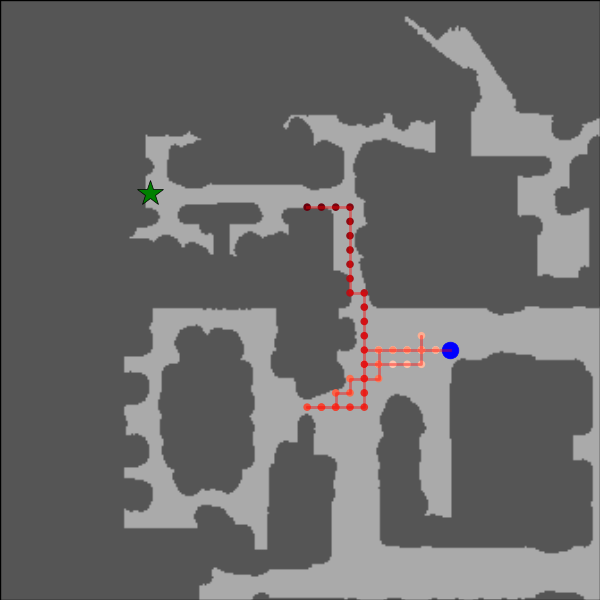} & & &%\unskip\ \vrule\ 
\insertW{0.1900}{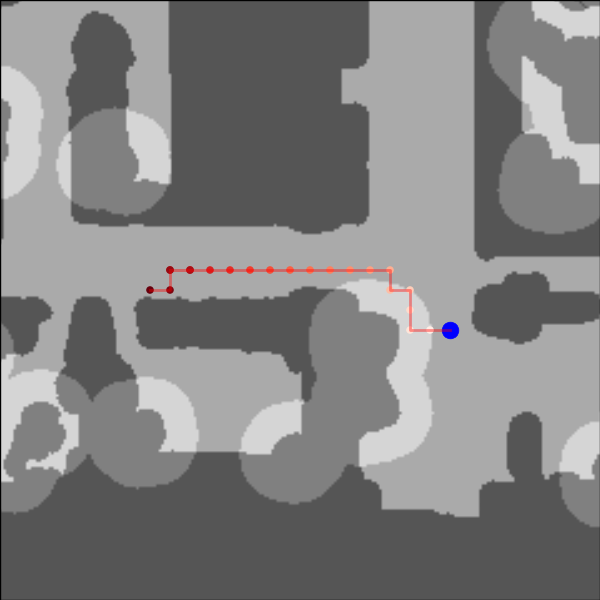} 
\end{tabular}
\caption{\textbf{Representative Success and Failure Cases for CMP}: We
visualize trajectories for some typical success and failure cases for CMP. Dark
gray regions show occupied space, light gray regions show free space. The agent
starts from the blue dot and is required to reach the green star (or semantic
regions shown in light gray). The agent's trajectory is shown by the dotted red
line. While we visualize the trajectories in the top view, note that the agent
only receives the first person view as input. \textbf{Top plots} show success
cases for geometric task. We see that the agent is able to traverse large
distances across multiple rooms to get to the target location, go around
obstacles and quickly resolve that it needs to head to the next room and not
the current room. The last two plots show cases where the agent successfully
backtracks. \textbf{Bottom plots} show failure cases for geometric task:
problems with navigating around tight spaces (entering through a partially
opened door, and getting stuck in the corner (the gap is not big enough to pass
through)), missing openings which would have lead to shorter paths, thrashing
around in space without making progress. \textbf{Right plots} visualize
trajectories for `go to the chair' semantic task. The top figure shows a
success case, while the bottom figure shows a typical failure case where the
agent walks right through a chair region.} \figlabel{trajectory-vis}
\end{figure*}

\pparagraph{Additional comparisons between LSTM and CMP.} We also conducted
additional experiments to further compare the performance of the LSTM baseline
with our model in the most competitive scenario where both methods use depth
images. We summarize the key conclusions here and provide more details in
\secrefext{supp_exp}. We report performance when the target is
much further away (64 time steps away) in \tableref{scenarios}~(top). We see a larger gap in performance
between LSTM and CMP for this test scenarios. We also compared performance of
CMP and LSTM over problems of different difficulty and observed that CMP is
generally better across all values of hardness, but for RGB images it is
particularly better for cases with high hardness (\figref{hardness}). We also evaluate how well
these models generalize when trained on a single scene, and when transferring
across datasets. We find that there is a smaller drop in performance for CMP as
compared to LSTM (\tableref{scenarios}~(bottom)). More details in
\secrefext{supp_exp}. \figref{trajectory-vis} visualizes and discusses some
representative success and failure cases for CMP, video examples are available
on the project website. 

\underline{\textbf{Semantic Task.}} We next present results for the semantic
task, where the goal is to find object of interest. The agent receives a
one-hot vector indicating the object category it must go to and is considered
successful if it can reach any instance of the indicated object category.  We
compare our method to the best performing reactive and LSTM based baseline
models from the geometric navigation task\footnote{This LSTM is impoverished
because it no longer receives the egomotion of the agent as input (because the
goal can not be specified as an offset relative to the robot). We did
experiment with a LSTM model which received egomotion as input but weren't able
to train it in initial experiments.}. This is a challenging task specially
because the agent may start in a location from which the desired object is not
visible, and it must learn to explore the environment to find the desired
object. \figref{sem-plot} and \tableref{sem-result} reports the success rate
and the SPL metric for the different categories we study. 
\figref{trajectory-vis} shows sample trajectories for this task for CMP.  We
summarize our findings below:
\begin{enumerate}
\item This is a hard task, performance for all methods is much lower than for
the geometric task of reaching a specified point in space.
\item CMP performs better than the other two learning based baselines across all
metrics.
\item Comparisons to the classical baseline of geometric exploration followed
by use of semantic segmentation (\figref{sem-plot}~(bottom) orange \vs blue
line) are also largely favorable to CMP. Performance for classical baseline
with \rgb input suffers due to inaccuracy in estimating the occupancy of the
environment. With depth input, this becomes substantially easier, leading to
better performance. A particularly interesting case is that of finding doors.
As the classical baseline explores, it comes close to doors as it exits the
room it started from. However, it is unable to stop (possibly being unable to
reliably detect them). This explains the spike in performance in
\figref{sem-plot}.
\item We also report SPL for this task for the different methods in
\tableref{sem-result}. We observe that though the success rates are high, SPL
numbers are low. In comparison to success rate, SPL additionally measures path
efficiency and whether the agent is able to reliably determine that it has
reached the desired target. \figref{sem-plot}~(bottom) shows that the success
rate continues to improve over the length of the episodes, implying that the
agent does realize that it has reached the desired object of interest. Thus, we
believe SPL numbers are low because of inefficiency in reaching the target,
specially as SPL measures efficiency with respect to the closest object of
interest using \textit{full} environment information. This can be particularly
strict in novel environments where the agent may not have any desired objects
in view, and thus needs to explore the environment to be able to find them.
Nevertheless, CMP outperforms learning-based methods on this metric, and also
outperforms our classical baseline when using RGB input. 
\item CMP when trained with additional data (6 additional buildings from the
Matterport3D dataset \cite{Matterport3D}) performs much better (green \vs
orange lines in \figref{sem-plot}~(bottom)), indicating scope for further
improvements in such polcies as larger datasets become available. Semantic
segmentation networks for the classical baseline can similarly be improved
using more data (possibly also from large-scale Internet datasets), but we
leave those experiments and comparisons for future work.
\end{enumerate}

%% file: geom_spl.tex
\input{nav_results_numbers}
\input{geo_spl_numbers}

\renewcommand{\arraystretch}{1.2} 
\setlength{\tabcolsep}{4pt}
\begin{table*}
\centering
\footnotesize
\resizebox{1.0\linewidth}{!}{
\begin{tabular}{lcccccccccggggggggcggg}\toprule 
  \multirow{2}{*}{$\quad$ Method} & 
  \multicolumn{2}{c}{Mean} & & \multicolumn{2}{c}{75\textsuperscript{th} \%ile} & & \multicolumn{2}{c}{Success \%age} & & 
  \multicolumn{2}{g}{Mean} & & \multicolumn{2}{g}{75\textsuperscript{th} \%ile} & & \multicolumn{2}{g}{Success \%age} & & 
  \multicolumn{2}{g}{SPL \%age}
  \\ \cmidrule(l{4pt}r{4pt}){2-3} \cmidrule(l{4pt}r{4pt}){5-6} \cmidrule(l{4pt}r{4pt}){8-9} \cmidrule(l{4pt}r{4pt}){11-12} \cmidrule(l{4pt}r{4pt}){14-15} \cmidrule(l{4pt}r{4pt}){17-18} \cmidrule(l{4pt}r{4pt}){20-21}
  & \rgb & Depth & & \rgb & Depth &  & \rgb & Depth &  & \rgb & Depth & & \rgb & Depth &  & \rgb & Depth  &  & \rgb  & Depth 
  \\ \midrule
  \textbf{Geometric Task} (4000 episodes) & & & & & & & & & & & & & & & & & & & & \\
  $\quad$ Initial       & \Ainit  & \Ainit  &  & \Binit  & \Binit  &  & \Cinit  & \Cinit  &  &  &  &  &  &  &  &  &  &  &  & \\
  $\quad$ No Image LSTM & \Ablind & \Ablind &  & \Bblind & \Bblind &  & \Cblind & \Cblind &  &  &  &  &  &  &  &  &  &  &  & \\
  $\quad$ Reactive (1 frame)  & \AreactiveAddRgbNoh & \AreactiveMulDNoh &  & \BreactiveAddRgbNoh & \BreactiveMulDNoh &  & \CreactiveAddRgbNoh & \CreactiveMulDNoh &  &  &  & & & & & & & & &\\
  $\quad$ Reactive (4 frames)    & \MAreactRGB    & \MAreactD    &  & \MCreactRGB    & \MCreactD     &  & \MEreactRGB    & \MEreactD    &
                                 & \MBreactRGB    & \MBreactD    &  & \MDreactRGB    & \MDreactD     &  & \MFreactRGB    & \MFreactD    &  & \MGreactRGB    & \MGreactD \\
  $\quad$ LSTM                   & \MAlstmRGB     & \MAlstmD     &  & \MClstmRGB     & \MClstmD      &  & \MElstmRGB     & \MElstmD     &
                                 & \MBlstmRGB     & \MBlstmD     &  & \MDlstmRGB     & \tb{\MDlstmD} &  & \MFlstmRGB     & \MFlstmD     &  & \MGlstmRGB     & \MGlstmD \\
  $\quad$ Our (CMP)              & \tb{\MAcmpRGB} & \tb{\MAcmpD} &  & \tb{\MCcmpRGB} & \tb{\MCcmpD}  &  & \tb{\MEcmpRGB} & \tb{\MEcmpD} &
                                 & \tb{\MBcmpRGB} & \tb{\MBcmpD} &  & \tb{\MDcmpRGB} & \tb{\MDcmpD}  &  & \tb{\MFcmpRGB} & \tb{\MFcmpD} &  & \tb{\MGcmpRGB} & \tb{\MGcmpD} \\
  \midrule
  \textbf{Geometric Task} (1000 episodes) & & & & & & & & & & & & & & & & & & & & \\
  $\quad$ Classical (900px, $4\times$ images)                & {\MAclassicalRGB} & {\MAclassicalD} &  & {\MCclassicalRGB} & {\MCclassicalD} &  & {\MEclassicalRGB} & \tb{\MEclassicalD} &
                                   & {\MBclassicalRGB} & {\MBclassicalD} &  & {\MDclassicalRGB} & {\MDclassicalD} &  & {\MFclassicalRGB} & {\MFclassicalD} &  & {\MGclassicalRGB} & {\MGclassicalD} \\
  $\quad$ Our (CMP)                & {\MAcmpsbpdRGB}   & {\MAcmpsbpdD}   &  & {\MCcmpsbpdRGB}   & {\MCcmpsbpdD}   &  & {\MEcmpsbpdRGB}   & {\MEcmpsbpdD}   &
                                   & {\MBcmpsbpdRGB}   & {\MBcmpsbpdD}   &  & \tb{\MDcmpsbpdRGB}   & \tb{\MDcmpsbpdD}   &  & {\MFcmpsbpdRGB}   & {\MFcmpsbpdD}   &  & {\MGcmpsbpdRGB}   & {\MGcmpsbpdD} \\
  $\quad$ Our (CMP [+6 MP3D Envs])    & \tb{\MAmpdRGB}    & \tb{\MAmpdD}    &  & \tb{\MCmpdRGB}    & \tb{\MCmpdD}    &  & \tb{\MEmpdRGB}    & {\MEmpdD}    &
                                   & \tb{\MBmpdRGB}    & \tb{\MBmpdD}    &  & \tb{\MDmpdRGB}    & \tb{\MDmpdD}    &  & \tb{\MFmpdRGB}    & \tb{\MFmpdD}    &  & \tb{\MGmpdRGB}    & \tb{\MGmpdD} \\
  \bottomrule
  \end{tabular}}
\caption{\textbf{Results for the Geometric Task:} 
We report the mean distance to goal location, 75\textsuperscript{th} percentile
distance to goal, success rate and SPL for various methods for the geometric
task. Unshaded table reports metrics at time step 39, while the shaded table
reports metrics for selected competitive methods at time step 199. Top part of
the table reports comparisons between learning-based methods, while the bottom
part report comparisons to classical purely geometry-based methods.}
\tablelabel{geom-result}
\end{table*}

%% file: nav_results_numbers.tex
\newcommand{\AreactiveMulRgbNoh}{21.0}
\newcommand{\BreactiveMulRgbNoh}{28}
\newcommand{\CreactiveMulRgbNoh}{\phz5.9}
\newcommand{\AreactiveMulDNoh}{17.0}
\newcommand{\BreactiveMulDNoh}{26}
\newcommand{\CreactiveMulDNoh}{21.9}
\newcommand{\AreactiveMulRgbH}{14.4}
\newcommand{\BreactiveMulRgbH}{25}
\newcommand{\CreactiveMulRgbH}{31.4}
\newcommand{\AreactiveMulDH}{\phz9.4}
\newcommand{\BreactiveMulDH}{19}
\newcommand{\CreactiveMulDH}{54.8}
\newcommand{\AreactiveAddRgbNoh}{20.9}
\newcommand{\BreactiveAddRgbNoh}{28}
\newcommand{\CreactiveAddRgbNoh}{\phz8.2}
\newcommand{\AreactiveAddDNoh}{17.5}
\newcommand{\BreactiveAddDNoh}{26}
\newcommand{\CreactiveAddDNoh}{19.7}
\newcommand{\AreactiveAddRgbH}{15.2}
\newcommand{\BreactiveAddRgbH}{25}
\newcommand{\CreactiveAddRgbH}{30.4}
\newcommand{\AreactiveAddDH}{\phz8.8}
\newcommand{\BreactiveAddDH}{18}
\newcommand{\CreactiveAddDH}{56.9}
\newcommand{\AcmpRgb}{\phz7.7}
\newcommand{\BcmpRgb}{14}
\newcommand{\CcmpRgb}{62.5}
\newcommand{\AcmpD}{\phz4.8}
\newcommand{\BcmpD}{\phz1}
\newcommand{\CcmpD}{78.3}
\newcommand{\AcmpNomsRgb}{\phz7.9}
\newcommand{\BcmpNomsRgb}{12}
\newcommand{\CcmpNomsRgb}{63.0}
\newcommand{\AcmpNomsD}{\phz4.9}
\newcommand{\BcmpNomsD}{\phz1}
\newcommand{\CcmpNomsD}{79.5}
\newcommand{\AcmpNoVinRgb}{\phz8.5}
\newcommand{\BcmpNoVinRgb}{16}
\newcommand{\CcmpNoVinRgb}{58.6}
\newcommand{\AcmpNoVinD}{\phz4.8}
\newcommand{\BcmpNoVinD}{\phz1}
\newcommand{\CcmpNoVinD}{79.0}
\newcommand{\AcmpAm}{\phz8.0}
\newcommand{\BcmpAm}{14}
\newcommand{\CcmpAm}{62.9}
\newcommand{\Ablind}{20.8}
\newcommand{\Bblind}{28}
\newcommand{\Cblind}{\phz6.2}
\newcommand{\Ainit}{25.3}
\newcommand{\Binit}{30}
\newcommand{\Cinit}{\phz0.7}
\newcommand{\AtwolstmD}{\phz3.9}
\newcommand{\BtwolstmD}{\phz0}
\newcommand{\CtwolstmD}{82.2}
\newcommand{\AtwocmpD}{\phz4.0}
\newcommand{\BtwocmpD}{\phz0}
\newcommand{\CtwocmpD}{81.6}
\newcommand{\Atwoinit}{25.2}
\newcommand{\Btwoinit}{30}
\newcommand{\Ctwoinit}{\phz0.5}
\newcommand{\AlonglstmD}{15.2}
\newcommand{\BlonglstmD}{29}
\newcommand{\ClonglstmD}{58.4}
\newcommand{\AlongcmpD}{11.9}
\newcommand{\BlongcmpD}{19.2}
\newcommand{\ClongcmpD}{66.3}
\newcommand{\Alonginit}{47.2}
\newcommand{\Blonginit}{58}
\newcommand{\Clonginit}{\phz0.0}
\newcommand{\AnoalllstmD}{\phz8.9}
\newcommand{\BnoalllstmD}{18}
\newcommand{\CnoalllstmD}{58.9}
\newcommand{\AnoallcmpD}{\phz7.0}
\newcommand{\BnoallcmpD}{10}
\newcommand{\CnoallcmpD}{67.9}
\newcommand{\Anoallinit}{25.3}
\newcommand{\Bnoallinit}{30}
\newcommand{\Cnoallinit}{\phz0.7}
\newcommand{\AtrlstmD}{11.0}
\newcommand{\BtrlstmD}{21}
\newcommand{\CtrlstmD}{48.6}
\newcommand{\AtrcmpD}{\phz8.5}
\newcommand{\BtrcmpD}{15}
\newcommand{\CtrcmpD}{61.1}
\newcommand{\Atrinit}{25.3}
\newcommand{\Btrinit}{30}
\newcommand{\Ctrinit}{\phz0.7}
\newcommand{\AlonglonglstmD}{12.5}
\newcommand{\BlonglonglstmD}{19}
\newcommand{\ClonglonglstmD}{69.0}
\newcommand{\AlonglongcmpD}{\phz9.3}
\newcommand{\BlonglongcmpD}{\phz0}
\newcommand{\ClonglongcmpD}{78.5}
\newcommand{\Alonglonginit}{47.2}
\newcommand{\Blonglonginit}{58}
\newcommand{\Clonglonginit}{\phz0.0}

%% file: geo_spl_numbers.tex
\newcommand{\MAreactRGB}{17.0}
\newcommand{\MBreactRGB}{16.7}
\newcommand{\MCreactRGB}{26}
\newcommand{\MDreactRGB}{26}
\newcommand{\MEreactRGB}{20.3}
\newcommand{\MFreactRGB}{22.8}
\newcommand{\MGreactRGB}{17.4}
\newcommand{\MAlstmRGB}{10.3}
\newcommand{\MBlstmRGB}{\phz8.1}
\newcommand{\MClstmRGB}{21}
\newcommand{\MDlstmRGB}{18}
\newcommand{\MElstmRGB}{52.1}
\newcommand{\MFlstmRGB}{69.5}
\newcommand{\MGlstmRGB}{51.3}
\newcommand{\MAcmpRGB}{\phz7.7}
\newcommand{\MBcmpRGB}{\phz5.4}
\newcommand{\MCcmpRGB}{14}
\newcommand{\MDcmpRGB}{\phz0}
\newcommand{\MEcmpRGB}{63.0}
\newcommand{\MFcmpRGB}{80.0}
\newcommand{\MGcmpRGB}{59.4}

\newcommand{\MAreactD}{\phz8.9}
\newcommand{\MBreactD}{\phz8.2}
\newcommand{\MCreactD}{18}
\newcommand{\MDreactD}{17}
\newcommand{\MEreactD}{56.1}
\newcommand{\MFreactD}{62.2}
\newcommand{\MGreactD}{52.0}
\newcommand{\MAlstmD}{\phz5.9}
\newcommand{\MBlstmD}{\phz3.5}
\newcommand{\MClstmD}{\phz6}
\newcommand{\MDlstmD}{\phz0}
\newcommand{\MElstmD}{71.3}
\newcommand{\MFlstmD}{88.5}
\newcommand{\MGlstmD}{69.1}
\newcommand{\MAcmpD}{\phz4.8}
\newcommand{\MBcmpD}{\phz3.3}
\newcommand{\MCcmpD}{\phz1}
\newcommand{\MDcmpD}{\phz0}
\newcommand{\MEcmpD}{78.8}
\newcommand{\MFcmpD}{89.3}
\newcommand{\MGcmpD}{73.7}

\newcommand{\MAclassicalRGB}{20.3}
\newcommand{\MBclassicalRGB}{20.4}
\newcommand{\MCclassicalRGB}{29}
\newcommand{\MDclassicalRGB}{29}
\newcommand{\MEclassicalRGB}{17.4}
\newcommand{\MFclassicalRGB}{17.7}
\newcommand{\MGclassicalRGB}{15.9}
\newcommand{\MAcmpsbpdRGB}{\phz7.7}
\newcommand{\MBcmpsbpdRGB}{\phz5.3}
\newcommand{\MCcmpsbpdRGB}{14}
\newcommand{\MDcmpsbpdRGB}{\phz0}
\newcommand{\MEcmpsbpdRGB}{63.0}
\newcommand{\MFcmpsbpdRGB}{80.6}
\newcommand{\MGcmpsbpdRGB}{59.6}
\newcommand{\MAmpdRGB}{\phz6.3}
\newcommand{\MBmpdRGB}{\phz3.8}
\newcommand{\MCmpdRGB}{\phz7}
\newcommand{\MDmpdRGB}{\phz0}
\newcommand{\MEmpdRGB}{71.5}
\newcommand{\MFmpdRGB}{89.7}
\newcommand{\MGmpdRGB}{70.8}

\newcommand{\MAclassicalD}{\phz3.3}
\newcommand{\MBclassicalD}{\phz3.3}
\newcommand{\MCclassicalD}{\phz2}
\newcommand{\MDclassicalD}{\phz2}
\newcommand{\MEclassicalD}{89.6}
\newcommand{\MFclassicalD}{90.7}
\newcommand{\MGclassicalD}{80.6}
\newcommand{\MAcmpsbpdD}{\phz5.2}
\newcommand{\MBcmpsbpdD}{\phz3.5}
\newcommand{\MCcmpsbpdD}{\phz1}
\newcommand{\MDcmpsbpdD}{\phz0}
\newcommand{\MEcmpsbpdD}{78.4}
\newcommand{\MFcmpsbpdD}{88.7}
\newcommand{\MGcmpsbpdD}{73.1}
\newcommand{\MAmpdD}{\phz2.8}
\newcommand{\MBmpdD}{\phz1.9}
\newcommand{\MCmpdD}{\phz0}
\newcommand{\MDmpdD}{\phz0}
\newcommand{\MEmpdD}{86.1}
\newcommand{\MFmpdD}{94.6}
\newcommand{\MGmpdD}{82.3}

%% file: semantic_spl.tex
\input{sem_spl_numbers.tex}

\newcommand{\uu}[1]{\textit{#1}} 
\renewcommand{\arraystretch}{1.2} 
\setlength{\tabcolsep}{8pt}
\begin{table*}
\centering
\footnotesize
\resizebox{1.0\linewidth}{!}{
\begin{tabular}{lccc>{\color{blue}}ccggg>{\color{blue}}gcggg>{\color{blue}}ggg}
  \toprule 
  \multirow{2}{*}{$\quad$ Method} & \multicolumn{4}{c}{Success \%age [39 steps]} & & \multicolumn{4}{g}{Success \%age [199 steps]} & & \multicolumn{4}{g}{SPL \%age [199 steps]} \\ 
  \cmidrule(l{0pt}r{0pt}){2-5} \cmidrule(l{0pt}r{0pt}){7-10} \cmidrule(l{0pt}r{0pt}){12-15} 
                                           & Chair               & Door                & Table               & Mean                & & Chair               & Door                & Table               & Mean                & & Chair               & Door                & Table               & Mean \\ \midrule
  \textbf{\rgb} (4000 episodes)            &                     &                     &                     &                     & &                     &                     &                     &                     & &                     &                     &                     & \\
  $\quad$ Initial                          & 10.7                & 12.1                & 10.3                & 11.1                & & 10.7                & 12.1                & 10.3                & 11.1                & &                     &                     &                     & \\
  $\quad$ Reactive (4 frames)              & \STAreactRGB        & \STBreactRGB        & \STCreactRGB        & \STDreactRGB        & & \STEreactRGB        & \STFreactRGB        & \STGreactRGB        & \STHreactRGB        & & \tb{\STMreactRGB}   & \STNreactRGB        & \tb{\STOreactRGB}   & \STPreactRGB \\
  $\quad$ LSTM                             & \STAlstmRGB         & \STBlstmRGB         & \STClstmRGB         & \STDlstmRGB         & & \STElstmRGB         & \STFlstmRGB         & \STGlstmRGB         & \STHlstmRGB         & & \STMlstmRGB         & \STNlstmRGB         & \STOlstmRGB         & \STPlstmRGB \\
  $\quad$ Our (CMP)                        & \tb{\STAcmpRGB}     & \tb{\STBcmpRGB}     & \tb{\STCcmpRGB}     & \tb{\STDcmpRGB}     & & \tb{\STEcmpRGB}     & \tb{\STFcmpRGB}     & \tb{\STGcmpRGB}     & \tb{\STHcmpRGB}     & & \STMcmpRGB          & \tb{\STNcmpRGB}     & {\STOcmpRGB}        & \tb{\STPcmpRGB} \\
  \textbf{\rgb} (500 epsiodes)             &                     &                     &                     &                     & &                     &                     &                     &                     & &                     &                     &                     & \\
  $\quad$ Classical (Explore + Sem. Segm.) & \STAclassicalRGB    & \STBclassicalRGB    & \STCclassicalRGB    & \STDclassicalRGB    & & \STEclassicalRGB    & \STFclassicalRGB    & \STGclassicalRGB    & \STHclassicalRGB    & & {\STMclassicalRGB}  & \STNclassicalRGB    & \STOclassicalRGB    & \STPclassicalRGB \\
  $\quad$ Our (CMP)                        & \tb{\STAcmpsbpdRGB} & \tb{\STBcmpsbpdRGB} & \tb{\STCcmpsbpdRGB} & \tb{\STDcmpsbpdRGB} & & \tb{\STEcmpsbpdRGB} & \tb{\STFcmpsbpdRGB} & \tb{\STGcmpsbpdRGB} & \tb{\STHcmpsbpdRGB} & & \tb{\STMcmpsbpdRGB} & \tb{\STNcmpsbpdRGB} & \tb{\STOcmpsbpdRGB} & \tb{\STPcmpsbpdRGB} \\
  \cmidrule(l{16pt}r{0pt}){1-15}
  $\quad$ \uu{Our (CMP [+ 6 MP3D Envs])}   & \uu{\STAcmpmpdRGB}  & \uu{\STBcmpmpdRGB}  & \uu{\STCcmpmpdRGB}  & \uu{\STDcmpmpdRGB}  & & \uu{\STEcmpmpdRGB}  & \uu{\STFcmpmpdRGB}  & \uu{\STGcmpmpdRGB}  & \uu{\STHcmpmpdRGB}  & & \uu{\STMcmpmpdRGB}  & \uu{\STNcmpmpdRGB}  & \uu{\STOcmpmpdRGB}  & \uu{\STPcmpmpdRGB} \\
  \midrule
  \textbf{Depth} (4000 episodes)           &                     &                     &                     &                     & &                     &                     &                     &                     & &                     &                     &                     & \\
  $\quad$ Reactive (4 frames)              & \STAreactD          & \STBreactD          & \STCreactD          & \STDreactD          & & \STEreactD          & \STFreactD          & \STGreactD          & \STHreactD          & & \STMreactD          & \STNreactD          & \STOreactD          & \STPreactD \\
  $\quad$ LSTM                             & \STAlstmD           & \STBlstmD           & \STClstmD           & \STDlstmD           & & \STElstmD           & \STFlstmD           & \STGlstmD           & \STHlstmD           & & \STMlstmD           & \STNlstmD           & \STOlstmD           & \STPlstmD \\
  $\quad$ Our (CMP)                        & \tb{\STAcmpD}       & \tb{\STBcmpD}       & \tb{\STCcmpD}       & \tb{\STDcmpD}       & & \tb{\STEcmpD}       & \tb{\STFcmpD}       & \tb{\STGcmpD}       & \tb{\STHcmpD}       & & \tb{\STMcmpD}       & \tb{\STNcmpD}       & \tb{\STOcmpD}       & \tb{\STPcmpD} \\
  \textbf{Depth} (1000 epsiodes)           &                     &                     &                     &                     & &                     &                     &                     &                     & &                     &                     &                     & \\
  $\quad$ Classical (Explore + Sem. Segm.) & \STAclassicalD      & \tb{\STBclassicalD} & \STCclassicalD      & \tb{\STDclassicalD} & & \STEclassicalD      & \tb{\STFclassicalD} & \STGclassicalD      & \STHclassicalD      & & \tb{\STMclassicalD} & \tb{\STNclassicalD} & \tb{\STOclassicalD} & \tb{\STPclassicalD} \\
  $\quad$ Our (CMP)                        & \tb{\STAcmpsbpdD}   & {\STBcmpsbpdD}      & \tb{\STCcmpsbpdD}   & {\STDcmpsbpdD}      & & \tb{\STEcmpsbpdD}   & {\STFcmpsbpdD}      & \tb{\STGcmpsbpdD}   & \tb{\STHcmpsbpdD}   & &    {\STMcmpsbpdD}   & {\STNcmpsbpdD}      &    {\STOcmpsbpdD}   & {\STPcmpsbpdD} \\
  \cmidrule(l{16pt}r{0pt}){1-15}
  $\quad$ \uu{Our (CMP [+ 6 MP3D Envs])}   & \uu{\STAcmpmpdD}    & \uu{\STBcmpmpdD}    & \uu{\STCcmpmpdD}    & \uu{\STDcmpmpdD}    & & \uu{\STEcmpmpdD}    & \uu{\STFcmpmpdD}    & \uu{\STGcmpmpdD}    & \uu{\STHcmpmpdD}    & & \uu{\STMcmpmpdD}    & \uu{\STNcmpmpdD}    & \uu{\STOcmpmpdD}    & \uu{\STPcmpmpdD} \\
  \bottomrule
  \end{tabular}}
\caption{\textbf{Results for Semantic Task:} 
We report success rate at 39 and 199 time steps, and SPL. We report
performance for individual categories as well as their average. Top part
reports comparisons with \rgb input, bottom part reports comparisons with depth
input.  We compare learning based methods and a classical baseline (based on
exploration and semantic segmentation). We also report performance of CMP when
trained with more data (+6 MP3D Envs).}
\tablelabel{sem-result}
% \vspace{-0.7cm}
\end{table*}

%% file: sem_spl_numbers.tex
\newcommand{\STAreactRGB}{24.8}
\newcommand{\STBreactRGB}{24.7}
\newcommand{\STCreactRGB}{18.6}
\newcommand{\STDreactRGB}{22.7}
\newcommand{\STEreactRGB}{19.1}
\newcommand{\STFreactRGB}{24.9}
\newcommand{\STGreactRGB}{19.7}
\newcommand{\STHreactRGB}{21.2}
\newcommand{\STIreactRGB}{23.2}
\newcommand{\STJreactRGB}{25.0}
\newcommand{\STKreactRGB}{18.1}
\newcommand{\STLreactRGB}{22.1}
\newcommand{\STMreactRGB}{12.7}
\newcommand{\STNreactRGB}{13.3}
\newcommand{\STOreactRGB}{10.2}
\newcommand{\STPreactRGB}{12.1}
\newcommand{\STAlstmRGB}{22.7}
\newcommand{\STBlstmRGB}{30.9}
\newcommand{\STClstmRGB}{19.1}
\newcommand{\STDlstmRGB}{24.2}
\newcommand{\STElstmRGB}{24.7}
\newcommand{\STFlstmRGB}{32.0}
\newcommand{\STGlstmRGB}{21.5}
\newcommand{\STHlstmRGB}{26.0}
\newcommand{\STIlstmRGB}{22.0}
\newcommand{\STJlstmRGB}{30.3}
\newcommand{\STKlstmRGB}{21.0}
\newcommand{\STLlstmRGB}{24.4}
\newcommand{\STMlstmRGB}{\phz7.8}
\newcommand{\STNlstmRGB}{14.6}
\newcommand{\STOlstmRGB}{\phz6.9}
\newcommand{\STPlstmRGB}{\phz9.7}
\newcommand{\STAcmpRGB}{25.0}
\newcommand{\STBcmpRGB}{40.2}
\newcommand{\STCcmpRGB}{25.7}
\newcommand{\STDcmpRGB}{30.3}
\newcommand{\STEcmpRGB}{46.9}
\newcommand{\STFcmpRGB}{44.6}
\newcommand{\STGcmpRGB}{30.1}
\newcommand{\STHcmpRGB}{40.5}
\newcommand{\STIcmpRGB}{21.1}
\newcommand{\STJcmpRGB}{31.8}
\newcommand{\STKcmpRGB}{17.6}
\newcommand{\STLcmpRGB}{23.5}
\newcommand{\STMcmpRGB}{11.9}
\newcommand{\STNcmpRGB}{18.9}
\newcommand{\STOcmpRGB}{10.0}
\newcommand{\STPcmpRGB}{13.6}

\newcommand{\STAreactD}{28.4}
\newcommand{\STBreactD}{23.6}
\newcommand{\STCreactD}{29.2}
\newcommand{\STDreactD}{27.1}
\newcommand{\STEreactD}{33.8}
\newcommand{\STFreactD}{23.8}
\newcommand{\STGreactD}{38.7}
\newcommand{\STHreactD}{32.1}
\newcommand{\STIreactD}{34.0}
\newcommand{\STJreactD}{19.8}
\newcommand{\STKreactD}{26.7}
\newcommand{\STLreactD}{26.8}
\newcommand{\STMreactD}{14.0}
\newcommand{\STNreactD}{\phz9.8}
\newcommand{\STOreactD}{11.9}
\newcommand{\STPreactD}{11.9}
\newcommand{\STAlstmD}{22.1}
\newcommand{\STBlstmD}{28.9}
\newcommand{\STClstmD}{26.8}
\newcommand{\STDlstmD}{25.9}
\newcommand{\STElstmD}{26.0}
\newcommand{\STFlstmD}{30.6}
\newcommand{\STGlstmD}{31.2}
\newcommand{\STHlstmD}{29.3}
\newcommand{\STIlstmD}{25.0}
\newcommand{\STJlstmD}{27.6}
\newcommand{\STKlstmD}{26.8}
\newcommand{\STLlstmD}{26.5}
\newcommand{\STMlstmD}{10.6}
\newcommand{\STNlstmD}{15.1}
\newcommand{\STOlstmD}{13.2}
\newcommand{\STPlstmD}{13.0}
\newcommand{\STAcmpD}{48.4}
\newcommand{\STBcmpD}{41.3}
\newcommand{\STCcmpD}{34.6}
\newcommand{\STDcmpD}{41.5}
\newcommand{\STEcmpD}{66.6}
\newcommand{\STFcmpD}{42.6}
\newcommand{\STGcmpD}{43.6}
\newcommand{\STHcmpD}{51.0}
\newcommand{\STIcmpD}{61.8}
\newcommand{\STJcmpD}{38.3}
\newcommand{\STKcmpD}{40.1}
\newcommand{\STLcmpD}{46.8}
\newcommand{\STMcmpD}{22.1}
\newcommand{\STNcmpD}{20.0}
\newcommand{\STOcmpD}{17.4}
\newcommand{\STPcmpD}{19.8}

\newcommand{\STAclassicalRGB}{11.6}
\newcommand{\STBclassicalRGB}{28.3}
\newcommand{\STCclassicalRGB}{13.6}
\newcommand{\STDclassicalRGB}{17.8}
\newcommand{\STEclassicalRGB}{16.2}
\newcommand{\STFclassicalRGB}{29.1}
\newcommand{\STGclassicalRGB}{22.2}
\newcommand{\STHclassicalRGB}{22.5}
\newcommand{\STIclassicalRGB}{16.2}
\newcommand{\STJclassicalRGB}{29.1}
\newcommand{\STKclassicalRGB}{22.2}
\newcommand{\STLclassicalRGB}{22.5}
\newcommand{\STMclassicalRGB}{\phz2.5}
\newcommand{\STNclassicalRGB}{10.1}
\newcommand{\STOclassicalRGB}{\phz5.1}
\newcommand{\STPclassicalRGB}{\phz5.9}
\newcommand{\STAcmpsbpdRGB}{17.9}
\newcommand{\STBcmpsbpdRGB}{41.5}
\newcommand{\STCcmpsbpdRGB}{25.9}
\newcommand{\STDcmpsbpdRGB}{28.4}
\newcommand{\STEcmpsbpdRGB}{35.3}
\newcommand{\STFcmpsbpdRGB}{44.6}
\newcommand{\STGcmpsbpdRGB}{32.1}
\newcommand{\STHcmpsbpdRGB}{37.3}
\newcommand{\STIcmpsbpdRGB}{15.6}
\newcommand{\STJcmpsbpdRGB}{31.8}
\newcommand{\STKcmpsbpdRGB}{21.0}
\newcommand{\STLcmpsbpdRGB}{22.8}
\newcommand{\STMcmpsbpdRGB}{\phz9.5}
\newcommand{\STNcmpsbpdRGB}{18.8}
\newcommand{\STOcmpsbpdRGB}{13.6}
\newcommand{\STPcmpsbpdRGB}{14.0}
\newcommand{\STAcmpmpdRGB}{42.8}
\newcommand{\STBcmpmpdRGB}{40.3}
\newcommand{\STCcmpmpdRGB}{21.0}
\newcommand{\STDcmpmpdRGB}{34.7}
\newcommand{\STEcmpmpdRGB}{80.3}
\newcommand{\STFcmpmpdRGB}{42.2}
\newcommand{\STGcmpmpdRGB}{40.7}
\newcommand{\STHcmpmpdRGB}{54.4}
\newcommand{\STIcmpmpdRGB}{73.4}
\newcommand{\STJcmpmpdRGB}{41.1}
\newcommand{\STKcmpmpdRGB}{38.3}
\newcommand{\STLcmpmpdRGB}{50.9}
\newcommand{\STMcmpmpdRGB}{21.1}
\newcommand{\STNcmpmpdRGB}{20.7}
\newcommand{\STOcmpmpdRGB}{15.1}
\newcommand{\STPcmpmpdRGB}{19.0}

\newcommand{\STAclassicalD}{38.0}
\newcommand{\STBclassicalD}{72.4}
\newcommand{\STCclassicalD}{28.5}
\newcommand{\STDclassicalD}{46.3}
\newcommand{\STEclassicalD}{38.9}
\newcommand{\STFclassicalD}{49.8}
\newcommand{\STGclassicalD}{43.0}
\newcommand{\STHclassicalD}{43.9}
\newcommand{\STIclassicalD}{38.9}
\newcommand{\STJclassicalD}{49.8}
\newcommand{\STKclassicalD}{43.0}
\newcommand{\STLclassicalD}{43.9}
\newcommand{\STMclassicalD}{27.5}
\newcommand{\STNclassicalD}{24.7}
\newcommand{\STOclassicalD}{24.3}
\newcommand{\STPclassicalD}{25.5}
\newcommand{\STAcmpsbpdD}{50.2}
\newcommand{\STBcmpsbpdD}{41.9}
\newcommand{\STCcmpsbpdD}{34.6}
\newcommand{\STDcmpsbpdD}{42.2}
\newcommand{\STEcmpsbpdD}{67.7}
\newcommand{\STFcmpsbpdD}{43.4}
\newcommand{\STGcmpsbpdD}{44.7}
\newcommand{\STHcmpsbpdD}{51.9}
\newcommand{\STIcmpsbpdD}{63.7}
\newcommand{\STJcmpsbpdD}{38.8}
\newcommand{\STKcmpsbpdD}{39.7}
\newcommand{\STLcmpsbpdD}{47.4}
\newcommand{\STMcmpsbpdD}{22.4}
\newcommand{\STNcmpsbpdD}{19.9}
\newcommand{\STOcmpsbpdD}{17.9}
\newcommand{\STPcmpsbpdD}{20.1}
\newcommand{\STAcmpmpdD}{43.9}
\newcommand{\STBcmpmpdD}{43.2}
\newcommand{\STCcmpmpdD}{42.5}
\newcommand{\STDcmpmpdD}{43.2}
\newcommand{\STEcmpmpdD}{60.4}
\newcommand{\STFcmpmpdD}{64.5}
\newcommand{\STGcmpmpdD}{66.5}
\newcommand{\STHcmpmpdD}{63.8}
\newcommand{\STIcmpmpdD}{55.8}
\newcommand{\STJcmpmpdD}{54.8}
\newcommand{\STKcmpmpdD}{60.3}
\newcommand{\STLcmpmpdD}{57.0}
\newcommand{\STMcmpmpdD}{19.5}
\newcommand{\STNcmpmpdD}{22.9}
\newcommand{\STOcmpmpdD}{26.1}
\newcommand{\STPcmpmpdD}{22.8}

%% file: visualizations.tex
\subsection{Visualizations}
\seclabel{vis}

\begin{figure}
  \centering
  \insertH{0.190}{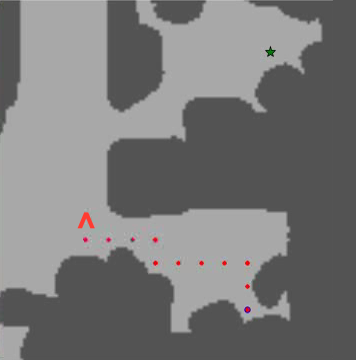} \insertH{0.190}{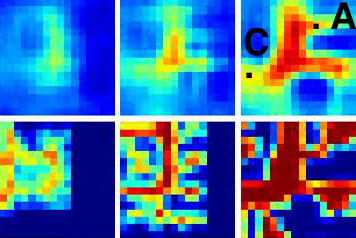} \\ \vspace{0.25cm}
  \insertH{0.190}{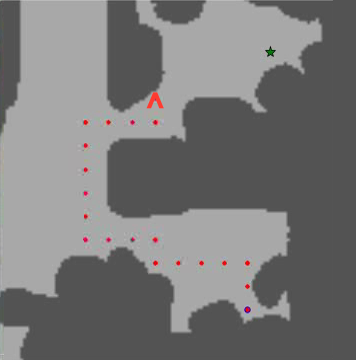} \insertH{0.190}{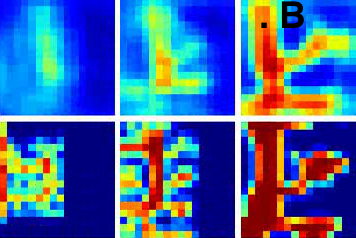} 
  \caption{We visualize the output of the map readout function trained on the
representation learned by the mapper (see text for details) as the agent moves
around. The two rows show two different time steps from an episode. For each
row, the gray map shows the current position and orientation of the agent (red
$\wedge$), and the locations that the agent has already visited during this
episode (red dots). The top three heatmaps show the output of the map readout
function and the bottom three heatmaps show the ground truth free space at the
three scales used by CMP (going from coarse to fine from left to right). We
observe that the readout maps capture the free space in the regions visited by
the agent (room entrance at point A, corridors at points B and C).}
\figlabel{rom}
\end{figure}

\begin{figure}
\centering
\insertWL{0.49}{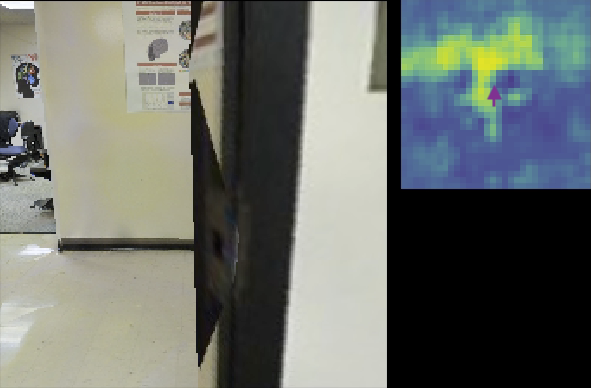} 
\insertWL{0.49}{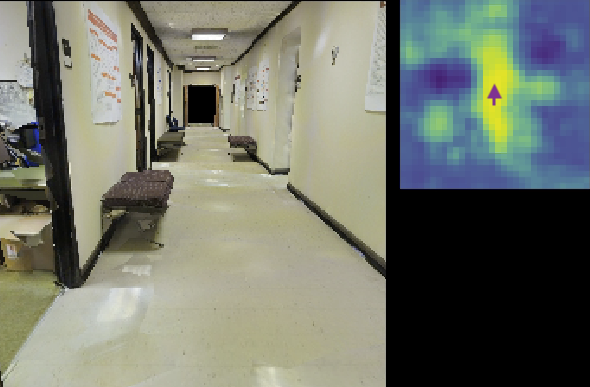} 
\caption{We visualize first-person images and the output of the readout function output for
free-space prediction derived from the representation produced by the mapper
module (in egocentric frame, that is the agent is at the center looking
upwards (denoted by the purple arrow)). In the left example, we can make a prediction behind the
wall, and in the right example, we can make predictions inside the room.}
\figlabel{rom2}
\end{figure}

\begin{figure}
\centering
\insertWL{1.00}{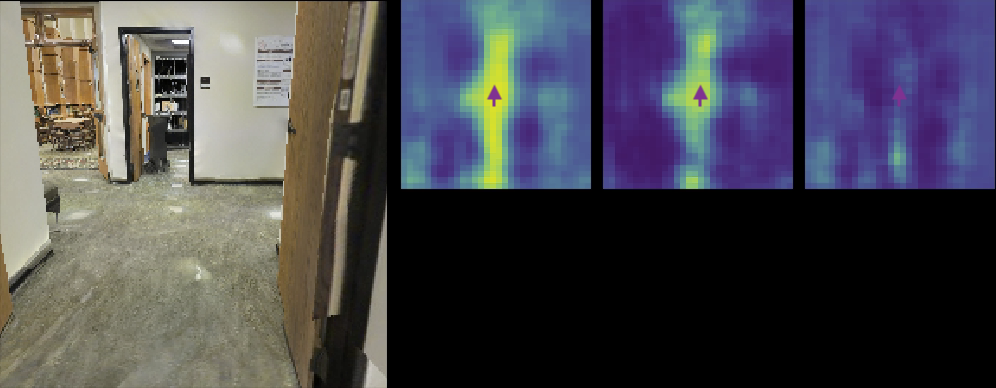} 
\insertWL{1.00}{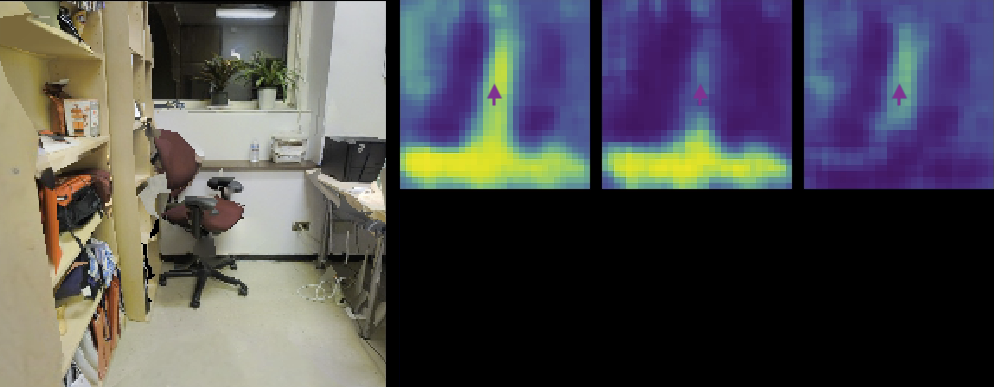} 
\caption{We visualize the first person image, prediction for all free space,
prediction for free space in a hallway, and prediction for free space inside a
room (in order). Once again, the predictions are in an egocentric coordinate
frame (agent (denoted by the purple arrow) is at the center and looking
upwards). The top figure pane shows the case when the agent is actually in a
hallway, while the bottom figure pane shows the case when the agent is inside a
room.}
\figlabel{rom3}
\end{figure}

\begin{figure}
\centering
\insertWL{1.00}{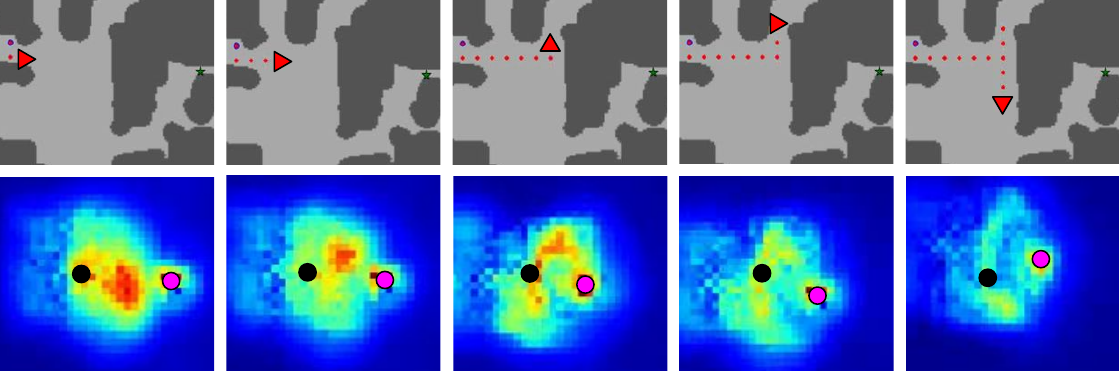} \\ \caption{\small We visualize the value
function for five snapshots for an episode for the single scale version of our
model. The top row shows the agent's location and orientation with a red
triangle, nodes that the agent has visited with red dots and the goal location
with the green star. Bottom row shows a 1 channel projection of the value maps
(obtained by taking the channel wise max) and visualizes the agent location by
the black dot and the goal location by the pink dot. Initially the agent plans
to go straight ahead, as it sees the wall it develops an inclination to turn
left. It then turns into the room (center figure), planning to go up and around
to the goal but as it turns again it realizes that that path is blocked (center
right figure). At this point the value function changes (the connection to the
goal through the top room becomes weaker) and the agent approaches the goal via
the downward path.} \figlabel{vis}
\end{figure}

We visualize activations at different layers in the CMP network to check if the
architecture conforms to the intuitions that inspired the design of the
network. We check for the following three aspects: 
a) is the representation produced by the mapper indeed spatial, 
b) does the mapper capture anything beyond what a purely geometric mapping
pipeline captures, and 
c) do the value maps obtained from the value iteration module capture the
behaviour exhibited by the agent.

\textbf{Is the representation produced by the mapper spatial?}
We train simple readout functions on the learned mapper representation to
predict free space around the agent. \figref{rom} visualizes the output of
these readout functions at two time steps from an episode as the agent moves.
We see that the representation produced by the mapper is in correspondence with
the actual free space around the agent. The representation produced by the
mapper is indeed spatial in nature. We also note that readouts are generally
better at finer scales. 

\textbf{What does the mapper representation capture?}
We next try to understand as to what information is captured in these spatial
representations. First, as discussed above the representation produced by the
mapper can be used to predict free space around the agent. Note that the agent
was never trained to predict free space, yet the representations produced by
the mapper carry enough information to predict free space reasonable well.
Second, \figref{rom2} shows free space prediction for
two cases where the agent is looking through a doorway. We see that the mapper
representation is expressive enough to make reasonable predictions for free
space behind the doorway. This is something that a purely geometric system that
only reasons about directly visible parts of the environment is simply
incapable of doing. Finally, we show the output of readout functions that were
trained for differentiating between free space in a hallway \vs free space in a
room. \figref{rom3}~(top) shows the prediction for when the agent is out in the
hallway, and \figref{rom3}~(bottom) shows the prediction for when the agent is
in a room. We see that the representation produced by the mapper can reasonably
distinguish between free space in a hallway \vs free space in a room, even
though it was never explicitly trained to do so. Once again, this is something
that a purely geometric description of the world will be unable to capture.

\begin{figure*}
\centering
\insertH{0.1700}{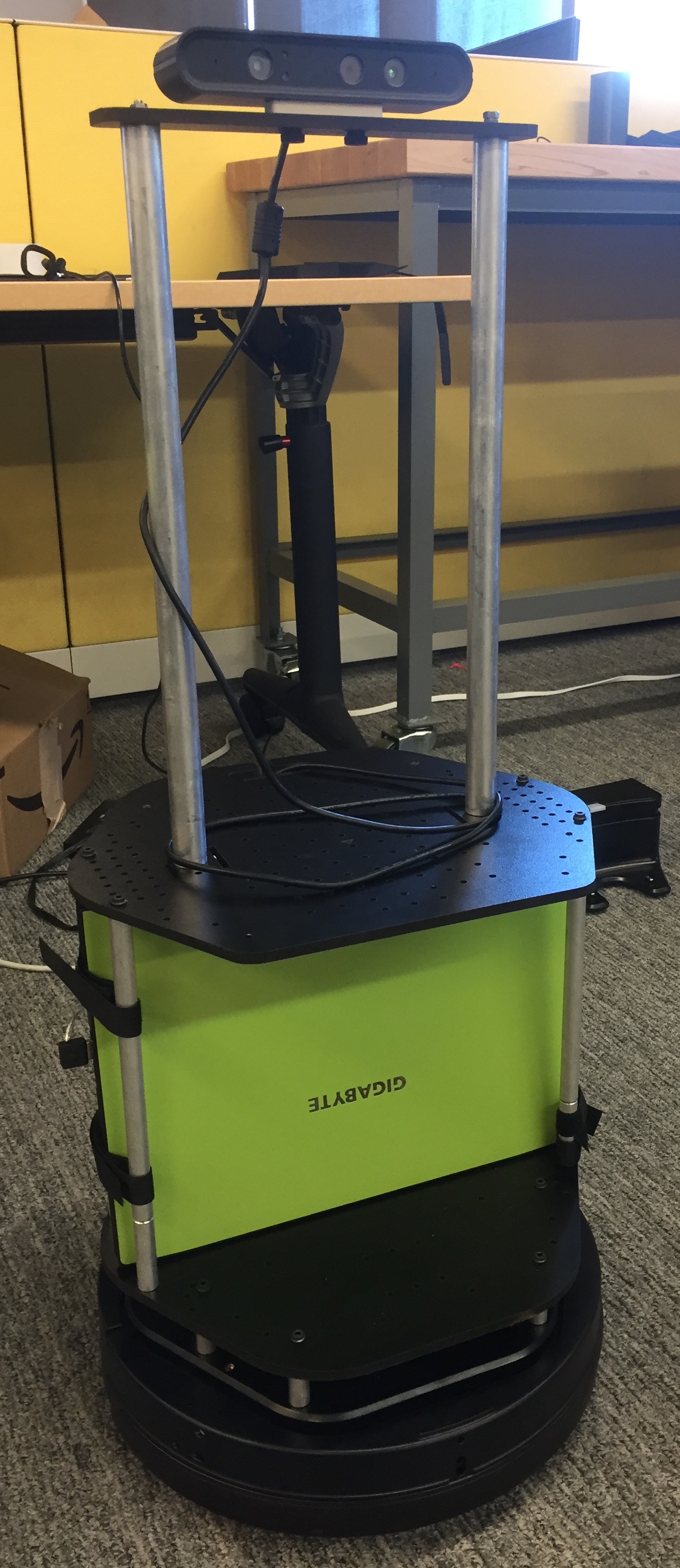} \insertH{0.1700}{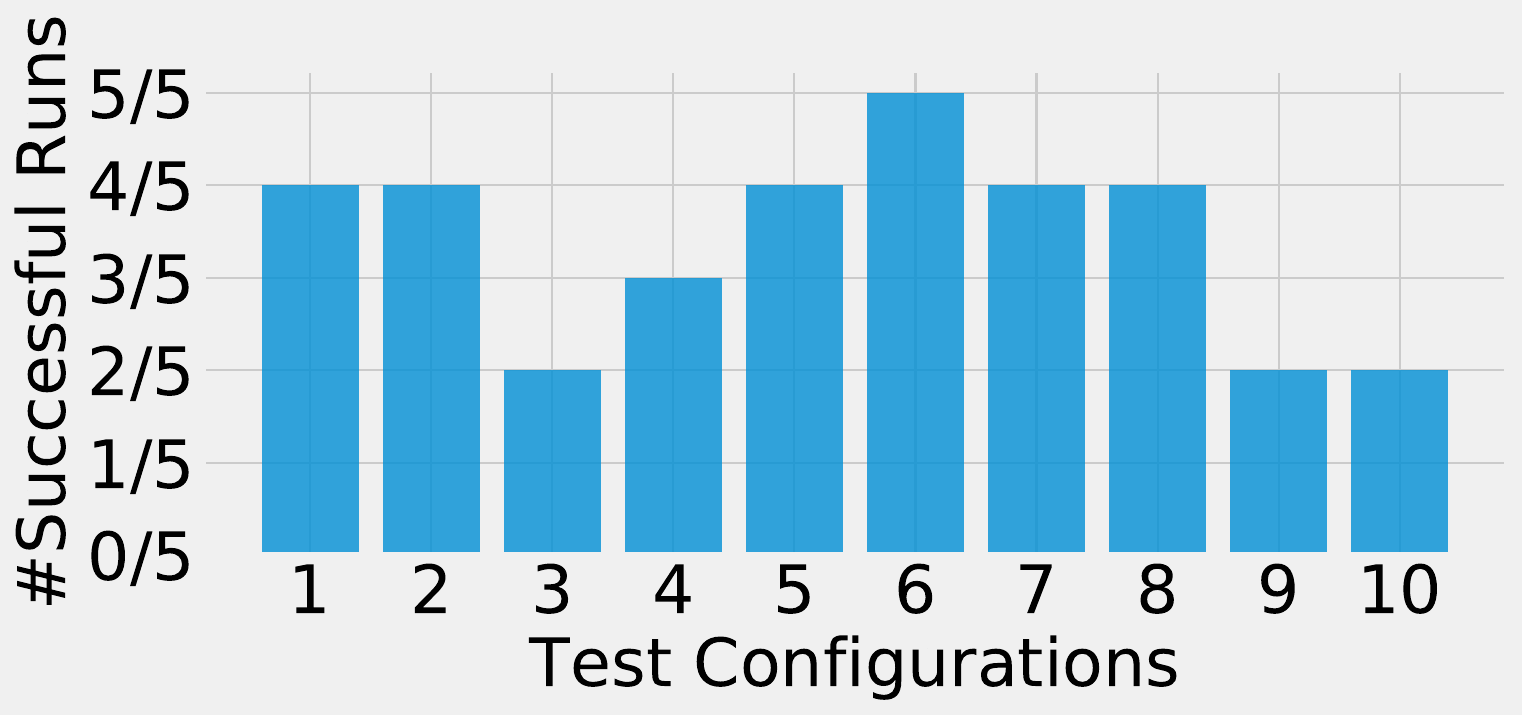} \insertH{0.1700}{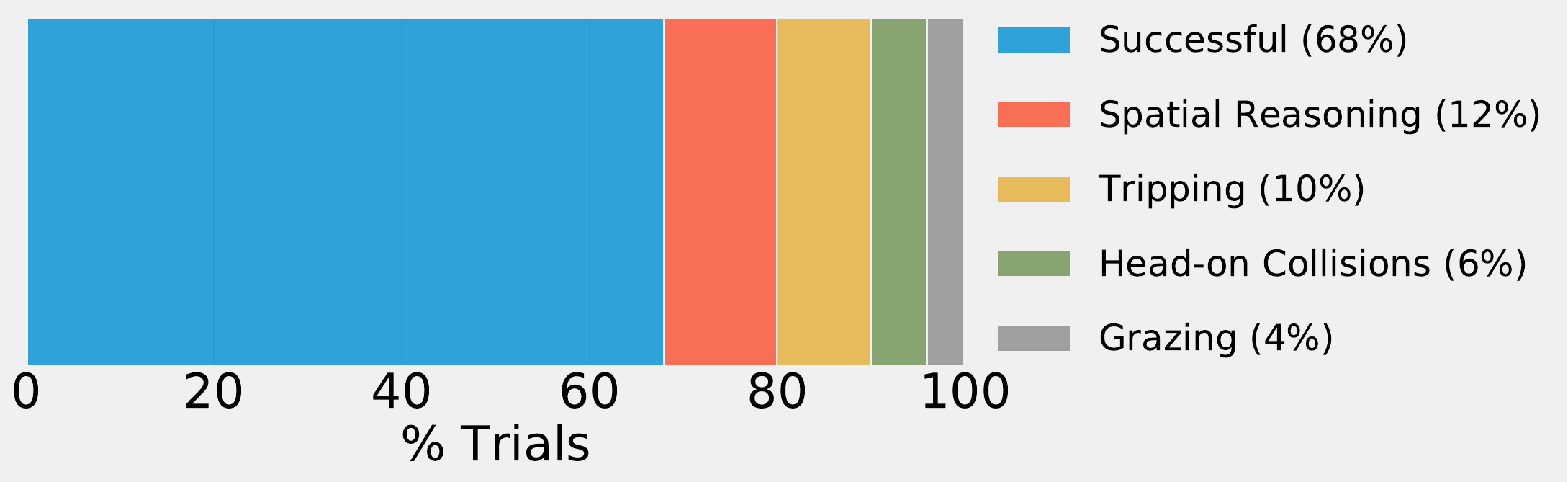}
\caption{\textbf{Real World Deployment:} We report success rate on different
test cases for real world deployment of our policy on TurtleBot~2. The policy
was trained for the geometric task using \rgb images in simulation. Right plot
shows breakdown of runs. 68\% runs succeeded, 20\% runs failed due to
infractions, and the remaining 12\% runs failed as the agent was unable to go
around obstacles.}
\figlabel{robot}
\end{figure*}

\textbf{Do the value maps obtained from the value iteration module capture the
behaviour exhibited by the agent?}
Finally, \figref{vis} visualizes a one channel projection of the value map for
the single scale version of our model at five time steps from an episode. We
can see that the value map is indicative of the current actions that the agent
takes, and how the value maps change as the agent discovers that the previously
hypothesised path was infeasible.

%% file: realrobot.tex
\input{realtests}

\section{Real World Deployment}
\seclabel{realrobot}
We have also deployed these learned policies on a real robot. We describe
the robot setup, implementation details and our results below.

\textbf{Robot description.}
We conducted our experiments on a TurtleBot~2 robot. TurtleBot~2 is a
differential drive platform based on the Yujin Kobuki Base. We mounted an Orbbec
Astra camera at a height of $80cm$, and a GPU-equipped high-end gaming laptop
(Gigabyte Aero 15'' with an NVIDIA 1060 GPU). The robot is shown in
\figref{robot}~(left). We used ROS to interface with the robot and the camera. We read
out images from the camera, and an estimate of the robot's $2D$ position and
orientation obtained from wheel encoders and an onboard inertial measurement
unit (IMU). We controlled the robot by specifying desired linear and angular
velocities. These desired velocity commands are internally used to determine
the voltage that is applied to the two motors through a
proportional integral derivative (PID) controller. Note that TurtleBot~2 is a
non-holonomic system. It only moves in the direction it is facing, and its
dynamics can be approximated as a Dubins Car.

\textbf{Implementation of macro-actions.}
Our policies output macro actions (rotate left or right by $90^\circ$, move
forward $40cm$). Unlike past work \cite{bruce2018learning} that uses human
operators to implement such macro-actions for such simulation to real transfer,
we implement these macro-actions using an iterative linear–quadratic regulator
(iLQR) controller \cite{jacobson1970differential, li2004iterative}.  iLQR
leverages known system dynamics to output a dynamically feasible local
reference trajectory (sequence of states and controls) that can convey the
system from a specified initial state to a specified final state (in our case,
rotation of $90^\circ$ or forward motion of $40cm$). Additionally, iLQR is a
state-space feedback controller. It estimates time-varying feedback matrices,
that can adjust the reference controls to compensate for deviations from the
reference trajectory (due to mis-match in system dynamics or noise in the
environment). These adjusted controls are applied to the robot. More details
are provided in \secrefext{ilqr}.

\textbf{Policy.} We deployed the policy for the geometric task onto the robot.
As all other policies, this policy was trained entirely in simulation. We used
the `CMP [+6 MP3D Env]' policy that was trained with the six additional large
environments from Matterport3D dataset \cite{Matterport3D} (on top of the 4
environments from the S3DIS \cite{armeni20163d} dataset). 
Apart from improves performance in simulation (SPL from $59.6\%$ to $70.8\%$),
it also exhibited better real world behavior in preliminary runs.

\textbf{Results.}
We ran the robot in 10 different test configurations (shown in
\figref{realtests}). These tests were picked such that there was no straight
path to the goal location, and involved situation like getting out of a room,
going from one cubicle to another, and going around tables and kitchen
counters. We found the depth as sensed from the Orbbec camera to be very noisy (and
different from depth produced in our simulator), and hence only conducted
experiments with \rgb images as input. We conducted 5 runs for each of the 10
different test configurations, and report the success rate for the 10
configurations in \figref{robot}~(middle). A run was considered successful
if the robot made it close to the specified target location (within $80cm$)
without brushing against or colliding with any objects. Sample videos of
execution are available on the project website. The policy achieved a success
rate of $68\%$. Executed trajectories are plotted in \figref{realtests}. This
is a very encouraging result, given that the policy was trained entirely in
simulation on very different buildings, and the lack of any form of domain
adaptation. Our robot, that only uses monocular \rgb images, successfully
avoids running into obstacles and arrives at the goal location for a number of
test cases.

\figref{robot}~(right) presents failure modes of our runs.  10 of the 16
failures are due to infractions (head-on collisions, grazing against objects,
and tripping over rods on the floor). These failures can possibly be mitigated
by use of a finer action space for more dexterous motion, additional
instrumentation such as near range obstacle detection, or coupling with a
collision avoidance system. The remaining 6 failures correspond to not going
around obstacles, possibly due to inaccurate perception.

%% file: realtests.tex
\begin{figure}
\centering
\insertH{0.095}{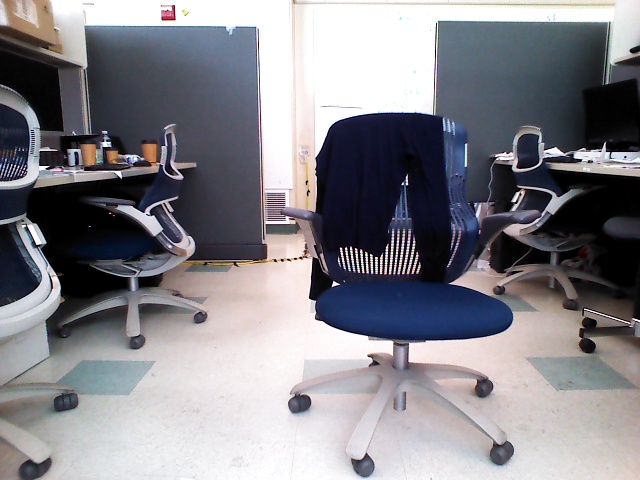}~
\insertH{0.095}{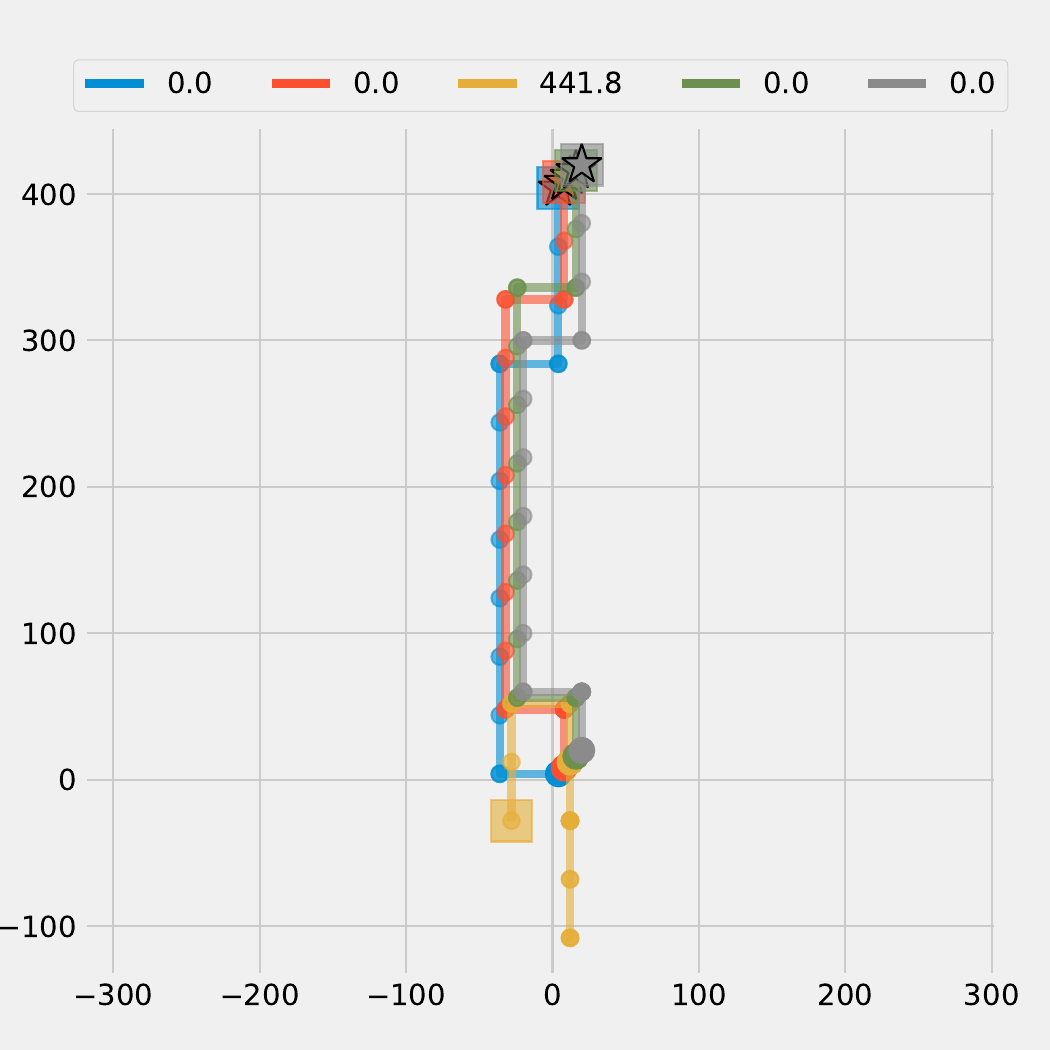} \:
\insertH{0.095}{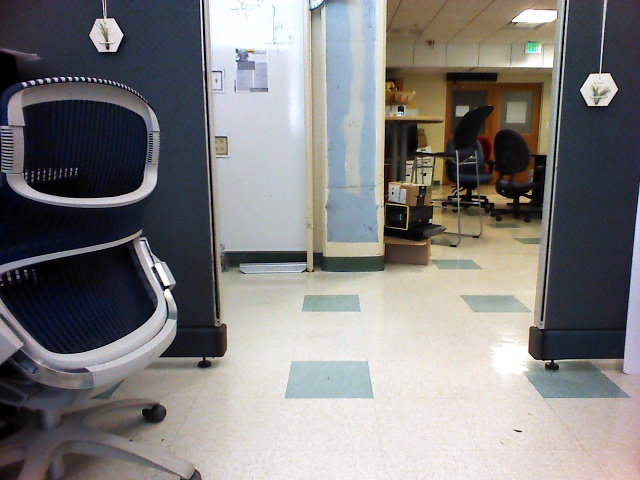}~
\insertH{0.095}{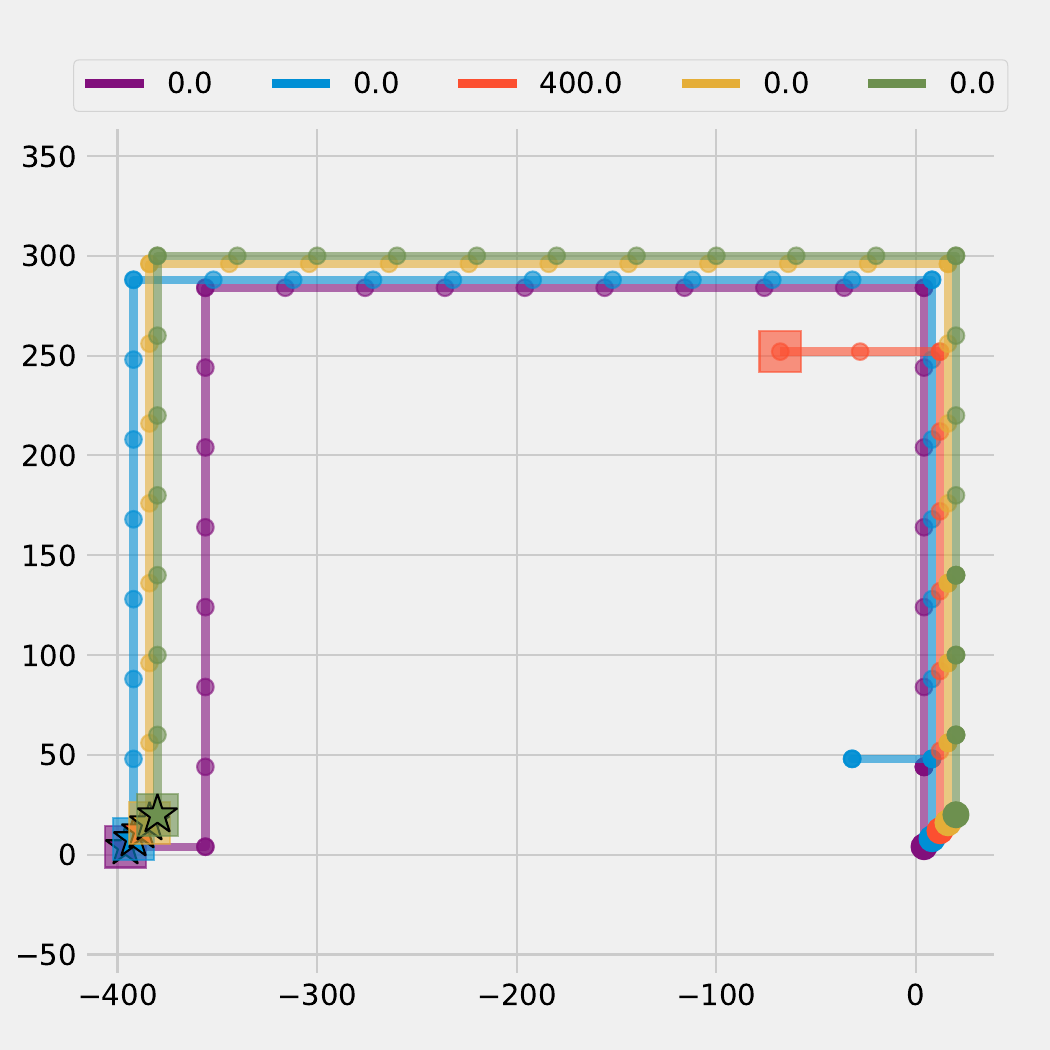} \:
\insertH{0.095}{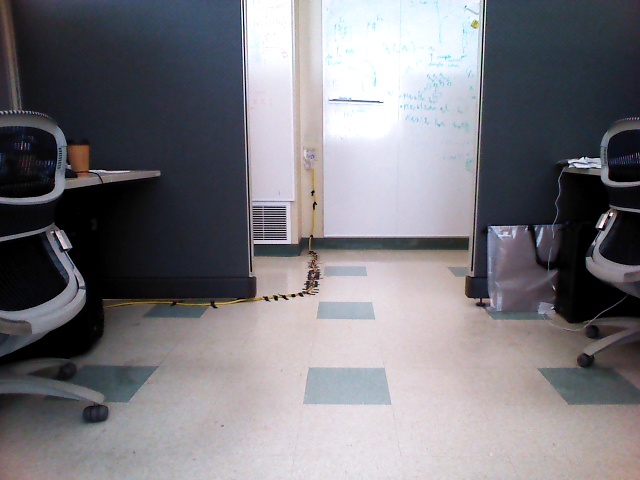}~
\insertH{0.095}{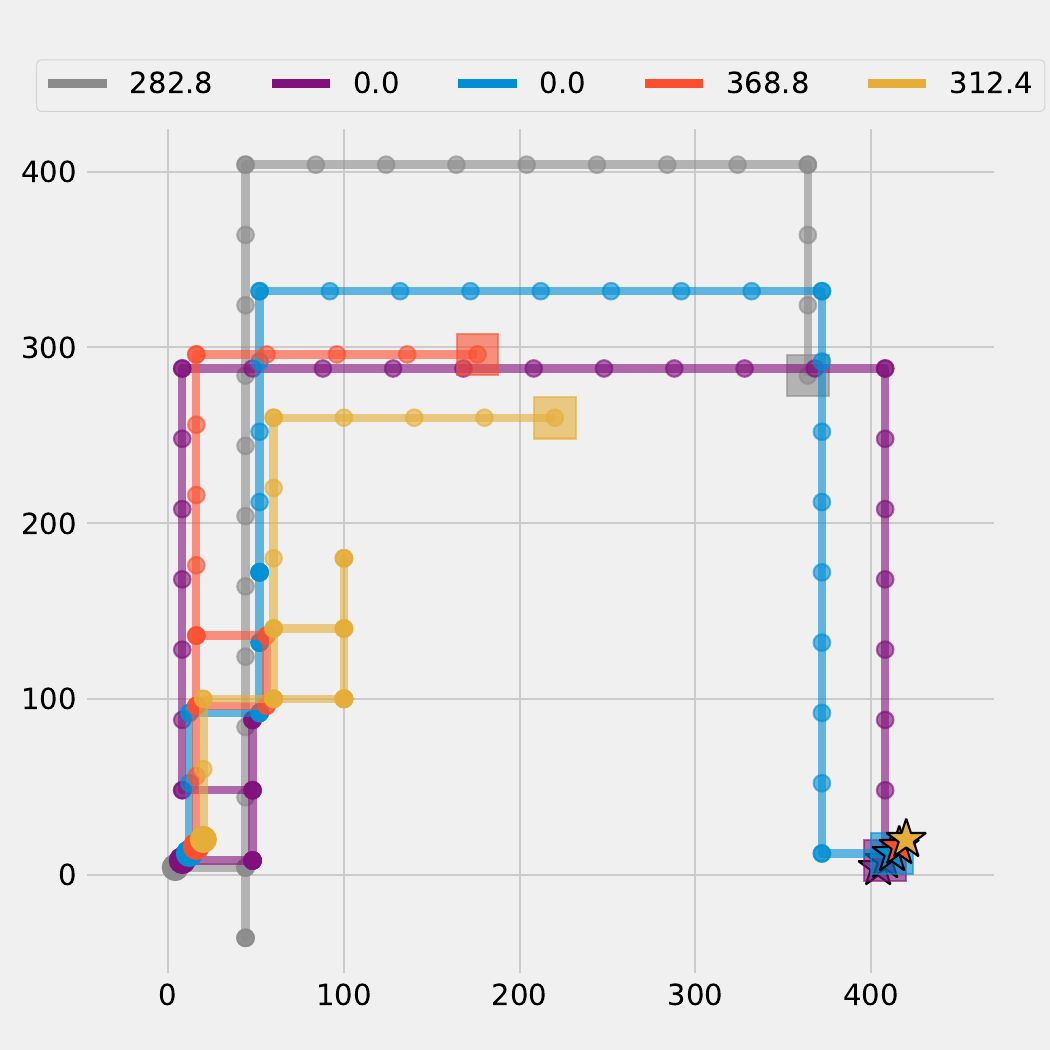} \:
\insertH{0.095}{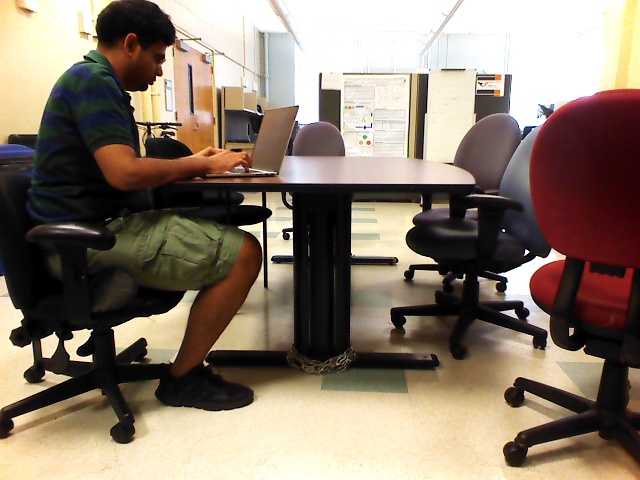}~
\insertH{0.095}{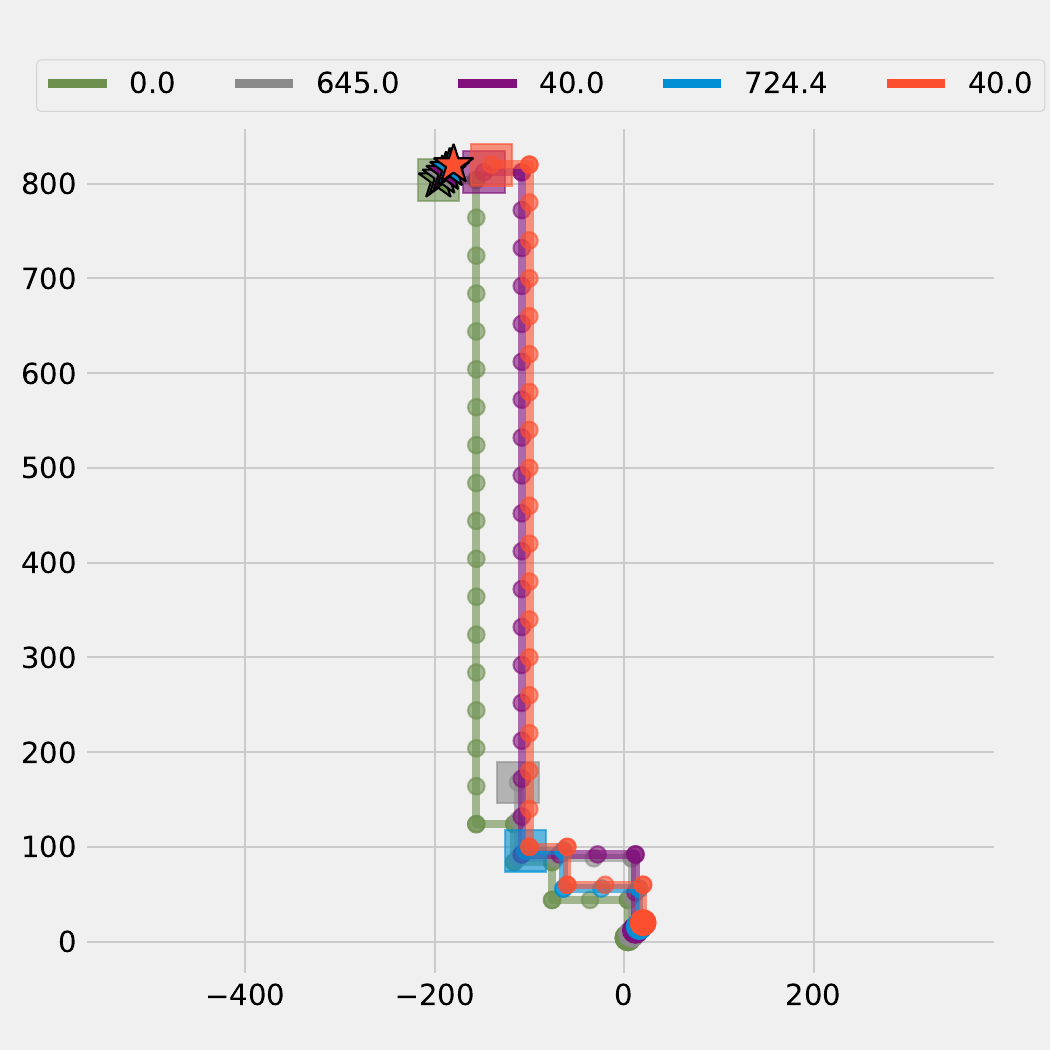} \:
\insertH{0.095}{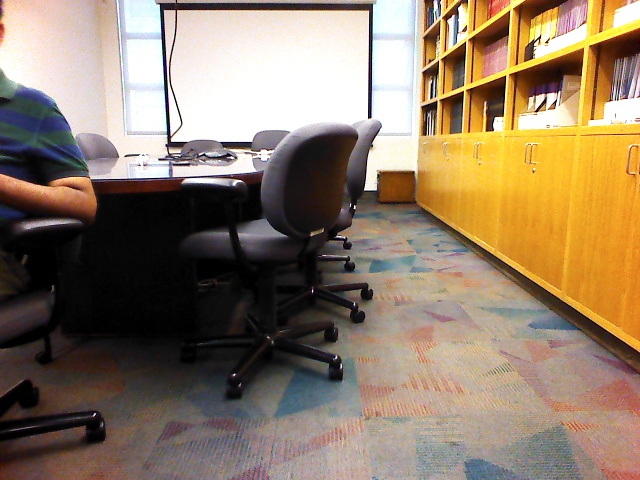}~
\insertH{0.095}{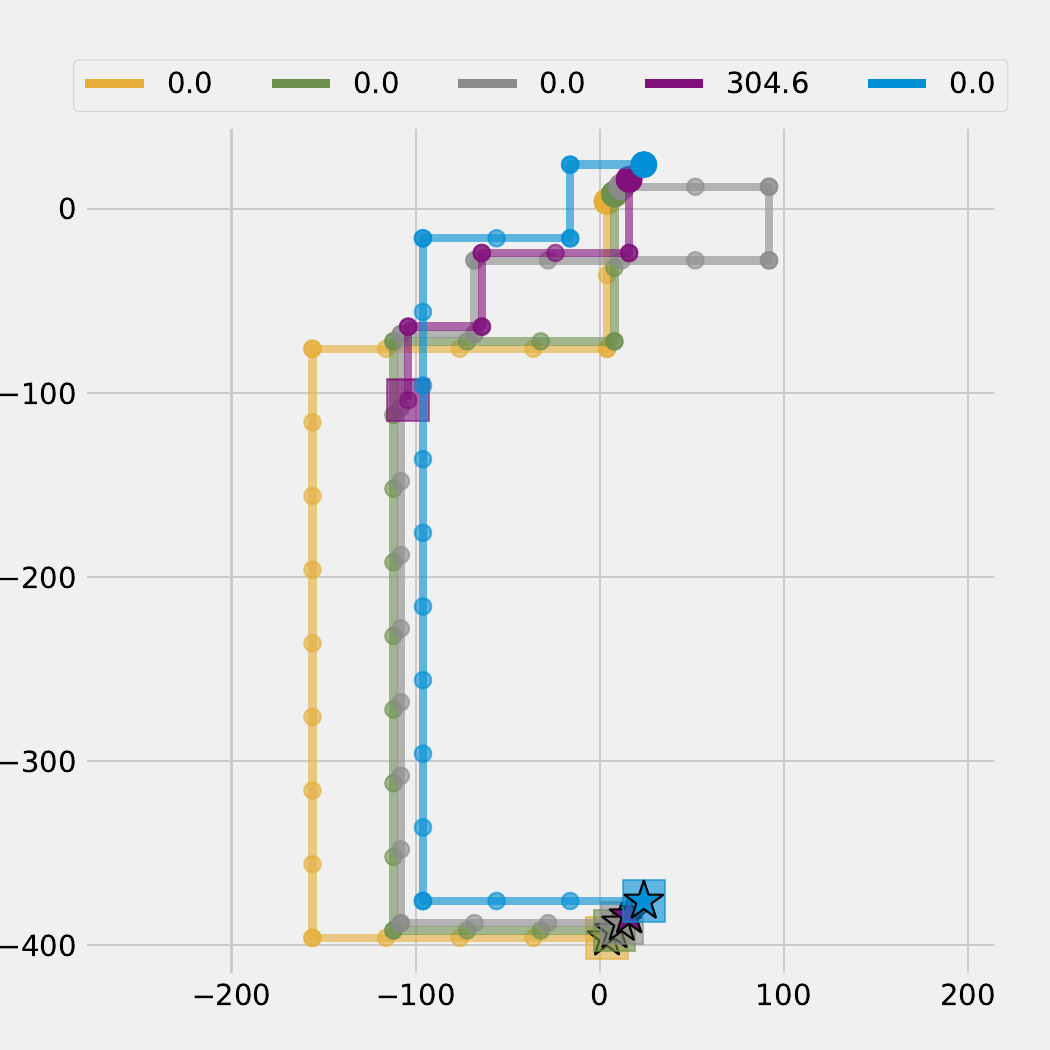} \:
\insertH{0.095}{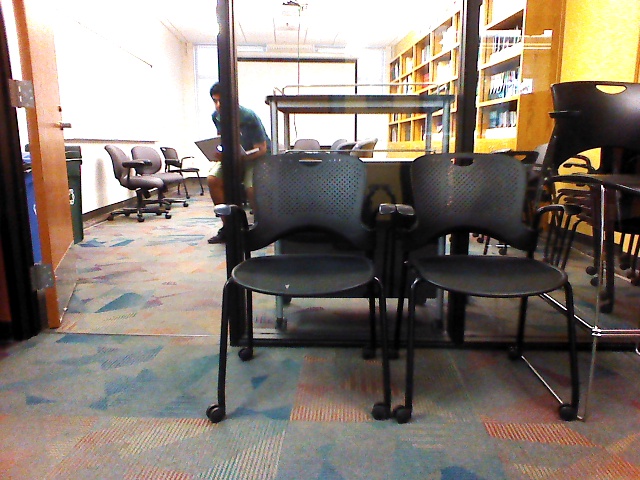}~
\insertH{0.095}{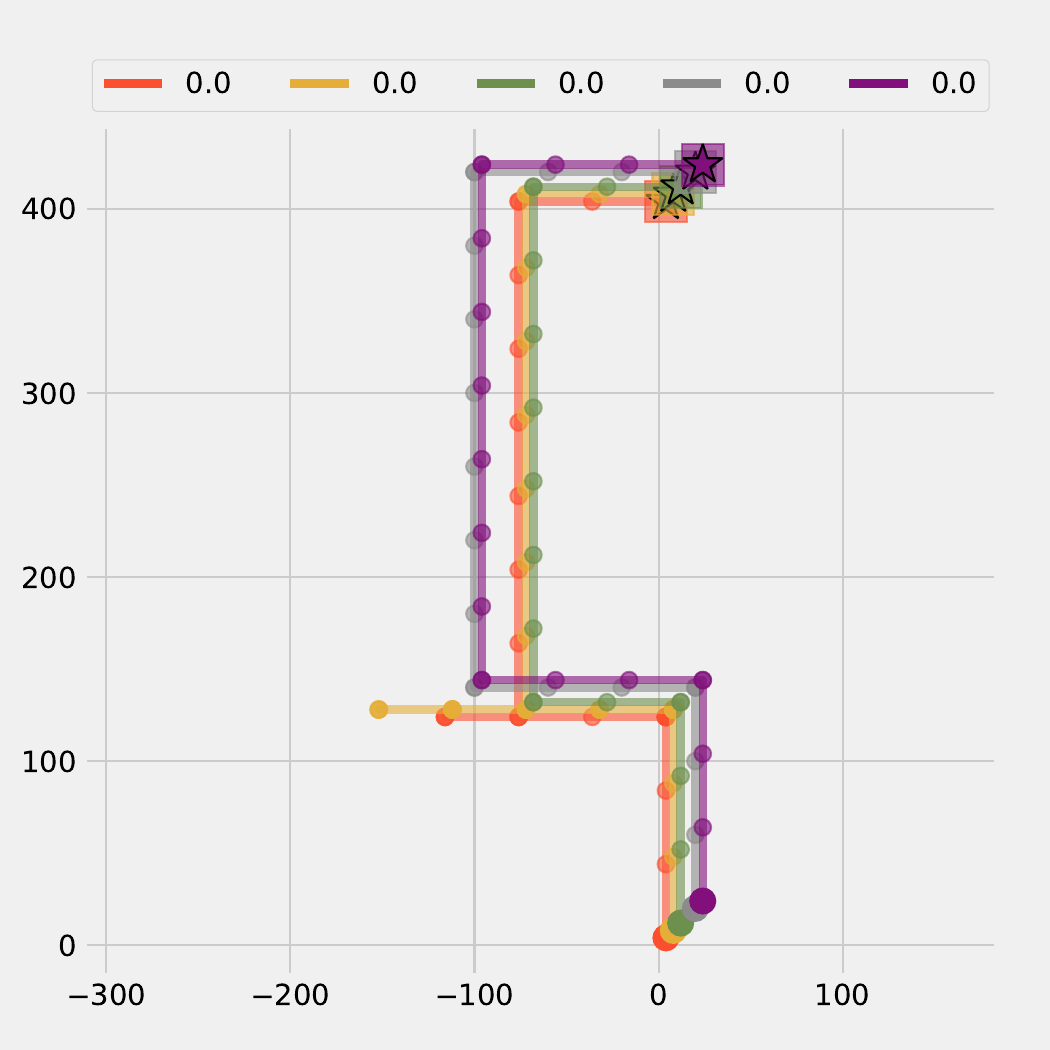} \:
\insertH{0.095}{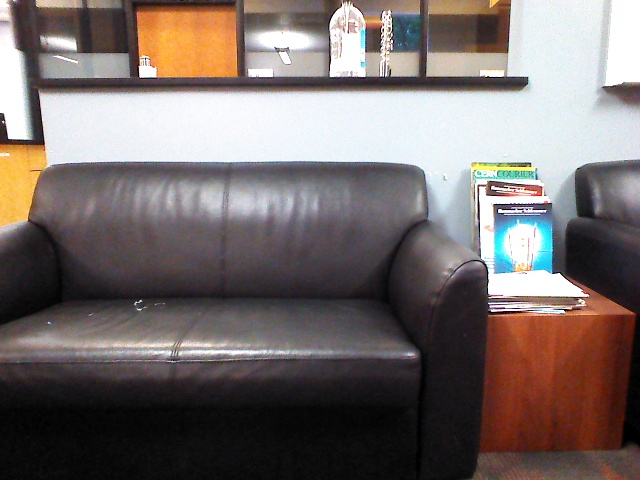}~
\insertH{0.095}{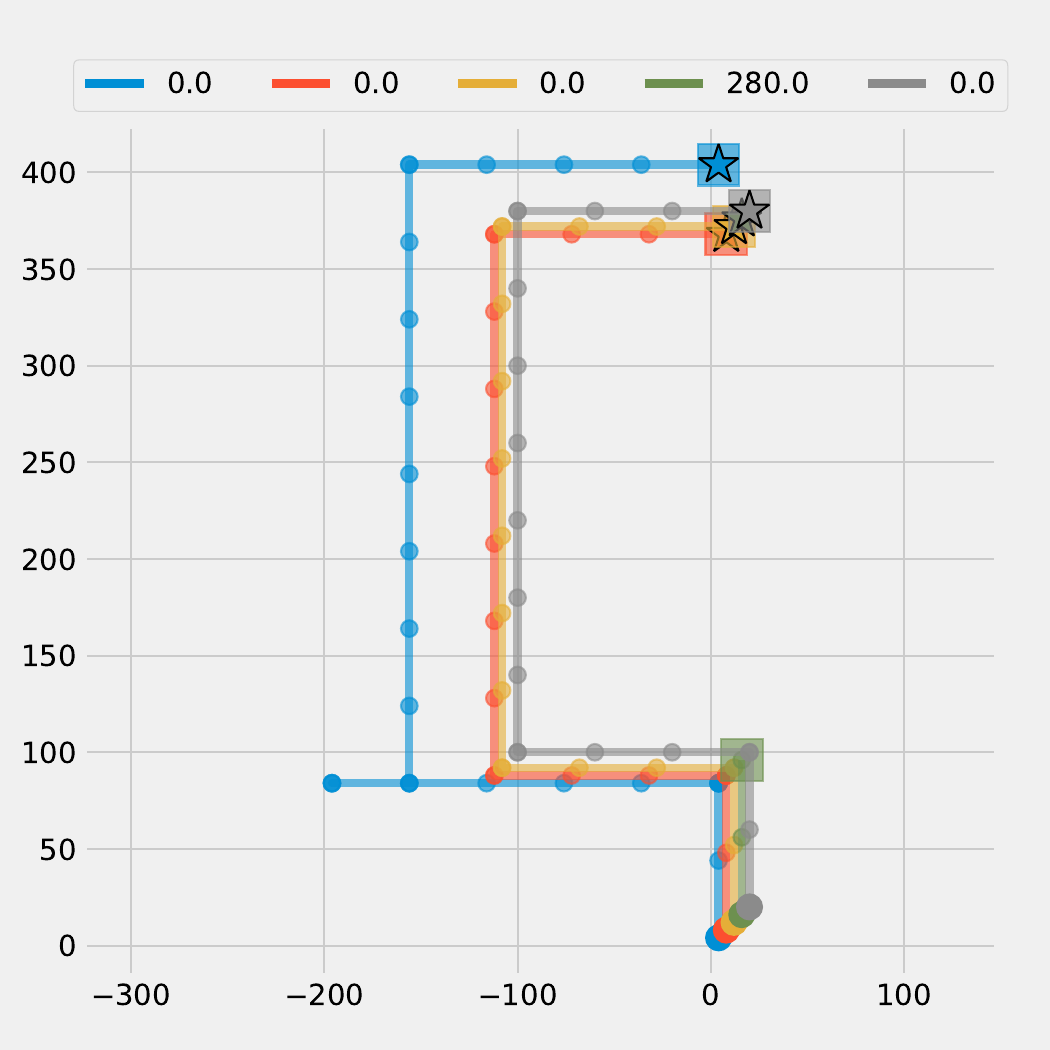} \:
\insertH{0.095}{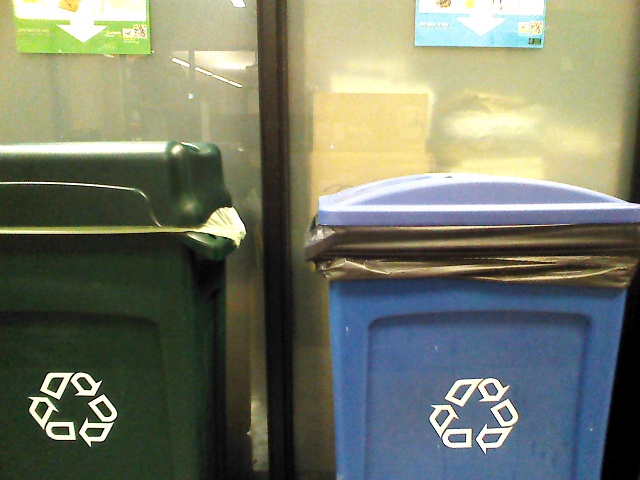}~
\insertH{0.095}{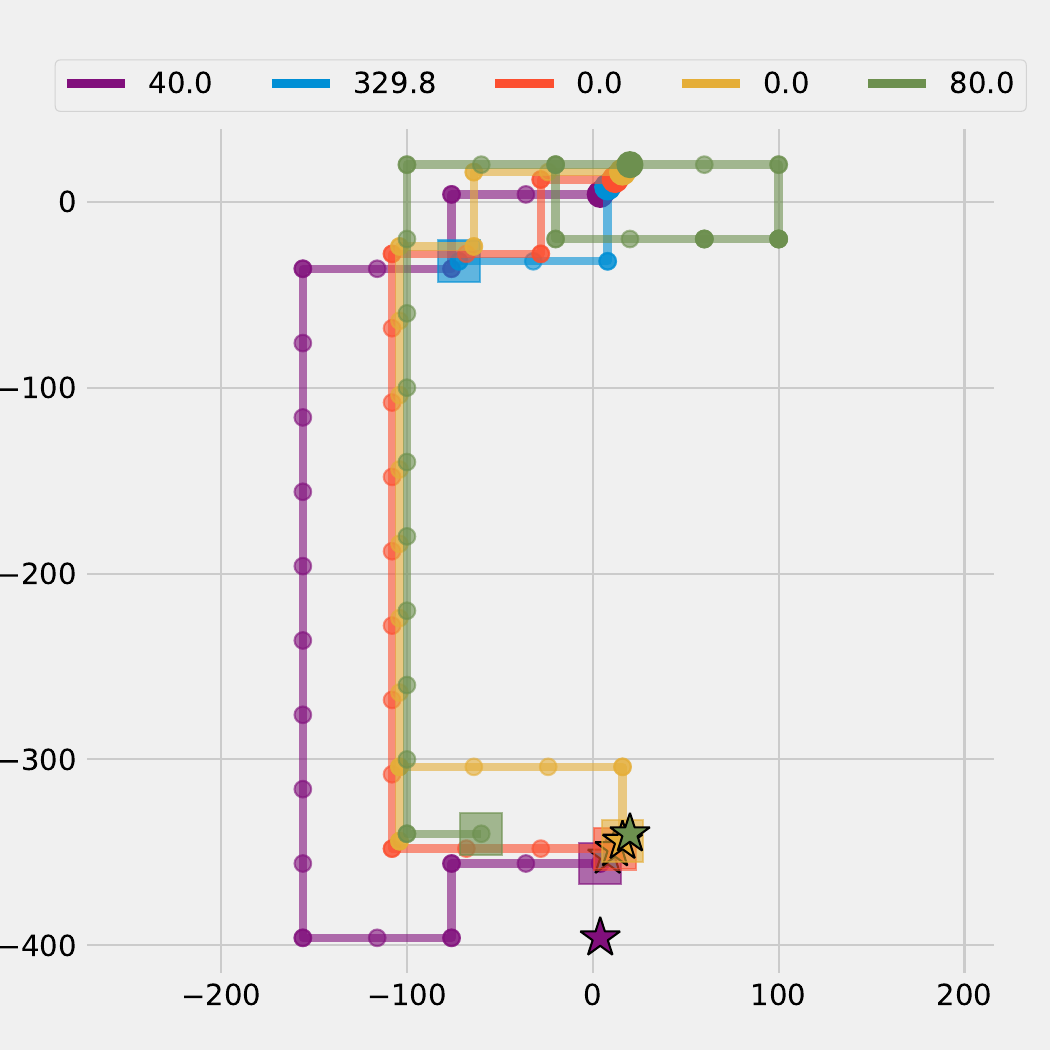} \:
\insertH{0.095}{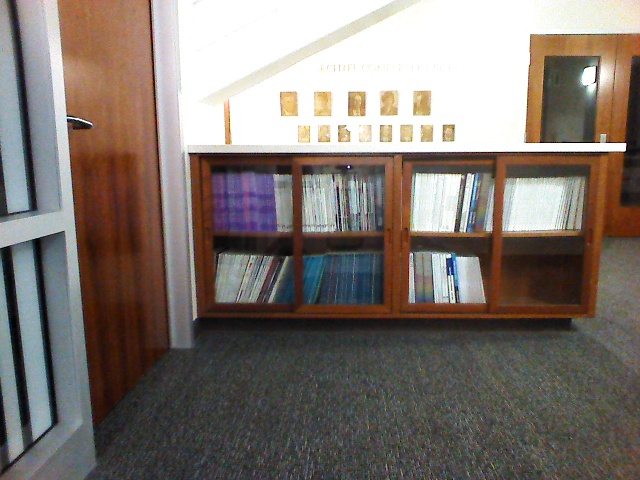}~
\insertH{0.095}{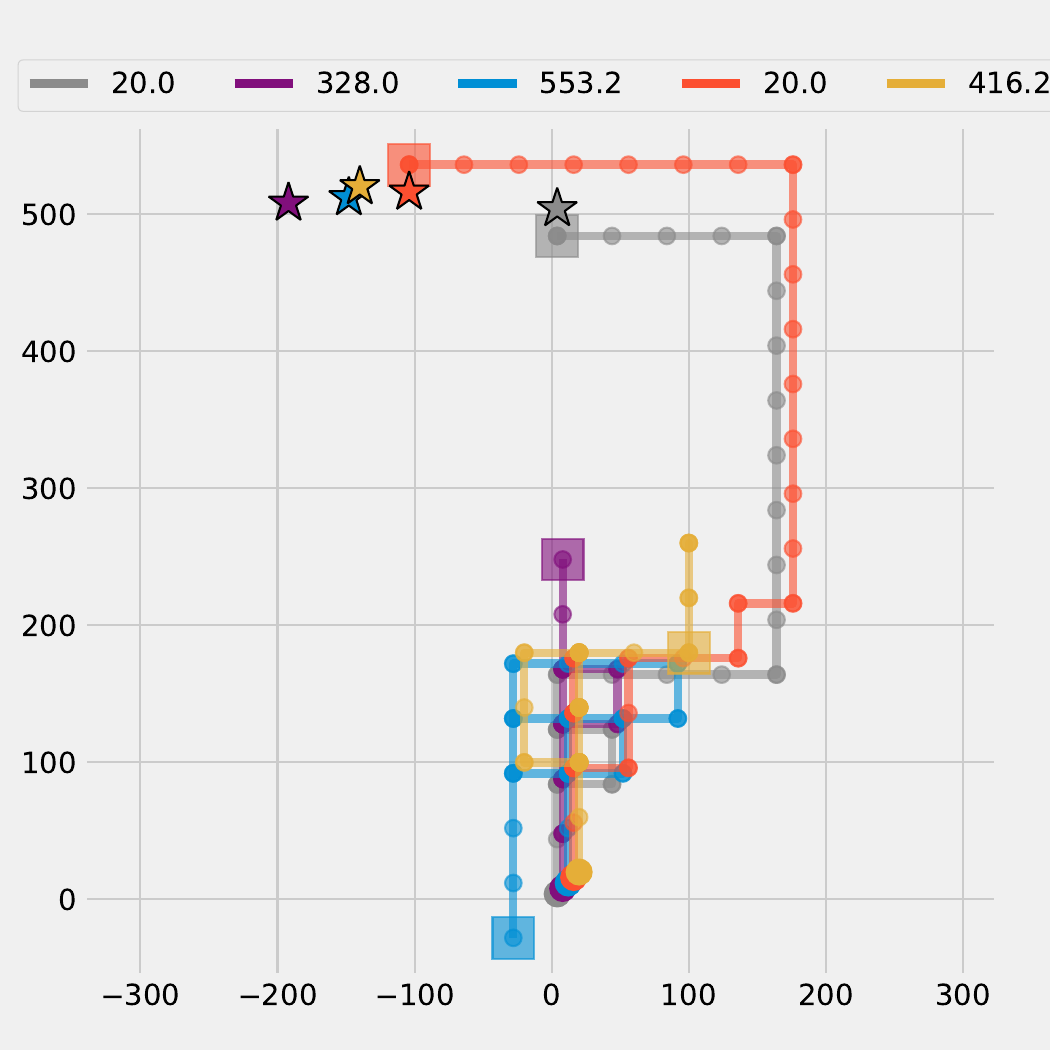} \:
\insertH{0.095}{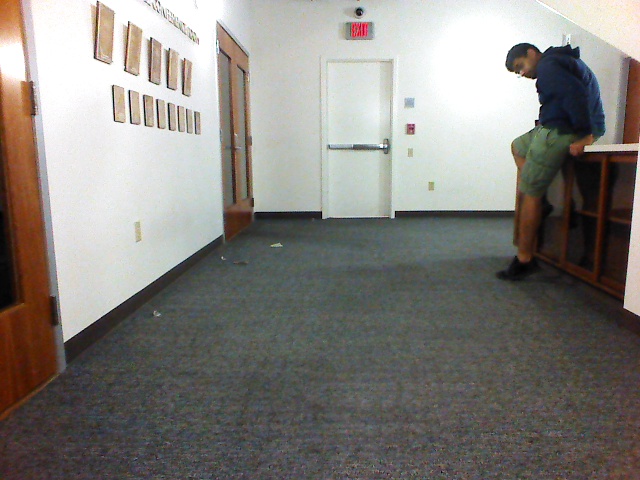}~
\insertH{0.095}{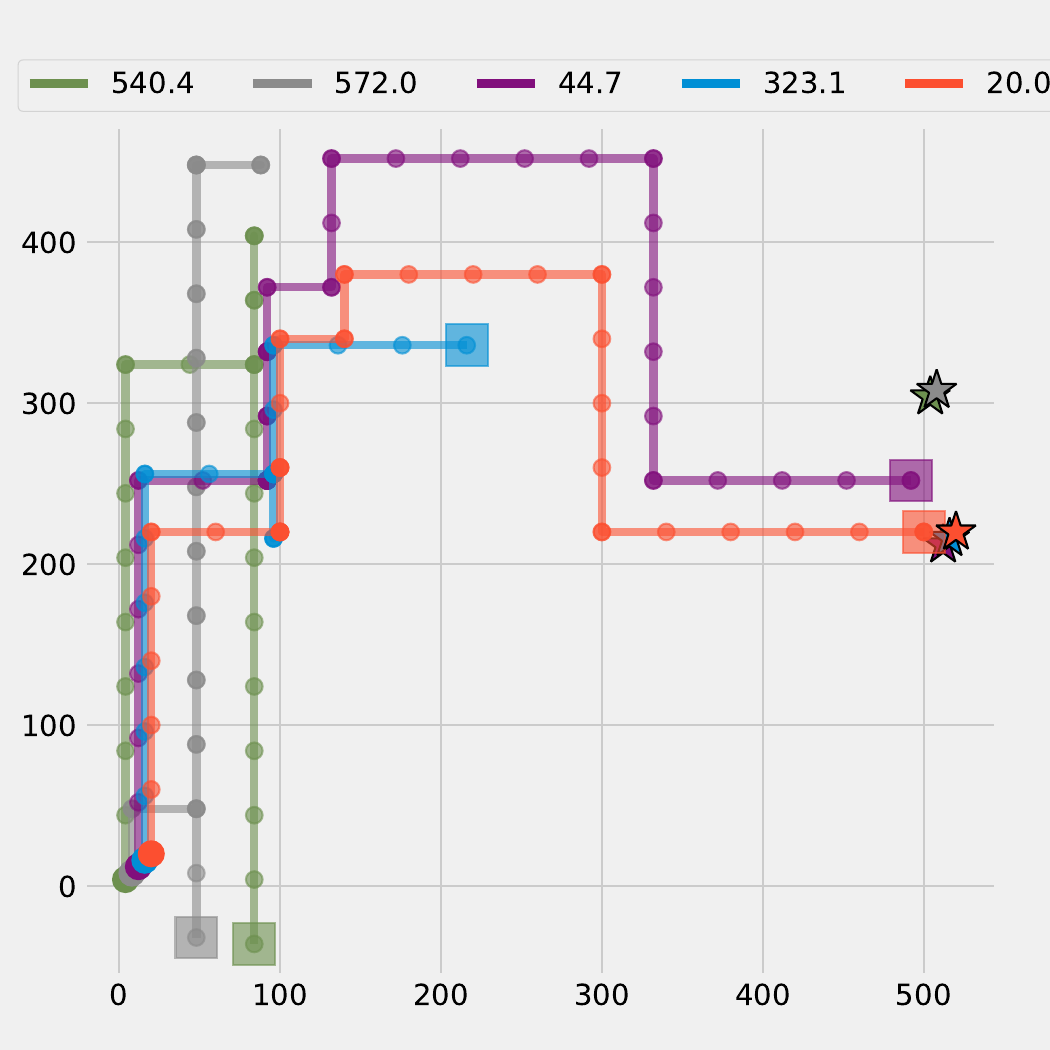} \:
\caption{\textbf{Real World Experiments:} Images and schematic sketch of the
executed trajectory for each of the 5 runs for the 10 test cases that were used
to test the policy in the real world. Runs are off-set from each other for
better visualization. Start location (always $(0,0)$) is denoted by a solid
circle, goal location by a start, and the final location of the agent is
denoted by a square. Legend notes the distance of the goal location from the
final position. Best seen in color on screen.}
\figlabel{realtests}
\end{figure}

%% file: discussion.tex
\section{Discussion}
\seclabel{discussion}
In this paper, we introduced a novel end-to-end neural architecture for navigation in
novel environments. Our architecture learns to map from first-person viewpoints and
uses a planner with the learned map to plan actions for navigating to different
goals in the environment. Our experiments demonstrate that such an approach
outperforms other direct methods which do not use explicit mapping and planning
modules. While our work represents exciting progress towards problems which
have not been looked at from a learning perspective, a lot more needs to be done for 
solving the problem of goal oriented visual navigation in novel environments. 

A central limitations in our work is the assumption of perfect odometry. Robots
operating in the real world do not have perfect odometry and a model that
factors in uncertainty in movement is essential before such a model can be
deployed in the real world.

A related limitation is that of building and maintaining metric representations
of space. This does not scale well for large environments. We overcome this by
using a multi-scale representation for space. Though this allows us to study
larger environments, in general it makes planning more approximate given lower
resolution in the coarser scales which could lead to loss in connectivity
information. Investigating representations for spaces which do not suffer from
such limitations is important future work.

In this work, we have exclusively used \rldagger for training our agents. Though
this resulted in good results, it suffers from the issue that the optimal
policy under an expert may be unfeasible under the information that the agent
currently has. Incorporating this in learning through guided policy search or
reinforcement learning may lead to better performance specially for the
semantic task.

%% file: supp_text.tex
\section{Backward Flow Field $\rho$ from Egomotion}
\seclabel{ego}
Consider a robot that rotates about its position by an angle $\theta$ and then
moves $t$ units forward. Corresponding points $p$ in the original top-view and
$p'$ in the new top-view are related to each other as follows ($R_\theta$ is a
rotation matrix that rotates a point by an angle $\theta$):
\begin{eqnarray}
p' = R_\theta^tp - t \text{ or } p = R_\theta(p'+t)
\end{eqnarray} 
Thus given the egomotion $\theta$ and $t$, for each point in the new top-view
we can compute the location in the original top-view from which it came from.

\section{Mapper Performance in Isolation}
\seclabel{mapper_vis}
To demonstrate that our proposed mapper architecture works we test it in
isolation on the task of free space prediction. We consider the scenario of an
agent rotating about its current position, and the task is to predict free
space in a 3.20 meter neighborhood of the agent. We only provide supervision
for this experiment at end of the agents rotation. \figref{mapper-vis}
illustrates what the mapper learns. Observe that our mapper is
able to make predictions where no observations are made.  We also report
the mean average precision for various versions of the mapper
\tableref{mapper} on the test set (consisting of 2000 locations from
the testing environment). We compare against an analytic mapping
baseline which projects points observed in the depth image into the top view
(by back projecting them into space and rotating them into the top-down view). 

\seclabel{mapper_performance}
\begin{figure}
  \centering
  \insertWL{1.00}{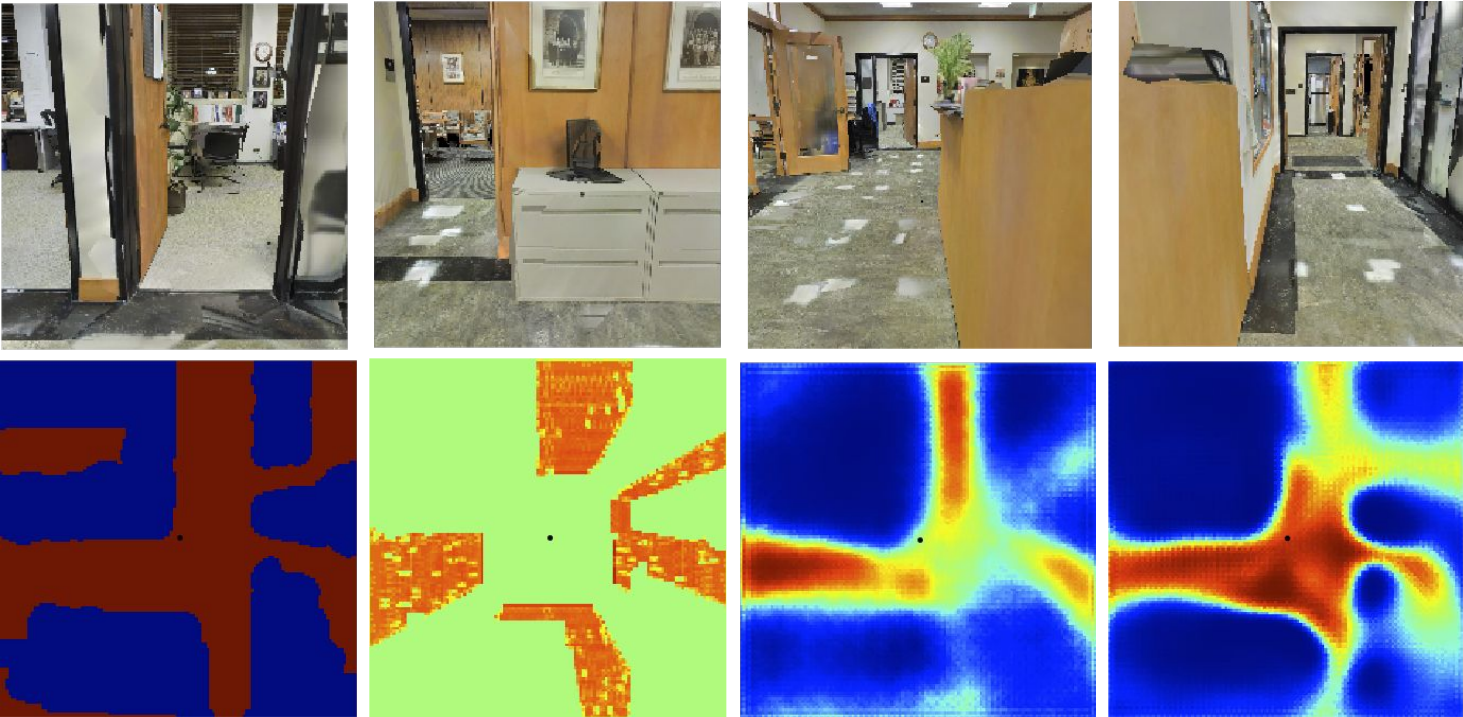}
  \caption{\textbf{Output Visualization for Mapper trained for Free Space
Prediction}: We visualize the output of the mapper when directly trained for
task of predicting free space.  We consider the scenario of an agent rotating
about its current position, the task is to predict free space in a 3.20
meter neighborhood of the agent, supervision for this
experiment at end of the agents rotation.  The top row shows the 4 input views.
The bottom row shows the ground truth free space, predicted free space by
analytically projecting the depth images, learned predictor using \rgb images
and learned predictor using \dd images.  Note that the learned approaches
produce more complete output and are able to make predictions where no
observations were made.}
\figlabel{mapper-vis}
\end{figure}

\begin{table}
\renewcommand{\arraystretch}{1.4} 
\setlength{\tabcolsep}{6pt}
\centering
\footnotesize
\resizebox{1.0\linewidth}{!}{
\begin{tabular}{lcccc} \toprule
Method                  & Modality & CNN Architecture                            & Free Space \\
                        &          &                                             & Prediction AP \\ \midrule
Analytic Projection     & \dd      & -                                           & 56.1\\
Learned Mapper          & \rgb     & ResNet-50                                   & 74.9 \\
Learned Mapper          & \dd      & ResNet-50 Random Init.                      & 63.4 \\
Learned Mapper          & \dd      & ResNet-50 Init. using \cite{gupta2016cross} & 78.4 \\
\bottomrule
 \end{tabular}}
  \caption{\textbf{Mapper Unit Test}: We report average precision for free
space prediction when our proposed mapper architecture is trained directly for
the task of free space prediction on a test set (consisting of 2000 locations
from the testing environment). We compare against an analytic mapping baseline
which projects points observed in the depth image into the top view (by back
projecting them into space and rotating them into the top-down view).} 
\tablelabel{mapper}
\end{table}

\section{Additional Experiments}
\seclabel{supp_exp}

\pparagraph{Additional experiment on an internal Matterport dataset.} We also
conduct experiments on an internal Matterport dataset consisting of 41 scanned
environments. We train on 27 of these environments, use 4 for validation and
test on the remaining 10. We show results for the 10 test environments in
\figref{gmd_error}. We again observe that CMP consistently outperforms the 4
frame reactive baseline and LSTM. 

\begin{figure}
\centering
\insertWL{1.0}{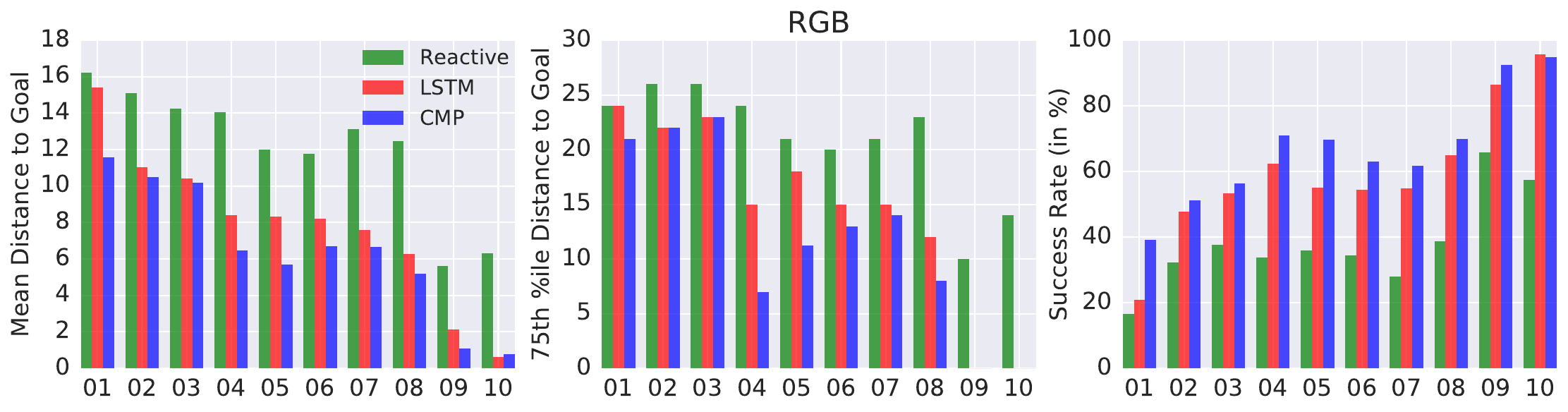}
\insertWL{1.0}{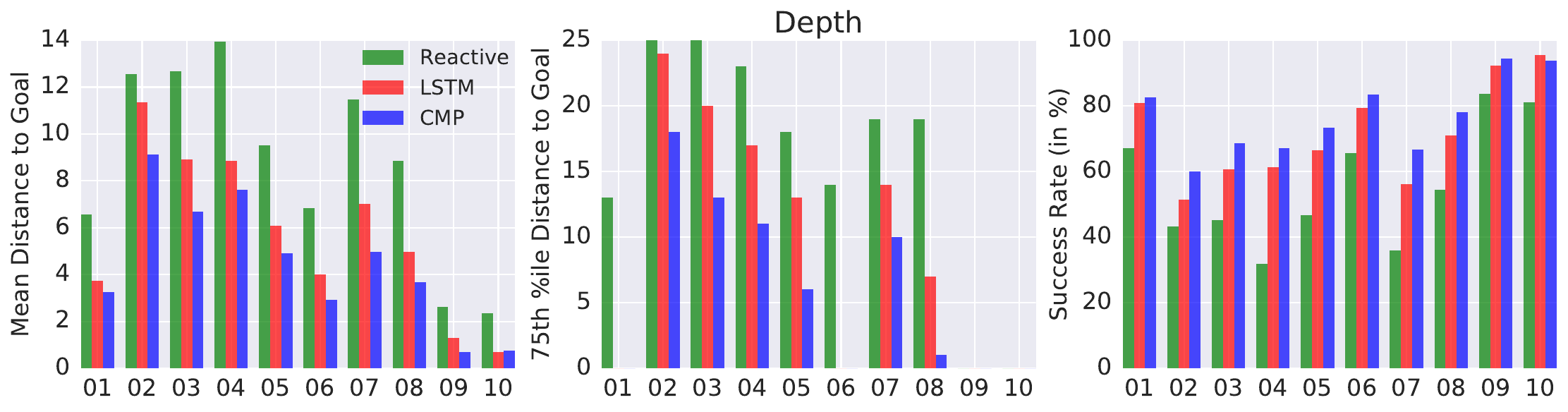} 
\caption{We report the mean distance to goal, 75\textsuperscript{th} percentile
distance to goal (lower is better) and success rate (higher is better) for
Reactive, LSTM and CMP based agents on different test environments from an
internal dataset of Matterport scans. We show performance when using RGB images
(top row) and depth images (bottom row) as input. We note that CMP consistently
outperforms Reactive and LSTM based agents.} \figlabel{gmd_error}
\end{figure}

\begin{figure}
  \centering
  \insertWL{1.0}{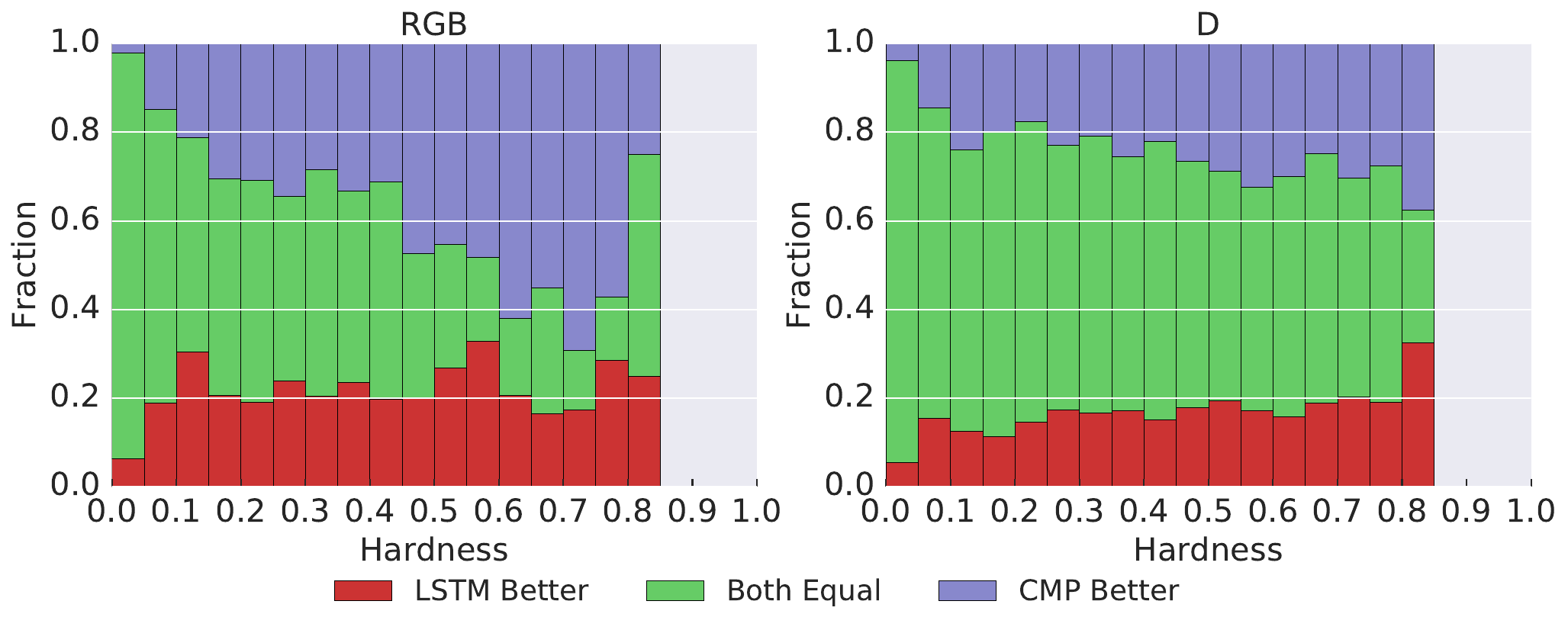} \caption{We show how performance of LSTM
and CMP compare across geometric navigation tasks of different hardness. We
define hardness as the gap between the ground truth and heuristic (Manhattan)
distance between the start and goal, normalized by the ground truth distance.
For each range of hardness we show the fraction of cases where LSTM gets closer
to the goal (LSTM Better), both LSTM and CMP are equally far from the goal
(Both Equal) and CMP gets closer to goal than LSTM (CMP Better). We show
results when using RGB images as input (left plot) and when using Depth images
as input (right plot). We observe that CMP is generally better across all
values of hardness, but for RGB images it is particularly better for cases with
high hardness.}
\figlabel{hardness}
\end{figure}
\input{nav_results_table_supp}

\begin{figure}
\centering
\insertW{0.490}{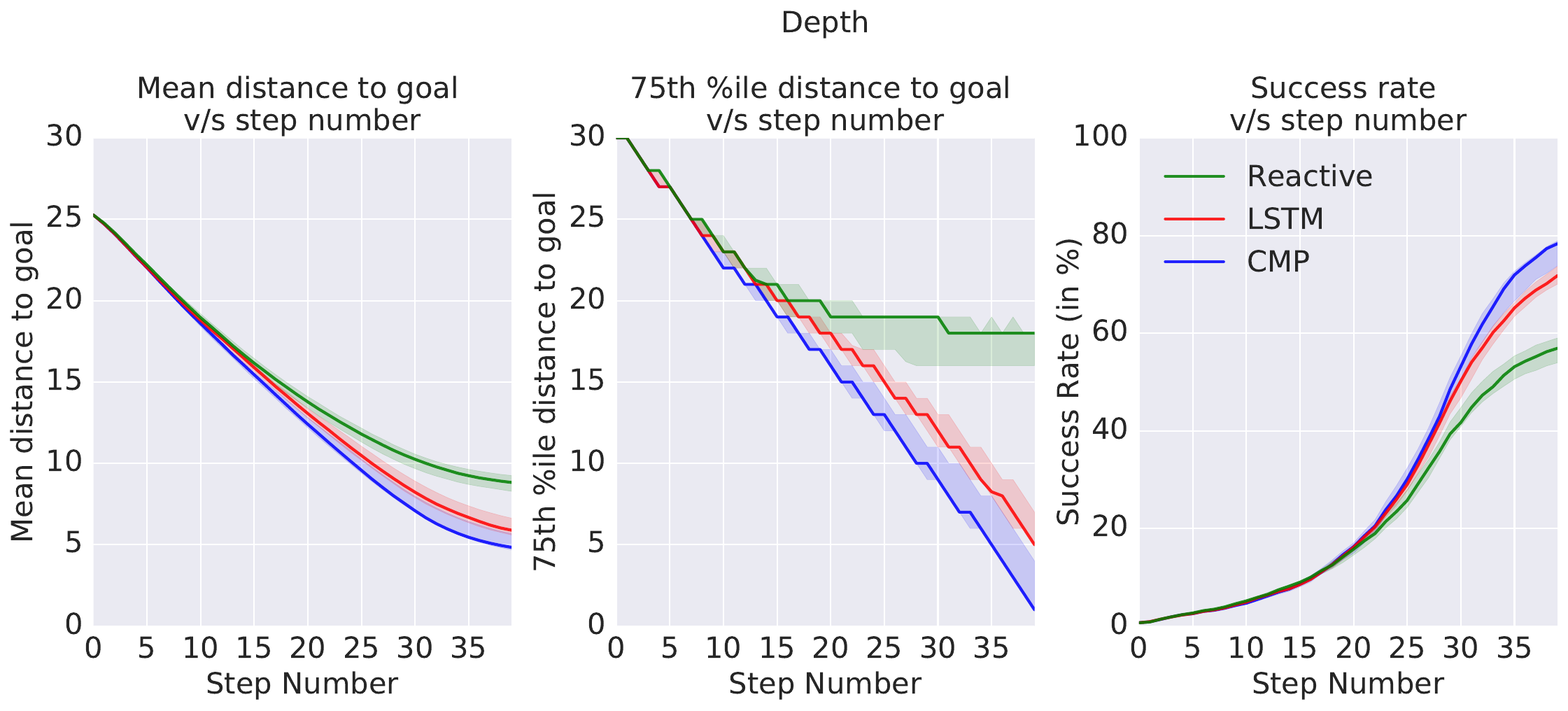} 
\caption{
We show the variance in performance over five re-trainings from different
random initializations of the agents when using depth images as input (the
solid line plots the median performance and  the surrounding shaded region
represents the minimum and maximum value across five different runs). We note
that the variation in performance is reasonably small for all models and CMP
consistently outperforms the two baseline.}
\figlabel{all_error}
\end{figure}

\pparagraph{Ablations.} We also present performance of ablated versions of our
proposed method in \tableref{ablation}.

\textit{Single Scale Planning.} We replace the multi-scale planner with a
single-scale planner. This results in slightly better performance but comes at
the cost of increased planning cost.

\textit{No Planning.} We swap out the planner CNN with a shallower CNN. This
also results in drop in performance specially for the RGB case as compared to
the full system which uses the full planner.

\textit{Analytic Mapper.} We also train a model where we replace our learned
mapper for an analytic mapper that projects points from the depth image into
the overhead view and use it with a single scale version of the planner. We
observe that this analytic mapper actually works worse than the learned one
thereby validating our architectural choice of learning to map. 

\pparagraph{Additional comparisons between LSTM and CMP.} We also report
additional experiments on the Stanford S3DIS dataset to further compare the
performance of the LSTM baseline with our model in the most competitive
scenario where both methods use depth images. These are reported in
\tableref{scenarios}.  We first evaluate how well do these models perform in
the setting when the target is much further away (instead of sampling problems
where the goal is within 32 time steps we sample problems where the goal is 64
times steps away). We present evaluations for two cases, when this agent is run
for 79 steps or 159 steps (see `Far away goal' rows in \tableref{scenarios}).
We find that both methods suffer when running for 79 steps only, because of
limited time available for back-tracking, and performance improves when running
these agents for longer. We also see a larger gap in performance between LSTM
and CMP for both these test scenarios, thereby highlighting the benefit of our
mapping and planning architecture. 

We also evaluate how well these models generalize when trained on a single
scene (`Train on 1 scene'). We find that there is a smaller drop in performance
for CMP as compared to LSTM. We also found CMP to transfer from internal
Matterport dataset to the Stanford S3DIS Dataset slightly better (`Transfer
from internal dataset'). 

We also study how performance of LSTM and CMP compares across geometric
navigation tasks of different hardness in \figref{hardness}. We define hardness
as the gap between the ground truth and heuristic (Manhattan) distance between
the start and goal, normalized by the ground truth distance. For each range of
hardness we show the fraction of cases where LSTM gets closer to the goal (LSTM
Better), both LSTM and CMP are equally far from the goal (Both Equal) and CMP
gets closer to goal than LSTM (CMP Better). We observe that CMP is generally
better across all values of hardness, but for RGB images it is particularly
better for cases with high hardness.

% \pparagraph{Semantic Task.}
% \tableref{semantic_results} reports per category performance for the semantic
% task. We also report an experiment where objects locations are explicitly
% marked by an easily identifiable `marker' (a floating cube at the location of
% the chair, reported as `ST + Markers' in \tableref{semantic_results}). We found
% this to boost performance suggesting incorporating appearance models in the
% form of object detectors from large-scale external datasets will improve
% performance for semantic tasks.  
% 
% \input{semantic_table}

\section{Simulation Testbed Details}
\seclabel{envvis}
We pre-processed the meshes to compute space traversable by the robot. Top
views of the obtained traversable space are shown in
\figsref{area_train1}{area_train2} (training and validation) and \figref{area4}
(testing) and indicate the complexity of the environments we are working with
and the differences in layouts between the training and testing environments.
Recall that robot's action space $\mathcal{A}_{x,\theta}$ consists of
macro-actions.  We pick $\theta$ to be $\pi/2$ which allows us to pre-compute
the set of locations (spatial location and orientation) that the robot can
visit in this traversable space. We also precompute a directed graph
$\mathcal{G}_{x,\theta}$ consisting of this set of locations as nodes and a
connectivity structure based on the actions available to the robot. 

Our setup allows us to study navigation but also enables us to independently
develop and design our mapper and planner architectures. We developed our mapper
by studying the problem of free space prediction from sequence of first person
view as available while walking through these environments. We developed our
planner by using the ground truth top view free space as 2D mazes to plan paths
through. Note that this division was merely done to better understand each
component, the final mapper and planner are trained jointly and there is no
restriction on what information gets passed between the mapper and the planner.

\newcommand{\defaulttext}[0]{Light area shows traversable space. Red bar in the
corner denotes a length of $32$ units ($12.80$ metres).  We also show some
example geometric navigation problems in these environments, the task is to go
from the circle node to the star node.} 
\begin{figure}
\centering
\insertH{0.2235}{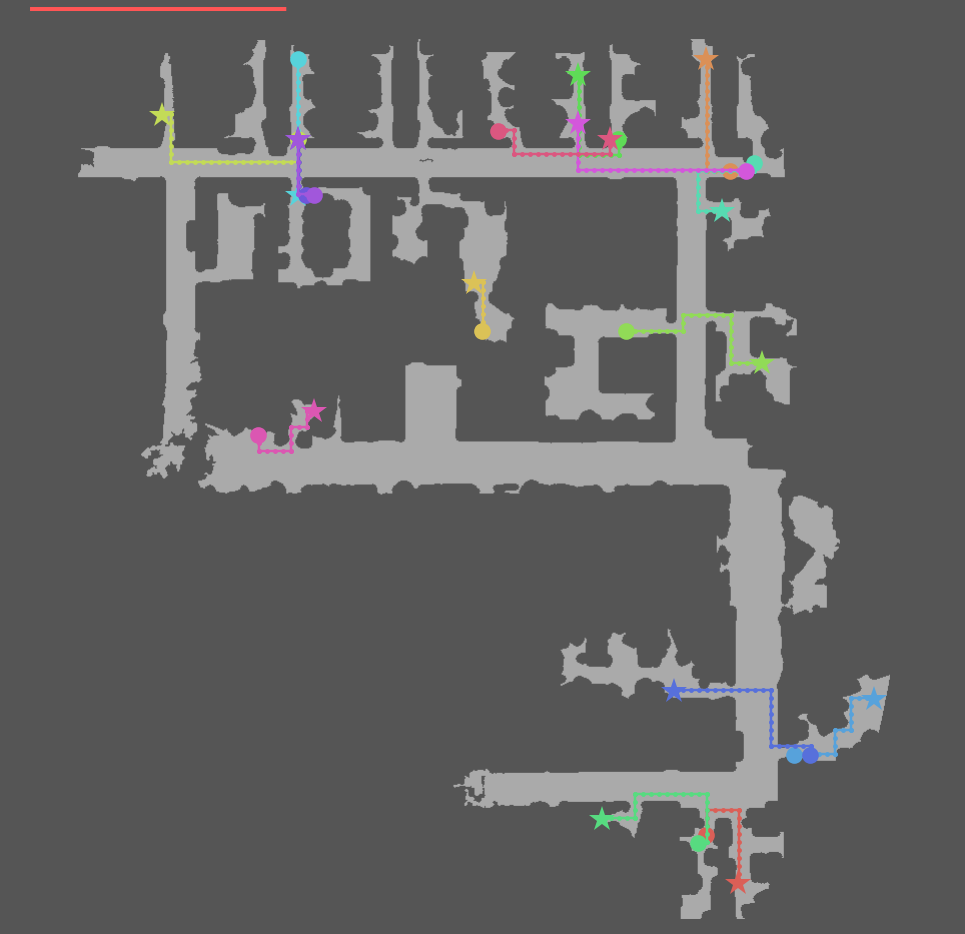} \hfill
\insertH{0.2235}{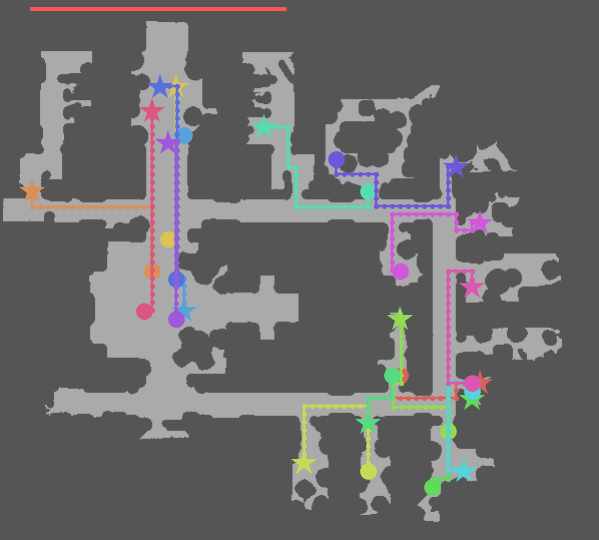}
\includegraphics[height=1.0\linewidth,angle=90]{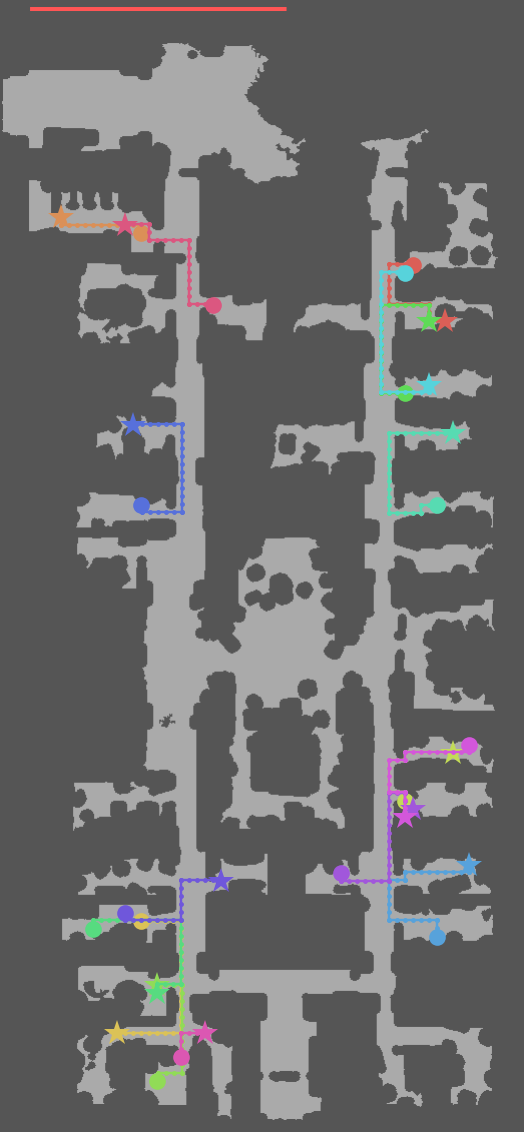}
\includegraphics[height=1.0\linewidth,angle=90]{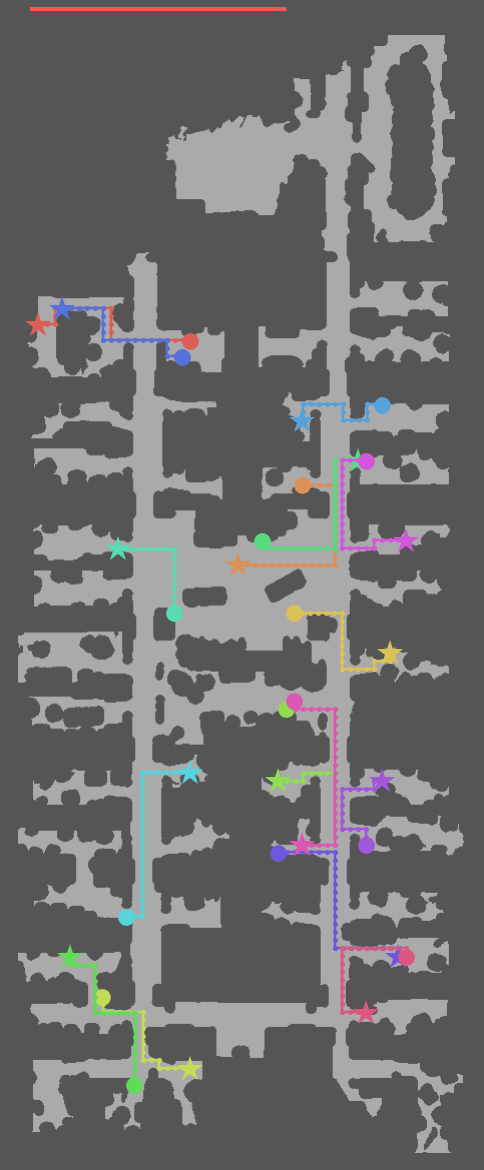}
\caption{Maps for \textit{area52}, \textit{area3} \textit{area1} and
\textit{area6}.  \defaulttext}
\figlabel{area_train1}
\end{figure}

\begin{figure}
\centering
\includegraphics[height=\linewidth,angle=90]{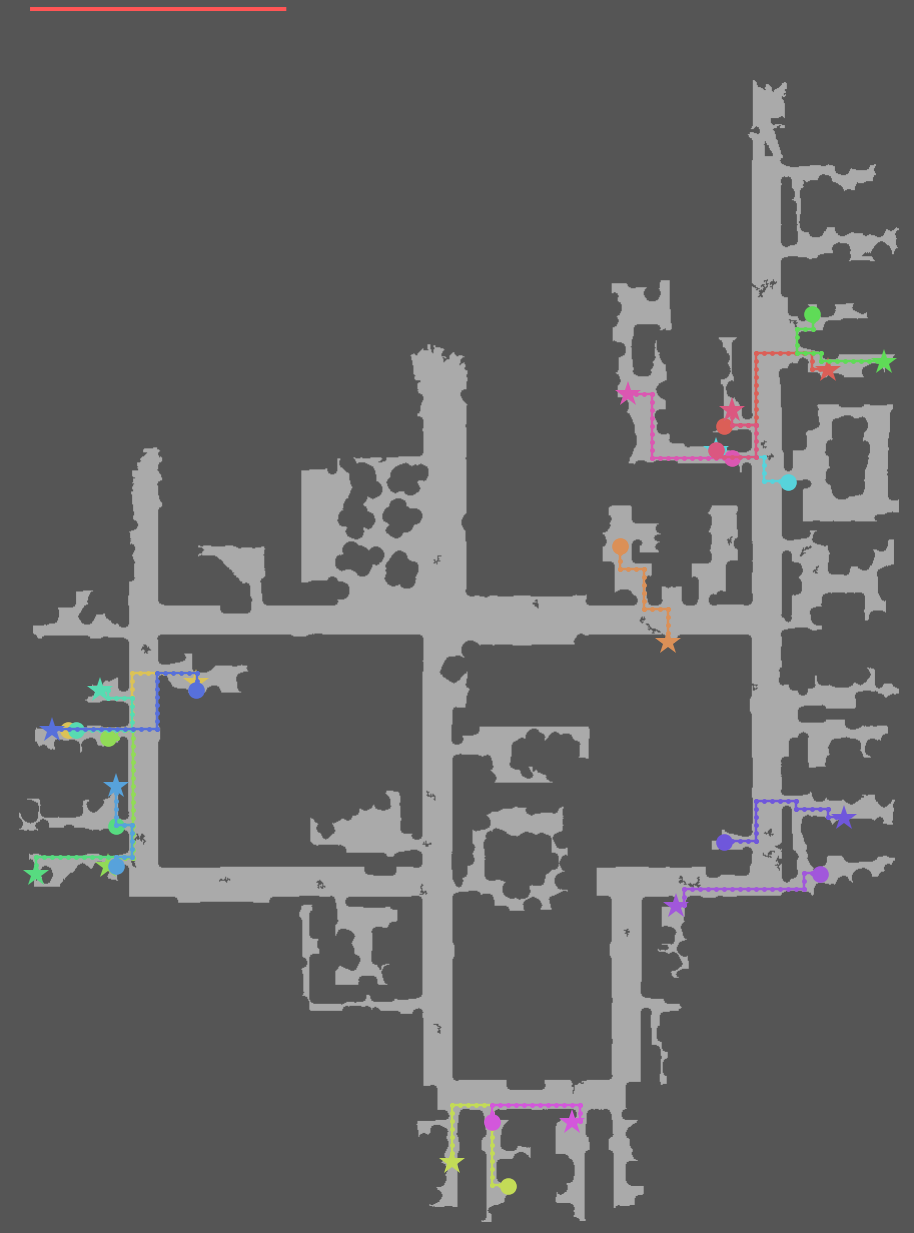} \\
\caption{Map for \textit{area51}. \defaulttext}
\figlabel{area_train2}
\end{figure}

\begin{figure}
\centering
\includegraphics[width=1.00\linewidth,trim={0  6cm 0 0}, clip]{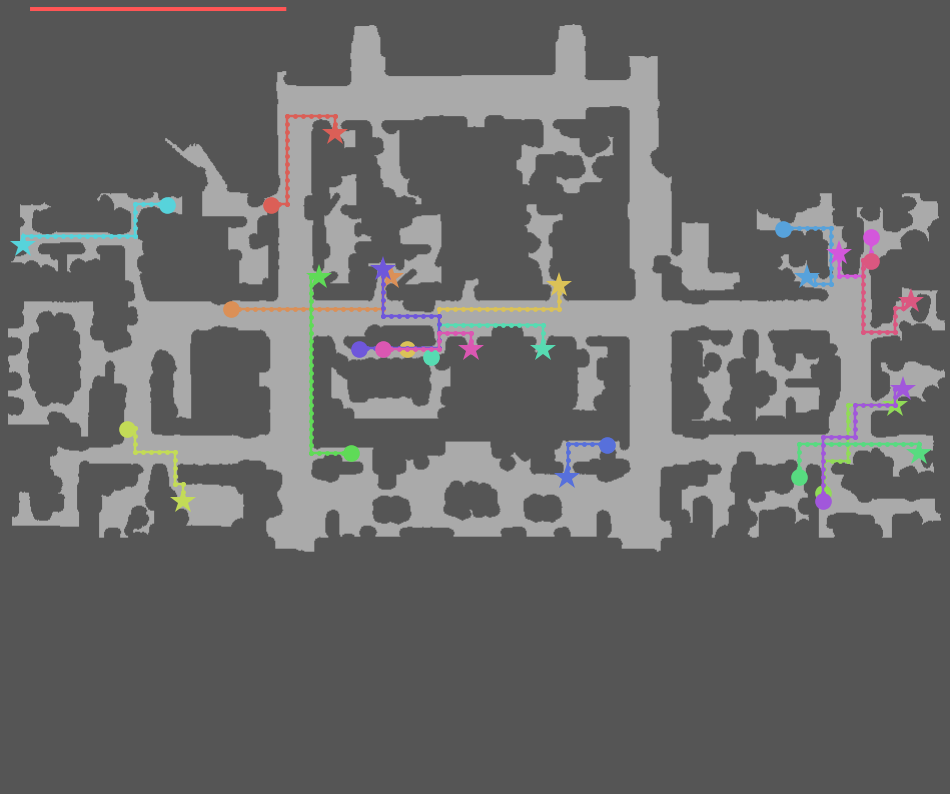}
\caption{Map for \textit{area4}. This floor was used for
testing all the models. \defaulttext}
\figlabel{area4}
\end{figure}

\input{realrobot-supp}

\input{change_log}

%% file: nav_results_table_supp.tex
\renewcommand{\arraystretch}{1.2} 
\setlength{\tabcolsep}{6pt}
\begin{table}
\centering
\footnotesize
\resizebox{1.0\linewidth}{!}{
\begin{tabular}{lccccccccc}\toprule 
  \multirow{2}{*}{$\quad$ Method} & \multicolumn{2}{c}{Mean} & & \multicolumn{2}{c}{75\textsuperscript{th} \%ile} & & \multicolumn{2}{c}{Success \%age} \\ 
  \cmidrule(l{4pt}r{4pt}){2-3} \cmidrule(l{4pt}r{4pt}){5-6} \cmidrule(l{4pt}r{4pt}){8-9}
  $\quad$                              & \rgb          & Depth       &  & \rgb          & Depth       &  & \rgb          & Depth \\ \midrule
  \textbf{Geometric Task} \\
  $\quad$ Initial                      & \Ainit        & \Ainit      &  & \Binit        & \Binit      &  & \Cinit        & \Cinit \\
  $\quad$ No Image LSTM                & \Ablind       & \Ablind     &  & \Bblind       & \Bblind     &  & \Cblind       & \Cblind \\
  $\quad$ CMP \\
  $\quad$ $\quad$ Full model           & \AcmpRgb      & \AcmpD      &  & \BcmpRgb      & \BcmpD      &  & \CcmpRgb      & \CcmpD  \\
  $\quad$ $\quad$ Single-scale planner & \AcmpNomsRgb  & \AcmpNomsD  &  & \BcmpNomsRgb  & \BcmpNomsD  &  & \CcmpNomsRgb  & \CcmpNomsD \\
  $\quad$ $\quad$ Shallow planner      & \AcmpNoVinRgb & \AcmpNoVinD &  & \BcmpNoVinRgb & \BcmpNoVinD &  & \CcmpNoVinRgb & \CcmpNoVinD \\
  $\quad$ $\quad$ Analytic map         & -             & \AcmpAm     &  & -             & \BcmpAm     &  & -             & \CcmpAm \\
  \bottomrule
  \end{tabular}}
\caption{\textbf{Ablative Analysis for CMP:} We follow the same experimental setup as used for table in the main text. See text for details.}
\tablelabel{ablation}
\end{table}

\renewcommand{\arraystretch}{1.2} 
\setlength{\tabcolsep}{2pt}
\begin{table}
\centering
\footnotesize
\resizebox{1.0\linewidth}{!}{
\begin{tabular}{lcccccccccc} \toprule
  & \multicolumn{3}{c}{Mean} & \multicolumn{3}{c}{75\textsuperscript{th} \%ile} & \multicolumn{3}{c}{Success Rate (in \%)} & \\
  \cmidrule(l{5pt}r{5pt}){2-4} \cmidrule(l{5pt}r{5pt}){5-7} \cmidrule(l{5pt}r{5pt}){8-10}
                                                               & Init.          & LSTM            & CMP            & Init.          & LSTM            & CMP            & Init.          & LSTM            & CMP            & \\ \midrule
  \multicolumn{6}{l}{\textbf{Far away goal (maximum 64 steps away)}} \\
  $\quad$ Run for 79 steps                                     & \Alonginit     & \AlonglstmD     & \AlongcmpD     & \Blonginit     & \BlonglstmD     & \BlongcmpD     & \Clonginit     & \ClonglstmD     & \ClongcmpD     & \\
  $\quad$ Run for 159 steps                                    & \Alonglonginit & \AlonglonglstmD & \AlonglongcmpD & \Blonglonginit & \BlonglonglstmD & \BlonglongcmpD & \Clonglonginit & \ClonglonglstmD & \ClonglongcmpD & \\ \\
  \multicolumn{6}{l}{\textbf{Generalization}} \\
  $\quad$ Train on 1 floor                                     & \Anoallinit    & \AnoalllstmD    & \AnoallcmpD    & \Bnoallinit    & \BnoalllstmD    & \BnoallcmpD    & \Cnoallinit    & \CnoalllstmD    & \CnoallcmpD    & \\
  $\quad$ Transfer from IMD                                    & \Atrinit       & \AtrlstmD       & \AtrcmpD       & \Btrinit       & \BtrlstmD       & \BtrcmpD       & \Ctrinit       & \CtrlstmD       & \CtrcmpD       & \\
  \bottomrule
  \end{tabular}}
\caption{We report additional comparison between best performing models. See
text for details.}
\tablelabel{scenarios}
\end{table}

%% file: realrobot-supp.tex
\section{Macro-action Implementation using ILQR}
\seclabel{ilqr}
We use the robot 2D location and orientation as the state
$\vec{s}$, the linear and angular velocity as the control inputs $\vec{u}$ to
the system, and function $f$ to model the dynamics of the system as follows:

\begin{eqnarray}
\vec{s_t} = \begin{bmatrix} x_t\\ y_t\\ \theta_t\end{bmatrix} \
\vec{u_t}= \begin{bmatrix} v_t\\ \omega_t\\ \end{bmatrix} \
f(\vec{s_t}, \vec{u_t}) = \begin{bmatrix}
x_t + v_t\dt \cos(\theta_t) \\
y_t + v_t \dt \sin(\theta_t) \\
\theta_t + \omega_t \dt 
\end{bmatrix}
\end{eqnarray}

Given an initial state $\vec{s_0}$, and a desired final state $\vec{s_T}$ ($=
\vec{0}$ without loss of generality), iLQR solves the following optimization
problem:
\begin{eqnarray}
\argmin_{\vec{u_t}} & \sum_t{\vec{s_t}^tQ\vec{s_t} + \vec{u_t}^tR\vec{u_t}} \\
\text{subject to} & \vec{s_{t+1}} = f(\vec{s_t}, \vec{u_t}) \text{for } t \in
[1, \ldots, T]
\end{eqnarray}
where, matrices $Q$ and $R$ are specified to be appropriately scaled identity
matrices, $\dt$ controls the frequency with which we apply the control input,
and $T$ determines the total time duration we have to finish executing the
macro-action. Matrix $Q$ incentives the system to reach the target state
quickly, and matrix $R$ incentives applying small velocities. The exact scaling
of matrices $Q$ and $R$, $\dt$ and $T$ are set experimentally by running the
robot on a variety of start and goal state pairs.

Given Dubins Car dynamics are non-linear, iLQR optimizes the cost function by
iteratively linearizing the system around the current solution. As mentioned,
iLQR outputs $\vec{x^{ref}_t}$, $\vec{u^{ref}_t}$, and a set of feedback
matrices $K_t$. The control to be applied to the system at time step $t$ is
obtained as $\vec{u^{ref}_t} + K_t \left( \tilde{\vec{s_t}}-\vec{s^{ref}_t}
\right)$, where $\tilde{\vec{s_t}}$ is the estimated state of the system as
measured from the robots wheel encoders and IMU (after appropriate coordinate
transforms).

%% file: change_log.tex
\section{Change Log}
\textbf{v1.} First Version.
\textbf{v2.} CVPR 2017 Camera Ready Version. Added more experiments for
semantic task. Code made available on project website. 
\textbf{v3.} IJCV Version. Putting work in context of more recent work in the
area, added SPL metric, added comparisons to classical methods, added details
about real world deployment of the learned policies, added more visualizations
of what is being learned inside the network.